\documentclass{article}
\PassOptionsToPackage{numbers,sort,comma,compress}{natbib}
\usepackage{natbib}

\usepackage{graphicx,xcolor}
\graphicspath{{images/}}
\usepackage[list=true]{subcaption}

\usepackage{mathtools}
\mathtoolsset{showonlyrefs}

\definecolor{light-blue}{rgb}{0.6,0.6,1}
\definecolor{orange}{rgb}{1,0.58,0	}
\definecolor{purplemountainmajesty}{rgb}{0.59, 0.47, 0.71}
\definecolor{applegreen}{rgb}{0.55, 0.71, 0.0}
\definecolor{hanpurple}{rgb}{0.32, 0.09, 0.98}
\definecolor{green(ryb)}{rgb}{0.4, 0.69, 0.2}
\definecolor{forestgreen(web)}{rgb}{0.13, 0.55, 0.13}
\definecolor{amethyst}{rgb}{0.6, 0.4, 0.8}
\definecolor{darkbyzantium}{rgb}{0.36, 0.22, 0.33}
\definecolor{darkslateblue}{rgb}{0.28, 0.24, 0.55}
\definecolor{darkblue}{rgb}{0.0, 0.0, 0.55}
\newcommand{\WF}[1]{\textcolor{darkblue}{#1}}
\usepackage{multirow}
\usepackage[rightcaption]{sidecap}
\usepackage{stmaryrd}
\usepackage{array}
\usepackage{floatrow}
\usepackage{amsmath}
\usepackage{pifont}
\newcommand{\cmark}{\ding{51}}%
\usepackage{graphicx} 

\newcommand\mlp{\textsc{MLPNet}}

\newcommand\cifar{\textsc{CIFAR10}}
\newcommand\mnist{\textsc{MNIST}}

\usepackage{wrapfig}

\newcommand{\ra}[1]{\renewcommand{\arraystretch}{#1}}
\newcommand{\lagr}{\mathcal{L}}
\newcommand{\w}{\mathbf{w}}
\newcommand{\e}{\mathbf{e}}
\newcommand{\tile}{\mathbf{\tilde{e}}}
\newcommand{\delw}{\delta \mathbf{w}}

\newcommand{\gradw}{\nabla_\w L}

\newcommand{\hess}{\mathbf{H}}
\newcommand{\hessinv}{\mathbf{H}^{-1}}
\newcommand{\hessqq}{\lbrack\hessinv\rbrack_{qq}}
\newcommand{\dell}{\delta L}
\renewcommand{\vec}[1]{\mathbf{#1}}

\usepackage{amsmath,amsfonts,bm}
\usepackage{amssymb}

\def\1{\bm{1}}

\def\va{{\mathbf{a}}}

\def\vg{{\mathbf{g}}}

\def\vs{{\mathbf{s}}}

\def\vu{{\mathbf{u}}}
\def\vv{{\mathbf{v}}}
\def\vw{{\mathbf{w}}}
\def\vx{{\mathbf{x}}}
\def\vy{{\mathbf{y}}}

\providecommand{\1}{\mathbf{1}}

\newcommand{\R}{\mathbb{R}}

 \usepackage[final]{neurips_2020}

\usepackage{appendix}

\usepackage[subfigure]{tocloft}

\usepackage[utf8]{inputenc} 
\usepackage[T1]{fontenc}    
\usepackage{hyperref}       
\usepackage{url}            
\usepackage{booktabs}       
\usepackage{amsfonts}       
\usepackage{nicefrac}       
\usepackage{microtype}      

\title{WoodFisher: Efficient Second-Order Approximation \\ for Neural Network Compression}

\author{%
	Sidak Pal Singh\thanks{Work done while at 	IST Austria.}\\
ETH Zurich, Switzerland\\
	\texttt{contact@sidakpal.com} \\
	\And
	Dan Alistarh \\ 
	IST Austria \& Neural Magic, Inc. \\
	\texttt{dan.alistarh@ist.ac.at} \\
}

\begin{document}
\maketitle
\addtocontents{toc}{\protect\setcounter{tocdepth}{0}}

\begin{abstract}
Second-order information, in the form of Hessian- or Inverse-Hessian-vector products, is a fundamental tool for solving optimization problems. Recently, there has been significant interest in utilizing this information in the context of deep neural networks; however, relatively little is known about the quality of existing approximations in this context. Our work examines this question, identifies issues with existing approaches, and proposes a method called WoodFisher to compute a faithful and efficient estimate of the inverse Hessian.

\hspace{1em} Our main application is to neural network compression, where we build on the classic Optimal Brain Damage/Surgeon framework. We demonstrate that WoodFisher significantly outperforms popular state-of-the-art methods for one-shot pruning. 
Further, even when iterative, gradual pruning is considered, our method results in a gain in test accuracy over the state-of-the-art approaches, for pruning popular neural networks (like \textsc{ResNet-50}, \textsc{MobileNetV1}) trained on standard image classification datasets such as ImageNet ILSVRC. 
We examine how our method can be extended to take into account first-order information, as well as illustrate its ability to automatically set layer-wise pruning thresholds and perform compression in the limited-data regime. The code is available at the following link, \color{blue}\url{https://github.com/IST-DASLab/WoodFisher}.
\end{abstract}

\section{Introduction}

The recent success of deep learning, e.g.~\citep{lecun2015deep, schmidhuber2015deep}, has brought about significant accuracy improvement in areas such as computer vision~\citep{NIPS2015_5638, He_2016} or natural language processing~\citep{devlin2018bert,Radford2018ImprovingLU}. Central to this performance progression has been the size of the underlying models, with millions or even billions of trainable parameters~\citep{He_2016,devlin2018bert}, a trend which seems likely to continue for the foreseeable future~\citep{rajbhandari2019zero}.  

Deploying such large models is taxing from the performance perspective. This has fuelled a line of work where researchers compress such parameter-heavy deep neural networks into ``lighter'', easier to deploy variants. This challenge is not new, and in fact, results in this direction can be found in the early work on neural networks, e.g.~\citep{NIPS1989_250, mozer1989skeletonization, NIPS1992_647}. 
Thus, most of the recent work to tackle this challenge can find its roots in these classic references~\cite{blalock2020state}, and in particular in the Optimal Brain Damage/Surgeon (OBD/OBS) framework \citep{NIPS1989_250, NIPS1992_647}. Roughly, the main idea behind this framework is to build a local quadratic model approximation based on the second-order Taylor series expansion to  determine  the optimal set of parameters to be removed. (We describe it precisely in Section~\ref{sec:compression}.)

A key requirement to apply this approach is to have an accurate estimate of the inverse Hessian matrix, or at least have access to accurate inverse-Hessian-vector-products (IHVPs). 
In fact, IHVPs are a central ingredient in many parts of machine learning, most prominently for optimization~\cite{nesterov2006cubic,Duchi2010AdaptiveSM,kingma2014adam,martens2015optimizing}, but also in other applications such as influence functions~\cite{koh2017understanding} or continual learning~\citep{Kirkpatrick3521}.
Applying second-order methods at the scale of model sizes described above might appear daunting, and so is often done via coarse-grained approximations (such as diagonal, block-wise, or Kronecker-factorization). However, relatively little is understood about the quality and scalability of such approximations.

\paragraph{Motivation.} Our work centers around two main questions.  
The first is analytical, and asks if  second-order approximations can be both \emph{accurate} and \emph{scalable} in the context of neural network models. 
The second is practical, and concerns the application of second-order approximations to neural network compression. 
In particular, we investigate whether these methods can be competitive with both industrial-scale methods such as (gradual) \emph{magnitude-based pruning}~\cite{gale2019state}, as well as with the series of non-trivial compression methods proposed by researchers over the past couple of years~\cite{zhu2017prune, theis2018faster, wang2019eigendamage, zeng2019mlprune, dettmers2019sparse, Lin2020Dynamic}.

\paragraph{Contribution.} 
We first examine second-order approximation schemes in the context of convolutional neural networks (CNNs). 
In particular, we identify a method of approximating Hessian-Inverse information by leveraging the observed structure of the empirical Fisher information matrix to approximate the Hessian, which is then used in conjunction with the Woodbury matrix identity~\citep{woodbury} to provide iteratively improving approximations of Inverse-Hessian-vector products. 
We show that this method, which we simply call WoodFisher, can be computationally-efficient, and that it faithfully represents the structure of the Hessian even for relatively low sample sizes. 
We note that early variants of this method have been considered previously~\cite{NIPS1992_647,doi:10.1162/089976600300015420}, but we believe we are the first to consider its accuracy, efficiency, and implementability in the context of large-scale deep models, as well as to investigate its extensions. 

To address the second, practical, question, we demonstrate in Section~\ref{sec:compression} how WoodFisher can be used together with variants of the OBD/OBS pruning framework, resulting in state-of-the-art compression of popular convolutional models such as \textsc{ResNet-50}~\citep{He_2016} and \textsc{MobileNet}~\citep{Howard2017MobileNetsEC} on the ILSVRC (ImageNet) dataset~\cite{russakovsky2015imagenet}. 
We investigate two practical application scenarios. 

The first is \emph{one-shot} pruning, in which the model has to be compressed in a single step, without any re-training. 
Here, WoodFisher easily outperforms all previous methods based on approximate second-order information or global magnitude pruning. 
The second scenario is \emph{gradual} pruning, allowing for re-training between pruning steps. 
Surprisingly, even here WoodFisher either matches or outperforms state-of-the-art pruning approaches, including recent dynamic pruners~\cite{Lin2020Dynamic, Kusupati2020SoftTW}. 
Our study focuses on \emph{unstructured} pruning, but we can exhibit non-trivial speed-ups for real-time inference by running on a CPU framework which efficiently supports unstructured sparsity~\cite{NM}.

WoodFisher has several useful features and extensions. 
Since it approximates the full Hessian inverse, it can provide a \emph{global} measure of parameter importance, and therefore removes the need for manually choosing sparsity targets per layer. 
Second, it allows us to apply compression in the limited-data regime, where either e.g. $99\%$ of the training is unavailable, or no data labels are available. 
Third, we show that we can also take into account the first-order (gradient) term in the local quadratic model, which leads to further accuracy gain, and the ability to prune models which are not fully converged.

\section{Background}

\paragraph{Deterministic Setting.} We consider supervised learning, where we are given a training set $S=\lbrace\big(\vx_i, \vy_i\big)\rbrace_{i=1}^N$, comprising of pairs of input examples $\vec{x} \in \mathcal{X}$ and outputs $\vec{y} \in \mathcal{Y}$. The goal is to learn a function $f: \mathcal{X} \mapsto \mathcal{Y}$,  parametrized by weights $\vw \in \R^d$, such that given input $\vec{x}$, the prediction $f(\vec{x}; \vec{w}) \approx \vec{y}$.  We consider the loss function $\ell:\mathcal{Y} \times \mathcal{Y} \mapsto \mathbb{R}$ to measure the accuracy of the prediction. The training loss  $L$ is defined as the average over training examples, i.e., $L(\vec{w})=\frac{1}{N} \sum_{n=1}^{N} \ell\big(\vec{y}_{n}, f\left(\vec{x}_{n}; \vec{w}\right)\big)$.

\textbf{{The Hessian Matrix.}} For a twice differentiable loss $L$, the Hessian matrix $\hess=\nabla^2_\vw L$,  takes into account the local geometry of the loss at a given point $\vw$ and allows building a faithful approximation to it in a small neighbourhood $\delw$ surrounding $\vw$. This is often referred to as the local quadratic model for the loss and is given by $L(\w + \delw) \approx  L(\w) + {\gradw}^{\top} \delta \mathbf{w}+\frac{1}{2} \delta \mathbf{w}^{\top} \mathbf{H} \; \delta  \mathbf{w} $.

\paragraph{Probabilistic Setting.} An alternative formulation is in terms of the underlying joint distribution $Q_{\vx,\vy} = Q_{\vx} \,Q_{\vy| \vx}$. 
The marginal distribution $Q_\vx$ is generally assumed to be well-estimated by the empirical distribution $\widehat{Q}_\vx$ over the inputs in the training set. As our task is predicting the output $\vy$ given input $\vx$, training the model is cast as learning the conditional distribution $P_{\vy|\vx}$, which is close to the true $Q_{\vy| \vx}$. 
If we formulate the training objective as minimizing the KL divergence between these conditional distributions, we obtain  the equivalence between  losses $\ell\big(\vy_n, f(\vx_n;\vw)\big) =- \log \big(p_\vw(\vy_n| \vx_n)\big)$, where $p_\vw$ is the density function corresponding to the model distribution.

\paragraph{The Fisher Matrix.} In the probabilistic view, the Fisher information matrix $F$ of the model's conditional distribution $P_{y|x}$ is defined as, 
\begin{equation}\label{eq:fisher}
	F=\mathrm{E}_{P_{\vx, \vy}}\left[\nabla_\vw \log p_{\vw}(\vx, \vy) \, \nabla_\vw \log p_{\vw}(\vx, \vy)^{\top}\right]\,.
\end{equation}
In fact, it can be proved that the Fisher $F=\mathrm{E}_{P_{\vx, \vy}}\left[- \nabla^2_\vw \log p_\vw(\vx, \vy)\right]\,$. Then, by expressing \(P_{\vy, \vx}=Q_{\vx} P_{\vy | \vx} \approx \widehat{Q}_{\vx} P_{\vy | \vx}\) and under the assumption that the model's conditional distribution $P_{\vy|\vx}$ matches the conditional distribution of the data $\widehat{Q}_{\vy|\vx}$, the Fisher and Hessian matrices are equivalent. 

\paragraph{The Empirical Fisher.} In practical settings, it is common to consider an  approximation to  the Fisher matrix introduced in Eq.~\eqref{eq:fisher}, where we replace the model distribution $P_{\vx, \vy}$ with the empirical training distribution $\widehat{Q}_{\vx, \vy}$. Thus we can simplify the expression of empirical Fisher as follows, 
{\footnotesize
\begin{equation*}
\begin{aligned}
\hat{F} &=\mathrm{E}_{\widehat{Q}_{\vx}}\left[\mathrm{E}_{\widehat{Q}_{\vy | \vx}}\left[\nabla \log p_{\vw}(\vy | \vx) \nabla \log p_{\vw}(\vy | \vx)^{\top}\right]\right] 
\stackrel{(a)}{=} \frac{1}{N} \sum_{n=1}^{N} \underbrace{\nabla \ell\left(\vy_{n}, f\left(\vx_{n}; \vw\right)\right)}_{\nabla \ell_n} \nabla \ell\left(\vy_{n}, f\left(\vx_{n}; w\right)\right)^{\top} 
\end{aligned}
\end{equation*}
}%
where (a) uses the equivalence of the loss between the probabilistic and deterministic settings. 
In the following discussion, we will use a shorthand $\ell_n$ to denote the loss for a particular training example $(\vx_n, \vy_n)$, and refer to the Fisher as \textit{true Fisher}, when needed to make the distinction relative to  empirical Fisher. For a detailed exposition, we refer the reader to \cite{martens2014new, kunstner2019limitations,Singh:277227}.

\section{Efficient Estimates of Inverse-Hessian Vector Products}

Second-order information in the form of Inverse-Hessian Vector Products (IHVP) has several uses in optimization and machine learning~\cite{nesterov2006cubic,Duchi2010AdaptiveSM,kingma2014adam,martens2015optimizing,Kirkpatrick3521,koh2017understanding}.  
Since computing and storing the Hessian and Fisher matrices directly is prohibitive, we will focus on  efficient ways to approximate this information. 

As we saw above, the Hessian and Fisher matrices are equivalent if the model and data distribution match. Hence, under this assumption, the Fisher can be seen as a reasonable approximation to the Hessian. 
Due to its structure, the Fisher is positive semidefinite (PSD), and hence can be made invertible by adding a small diagonal dampening term. This approximation is therefore fairly common~\citep{doi:10.1162/089976698300017746,martens2015optimizing,theis2018faster}, although there is relatively little work examining the quality of this approximation in the context of neural networks.  

Further, one can ask whether the \emph{empirical Fisher} is a good approximation of the true Fisher. 
The latter is known to converge to the Hessian as the training loss approaches zero via relation to Gauss-Newton~\cite{Schraudolph02,kunstner2019limitations}. 
The empirical Fisher does not enjoy this property, but is far more computationally-efficient than the Fisher, as it can be obtained after a limited number of back-propagation steps\footnote{In comparison to empirical Fisher, the true Fisher requires $m \times$ more back-propagation steps for each sampled $\vy$ or $k \times $ more steps to compute the Jacobian when the hessian $\mathbf{H}_\ell$ of the loss  $\ell$ with respect to the network output $f$ can be computed in a closed-form (like for exponential distributions)}. Hence, this second approximation would trade off theoretical guarantees for practical efficiency. In the next section, we examine how these approximations square off in practice for neural networks.

\subsection{The (Empirical) Fisher and the Hessian: A Visual Tour}\label{sec:visual}

\begin{figure}[h]
	\centering
	\begin{subfigure}{0.4\textwidth}
		\centering
		\includegraphics[width=\linewidth]{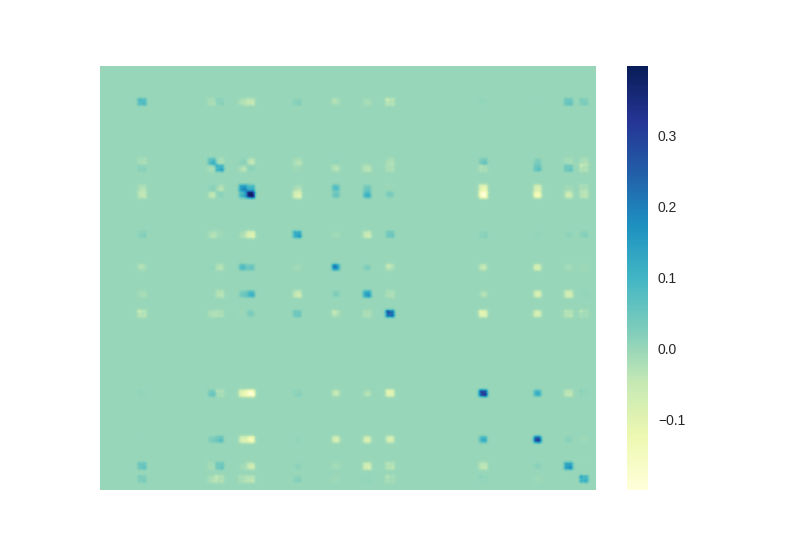}
		\caption{$\hess_{2,2}$}
		\label{fig:cifar_2nd_hess}
	\end{subfigure}
	\begin{subfigure}{0.4\textwidth}
		\centering
		\includegraphics[width=\linewidth]{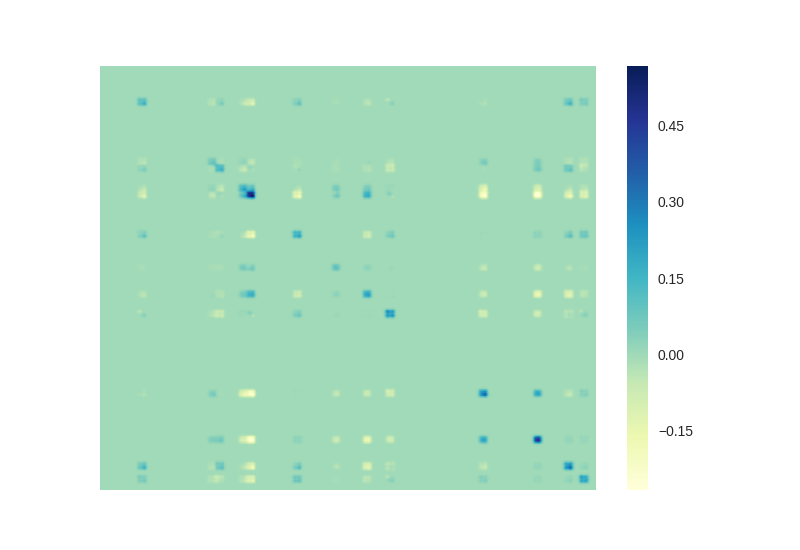}
		\caption{$\widehat{F}_{2,2}$}
		\label{fig:cifar_2nd_ef}
	\end{subfigure}
	\begin{subfigure}{0.4\textwidth}
		\centering
		\includegraphics[width=\linewidth]{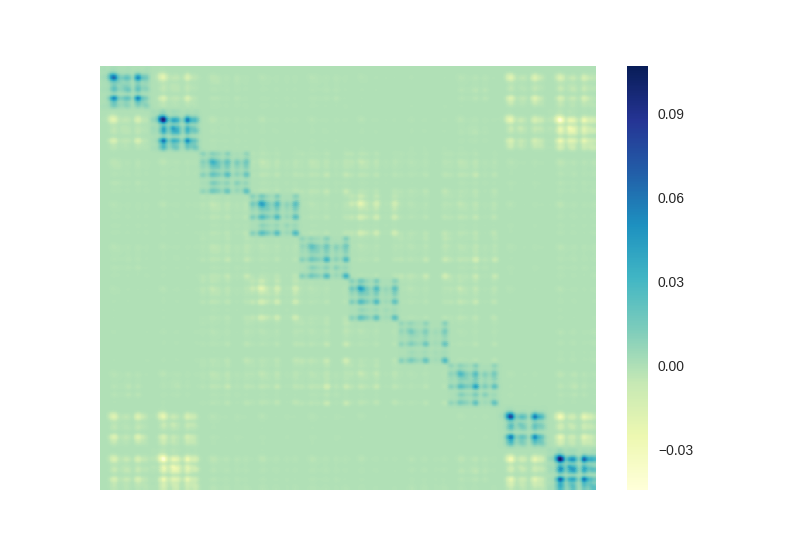}
		\caption{$\hess_{3,3}$}
		\label{fig:cifar_3rd_hess}
	\end{subfigure}
	\begin{subfigure}{0.4\textwidth}
		\centering
		\includegraphics[width=\linewidth]{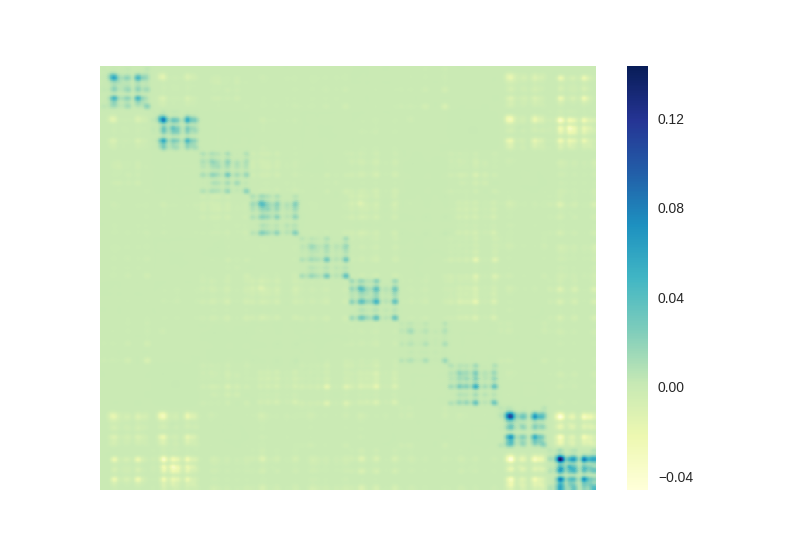}
		\caption{$\widehat{F}_{3,3}$}
		\label{fig:cifar_3rd_ef}
	\end{subfigure}
	\caption{\small \textbf{Similarity in structures of the Hessian and empirical Fisher:} Hessian ($\hess$) and empirical Fisher ($\widehat{F}$) blocks for \textsc{CIFARNet} ($3072 \rightarrow 16 \rightarrow 64 \rightarrow10$) corresponding to second and third hidden layers when trained on \cifar. Figures have been smoothened slightly with a Gaussian kernel for better visibility. Both Hessian and empirical Fisher have been estimated over a batch of $100$ examples in all the figures. } 
	\label{fig:cifar-figures-diag}
\end{figure}

We consider the Hessian ($\hess$) and empirical Fisher ($\hat{F}$) matrices for neural networks trained on standard datasets like \textsc{CIFAR10} and \textsc{MNIST}. 
Due to practical considerations (involving the computation of full Hessian), we consider relatively small models: on \textsc{CIFAR10}, we utilize a fully-connected network with two hidden layers  $3072 \rightarrow 16 \rightarrow 64 \rightarrow10$, which we refer to as \textsc{CIFARNet}. 

Figure~\ref{fig:cifar-figures-diag} compares the Hessian and empirical Fisher blocks corresponding to the second and third hidden layers of this network. 
Visually, there is a clear similarity between the structure of these two matrices, for both the layers. 
A similar trend holds for the first hidden layer as well as the cross-layer blocks, and likewise can be noted in the case of \textsc{MNIST}. 
Surprisingly, this behaviour occurs even if the network is not at full convergence, where we would expect the data and model distribution to match, but even at early stages of training (e.g., after one epoch of training). Please see Appendix \ref{app:hessian_pics_mnist} for full experimental results.
This observation is consistent with recent work~\cite{thomas2019interplay} finding high cosine similarity between the Hessian and empirical Fisher matrices just after a few gradient steps.

As can be noted from the Figure~\ref{fig:cifar-figures-diag}, the main difference between these matrices is not in terms of structure, but in terms of \emph{scale}. 
As a result, we could consider that the empirical Fisher $\hat{F} \propto \hess$, modulo scaling, as long as our target application is not scale-dependent, or if we are willing to adjust the scaling through hyper-parametrization. 
Assuming we are willing to use the empirical Fisher as a proxy for the Hessian, the next question is:  how can we estimate its inverse efficiently?

\subsection{The WoodFisher Approximation}

\paragraph{The Woodbury Matrix Identity.} Clearly, direct inversion techniques would not be viable, since their runtime is cubic in the number of model parameters.  Instead, we start from the Woodbury matrix identity~\citep{woodbury}, which provides the formula for computing the inverse of a low-rank correction to a given invertible matrix $A$. The Sherman-Morrison formula is a simplified variant, given as $\left(A+\vu \vv^{\top}\right)^{-1}=A^{-1}-\frac{A^{-1} \vu \vv^{\top} A^{-1}}{1+\vv^{\top} A^{-1} \vu}$.
We can then express the empirical Fisher as the following recurrence,
\begin{equation}\label{eq:emp-fisher-rec}
\widehat{F}_{n+1} = \widehat{F}_{n} + \frac{1}{N} \nabla {\ell_{n+1}} \nabla {\ell_{n+1}}^\top, \quad \text{where} \quad \widehat{F}_0 = \lambda I_{d}.
\end{equation}
Above, $\lambda$ denotes the \emph{dampening} term, i.e., a positive scalar $\lambda$ times the identity $I_{d}$ to render the empirical Fisher positive definite. 
Then, the recurrence for calculating the inverse of empirical Fisher becomes: 
\begin{equation}\label{eq:woodfisher-inv}
\widehat{F}_{n+1}^{-1}=\widehat{F}_{n}^{-1}-\frac{\widehat{F}_{n}^{-1} \nabla \ell_{n+1} \nabla \ell_{n+1}^{\top} \widehat{F}_{n}^{-1}}{N+\nabla \ell_{n+1}^{\top} \widehat{F}_{n}^{-1} \nabla \ell_{n+1}}, \quad \text{where} \quad \widehat{F}_0^{-1} = \lambda^{-1} I_{d}.
\end{equation}

Finally, we can express the inverse of the empirical Fisher as $\widehat{F}^{-1} = \widehat{F}_{N}^{-1}$, where $N$ denotes the number of examples from the dataset over which it is estimated.  Stretching the limits of naming convention, we refer to this method of using the empirical Fisher in place of Hessian and computing its inverse via the Woodbury identity as \textit{WoodFisher}.

\begin{figure}[h!]
	\centering
	\begin{subfigure}{0.3\textwidth}
		\centering
		\includegraphics[width=\linewidth]{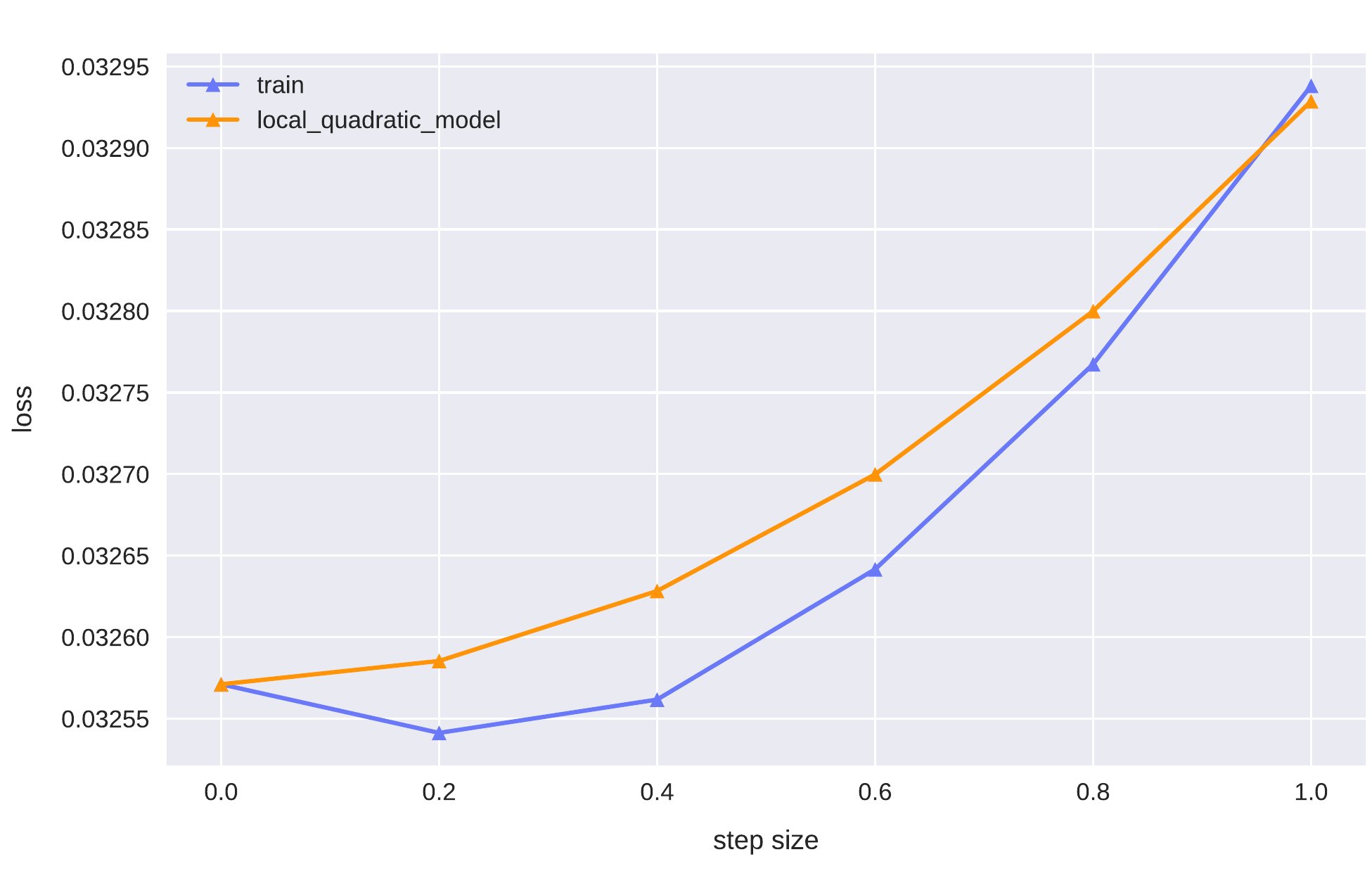}
		\caption{\textsc{layer1.0.conv1}}
		\label{fig:quad-model1}
	\end{subfigure}
	\begin{subfigure}{0.3\textwidth}
		\centering
		\includegraphics[width=\linewidth]{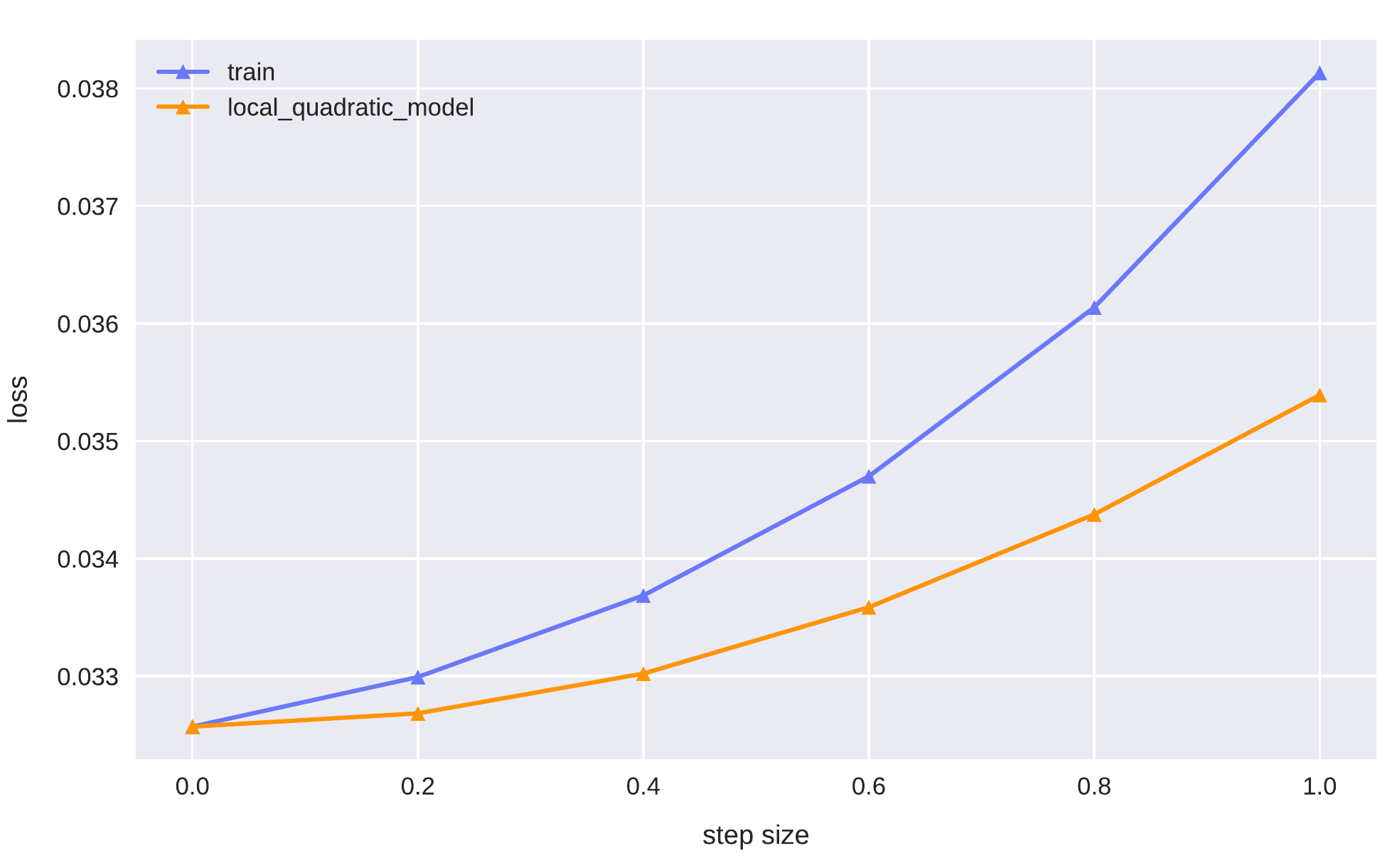}
		\caption{\textsc{layer2.0.conv1}}
		\label{fig:quad-model2}
	\end{subfigure}
	\begin{subfigure}{0.3\textwidth}
		\centering
		\includegraphics[width=\linewidth]{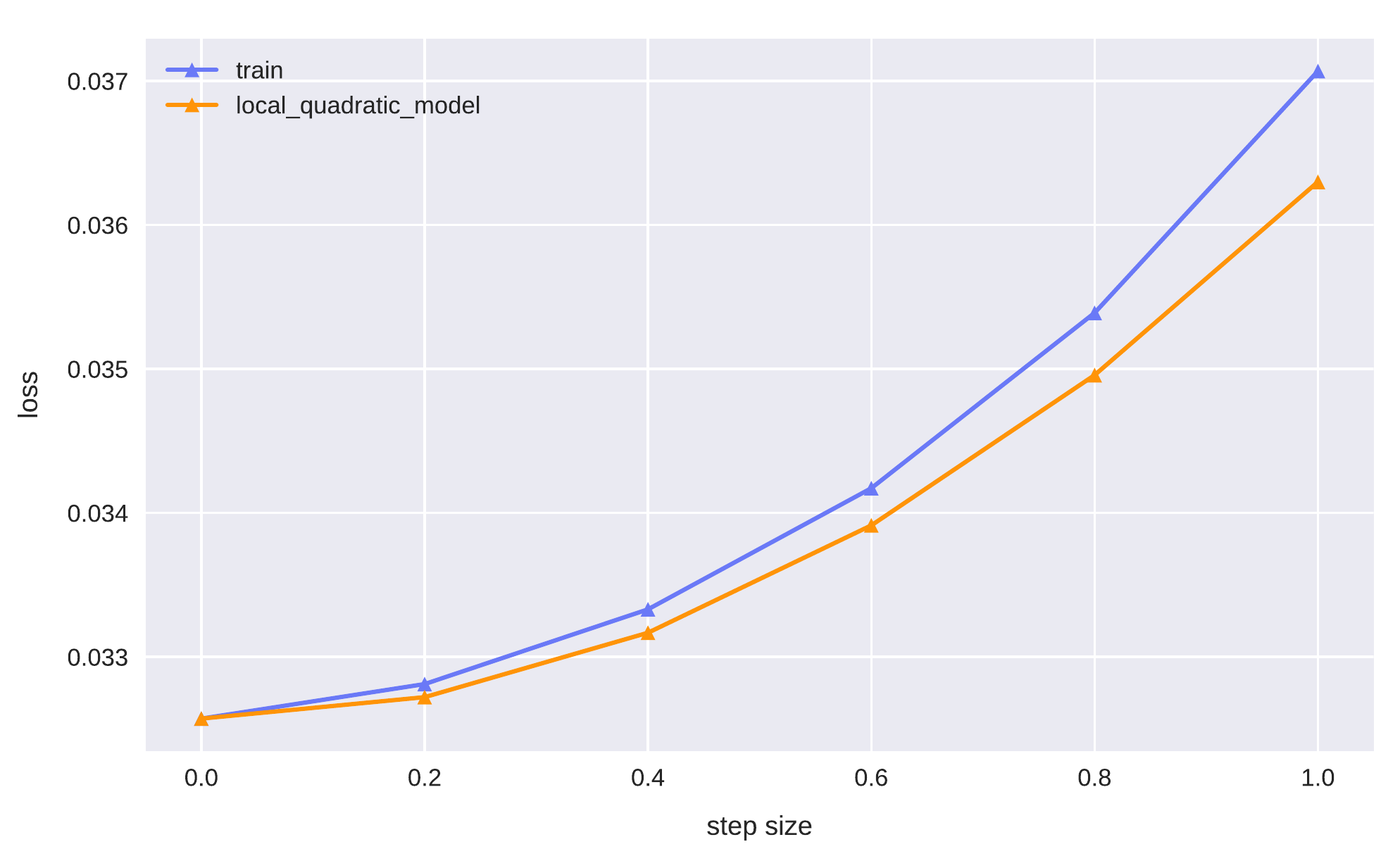}
		\caption{\footnotesize\textsc{layer3.0.conv1}}
		\label{fig:quad-model3}
	\end{subfigure}
	\caption{\small Approximation quality of the loss suggested by the local quadratic model using WoodFisher with respect to the actual training loss. The three plots measure the quality of local quadratic model along three different directions, each corresponding to pruning the respective layers to $50\%$ sparsity. The `blue' curve denotes the actual training loss along the pruning direction, while the `orange' curve is the training loss predicted by the local quadratic model using WoodFisher.}
	\label{fig:approx-quality}
\end{figure}

\paragraph{Approximation quality for the local quadratic model.} To evaluate the accuracy of our local quadratic model estimated via WoodFisher, we examine how the loss predicted by it compares against the actual training loss. (Since we use a pre-trained network, the first-order gradient term is ignored; we will revisit this assumption later.) We test the approximation on three different directions, each corresponding to the pruning direction (of the form $\hess^{-1}\delw$) obtained when compressing a particular layer to $50\%$ sparsity. We choose three layers from different stages of a pre-trained \textsc{ResNet-20} on \textsc{CIFAR10}, 
and Figure~\ref{fig:approx-quality} presents these results. 

In all  three cases, the local quadratic model using WoodFisher predicts an accurate approximation to the actual underlying loss. Further, it is possible to use the dampening $\lambda$ to control whether a more conservative or relaxed estimate is needed. Overall, this suggests that the approach might be fairly accurate, a hypothesis we examine in more detail in Section~\ref{sec:compression}. 

\paragraph{Computational Efficiency and Block-wise Approximation.} 
While the expression in Eq.~\eqref{eq:woodfisher-inv} goes until $n=N$, in practice, we found the method only needs a small subset of examples, $m$, typically ranging between $100$ to $400$. The runtime is thus reduced from cubic to quadratic in $d$, which can still be excessive for neural networks with millions of parameters. 

Thus, for large models we will need to employ a \emph{block-wise approximation}, whereby we maintain and estimate limited-size blocks (`chunks') on the diagonal and ignore the off-diagonal parts. This block-wise simplification is is motivated by the observation that Hessians tend to be diagonally-dominant, and has been employed in previous work, e.g.~\cite{leroux2008topmoumoute,dong2017learning}. 
Assuming uniform blocks of size $c \times c$ along the diagonal, the runtime of this inversion operation becomes $\mathcal{O}(m c d)$, and hence linear in the dimension $d$. This restriction appears necessary for computational tractability.

\subsection{Context and Alternative Methods}

There is a large body of work utilizing second-order information in machine learning and optimization, to the extent that, it is infeasible to discuss every alternative in detail here. 
We therefore highlight the main methods for estimating inverse Hessian-vector products (IHVPs) in our context of neural networks.  See Appendix~\ref{app:ihvp} for detailed discussion. 

A tempting first approach is the \emph{diagonal approximation}, which only calculates the elements along the diagonal, and inverts the resulting matrix. 
Variants of this approach have been employed in optimization~\citep{kingma2014adam,Duchi2010AdaptiveSM} and model compression~\citep{theis2018faster}. 
Yet, as we show experimentally (Figure~\ref{fig:one-shot-cifarx}), this local approximation can be surprisingly inaccurate. 
By comparison, WoodFisher costs an additional constant factor, but provides significantly better IHVP estimates. 
\emph{Hessian-free} methods are another approach, which forgoes  the explicit computation of Hessians \citep{martens_free} in favour of computing IHVP with a vector $\vv$ by solving the linear system $\mathbf{H} \vx = \vv$ for some given $\vx$. 
Unfortunately, a disadvantage of these methods, which we observed practically, is that they require many iterations to converge, since the underlying Hessian matrix can be  ill-conditioned. 
\emph{Neumann-series-based methods}  \citep{krishnan2018neumann,agarwal2016secondorder}  exploit the infinite series expression for the inverse of a matrix with eigenspectrum in $[0, 1]$. 
This does not hold by default for the Hessian, and requires using the Power method to estimate the largest and smallest absolute eigenvalues, which increases cost substantially, while the Power method may fail to return the smallest negative eigenvalue.

\paragraph{K-FAC.} Kronecker-factorization (K-FAC) methods \citep{Heskes2000OnNL, martens2015optimizing} replace the expectation of a Kronecker product between two matrices (that arises in the formulation of Fisher blocks between two layers) as the Kronecker product between the expectations of two matrices. While this is known to be a significant approximation~\citep{martens2015optimizing}, the benefit of K-FAC is that the inverse can be efficiently computed~\citep{ba2016distributed, Osawa_2019,zeng2019mlprune, wang2019eigendamage}. 
However, a significant drawback is that the Kronecker factorization form only exists naturally for fully-connected networks. When applied to convolutional or recurrent neural networks, additional approximations \citep{grosse2016kroneckerfactored,martens2018kroneckerfactored} are required in order to have a Kronecker structure, thus limiting its applicability. Furthermore, even in regards to its efficiency, often approximations like the chunking the layer blocks or channel-grouping are required~\cite{laurent2018an}. For a quantitative comparison against WoodFisher, please refer to Section~\ref{sec:kfac_comparison} where we show how WoodFisher can outperform K-FAC even for fully-connected networks.

\paragraph{WoodFisher.} In this context, with WoodFisher, we propose a new method of estimating second-order information that addresses some of the shortcomings of previous methods, and validate it in the context of network pruning.  We emphasize that a similar approach was suggested in the early works of~\cite{NIPS1992_647,doi:10.1162/089976600300015420} for the case of a one-hidden layer neural network. Our contribution is in extending this idea, and showing how it can scale to modern network sizes. 

The Woodbury matrix identity was also used in L-OBS~\cite{dong2017learning} by defining separate layer-wise objectives, 
and was applied to carefully-crafted blocks at the level of neurons. Our approach via empirical Fisher is more general, and we show experimentally that it yields better approximations at scale (c.f. Figure~\ref{fig:one-shot-lobs} in Section~\ref{sec:one-shot-pruning}).

\paragraph{Use of Empirical Fisher.} \citet{kunstner2019limitations} questioned the use of empirical Fisher since, as the training residuals approach zero, the empirical Fisher goes to zero while the true Fisher approaches the Hessian. However, this rests on the assumption that each individual gradient vanishes for well-optimized networks, which we did not find to hold in our experiments. Further, they argue that a large number of samples  are needed for the empirical Fisher to serve as a good approximation---in our experiments, we find that a few hundred samples suffice for our applications (e.g. Figure~\ref{fig:cifar-figures-diag}).

\section{Model Compression}\label{sec:compression}

This area has seen an explosion of interest in recent years--due to space constraints, we refer the reader to the recent survey of~\cite{blalock2020state} for an overview, and here we mainly focus on closely related work on \emph{unstructured} pruning. 
Broadly, existing methods can be split into four classes: 
(1) methods based on approximate second-order information, e.g.~\cite{dong2017learning, theis2018faster,zeng2019mlprune}, usually set in the classical OBD/OBS framework~\cite{NIPS1989_250, NIPS1992_647}; 
(2) iterative methods, e.g.~\cite{han2015deep, zhu2017prune, gale2019state}, which apply magnitude-based weight pruning in a series of incremental steps over fully- or partially-trained models; 
(3) dynamic methods, e.g.~\cite{Evci2019RiggingTL, dettmers2019sparse, Lin2020Dynamic, Kusupati2020SoftTW}, which prune during regular training and can additionally allow the re-introduction of weights during training;
(4) variational or regularization-based methods, e.g.~\cite{molchanov2017variational,louizos2017learning}. 
Recently, pruning has also been linked to intriguing properties of neural network training~\cite{frankle2018lottery}. 
WoodFisher belongs to the first class of methods, but can be used together with both iterative and dynamic methods. 

\paragraph{Optimal Brain Damage.} 
We start from the idea of pruning (setting to $0$) the parameters which, when removed,  lead to a minimal increase in training loss. 
Denote the dense weights by $\w$, and the new weights after pruning as $\w + \delw$. Using the local quadratic model, we seek to minimize $\delta L = L(\w + \delw) - L(\w)  \approx   {\gradw}^{\top} \delta \mathbf{w} + \frac{1}{2} \delta \mathbf{w}^{\top} \hess \; \delta  \mathbf{w}$. It is often assumed that the network is pruned at a local optimum, which eliminates the first term, and the expression simplifies to Eq.~\eqref{eq:changeloss} ahead:
\begin{equation}\label{eq:changeloss}
\delta L \approx  \frac{1}{2} \delta \mathbf{w}^{\top} \hess \; \delta  \mathbf{w}.
\end{equation}

\subsection{Removing a single parameter $w_q$}\label{sec:zerograd}
Remember, our goal in pruning is to remove parameters that do not change the loss by a significant amount. For now, let us consider the case when just a single parameter at index $q$ is removed. The corresponding perturbation $\delw$ can be expressed by the constraint $\e_q ^\top \delw + w_q = 0$, where $\e_q$ denotes the $q^{\text{th}}$ canonical basis vector. Then pruning can be formulated as finding the optimal perturbation that satisfies this constraint, and the overall problem can written as follows:
\begin{equation}\label{eq:prob_w}
\min _{\delw \, \in  \,\mathbb{R}^d} \;\; \left(\frac{1}{2} \delw^{\top} \, \hess \, \delw\right), \quad \text { s.t. } \quad \e_q ^\top \delw + w_q = 0.
\end{equation}

In order to impose the best choice for the parameter to be removed, we can further consider the following constrained minimization problem.
\begin{equation}\label{eq:prob_w_q}
\min_{q  \,\in  \, [d]} \;\;\; \bigg\lbrace \min _{\delw \, \in  \,\mathbb{R}^d} \;\; \left(\frac{1}{2} \delw^{\top} \, \hess \, \delw\right), \quad \text { s.t. } \quad \e_q ^\top \delw + w_q = 0 \; \bigg\rbrace.
\end{equation}

However, let us first focus on the inner problem, i.e., the one from Eq.~\eqref{eq:prob_w}. As this is a constrained optimization problem, we can consider the Lagrange multiplier $\lambda$ for the constraint and write the Lagrangian $\lagr(\delw, \lambda)$ as follows,
\begin{equation}\label{eq:lagr}
\lagr(\delw, \lambda) = \frac{1}{2} \delw^{\top} \, \hess \, \delw + \lambda \left( \e_q ^\top \delw + w_q \right). 
\end{equation}
The Lagrange dual function $g(\lambda)$, which is the infimum of the Lagrangian in Eq.~\eqref{eq:lagr} with respect to $\w$, can be then obtained by first differentiating Eq.~\ref{eq:lagr} and setting it to $0$, and then substituting the obtained value of $\delw$. These steps are indicated respectively in Eq.~\eqref{eq:lagr_derivative} and Eq.~\eqref{eq:lagr_dual} below.

\begin{equation}\label{eq:lagr_derivative}
\hess \delw + \lambda \e_q = 0 \implies \delw = - \lambda {\hessinv} e_q.
\end{equation}
\begin{equation}\label{eq:lagr_dual}
g(\lambda)\;\; = \;\;\frac{\lambda^2}{2} \e_q^\top \hessinv \e_q - \lambda^2 \e_q^\top \hessinv \e_q + \lambda w_q 
= \;\; -\frac{\lambda^2}{2} \e_q^\top \hessinv \e_q + \lambda w_q.                                              
\end{equation}                                                                                                     

Now, maximizing with respect to $\lambda$, we obtain that the optimal value $\lambda^\ast$ of this Lagrange multiplier as
\begin{equation}\label{eq:opt_mult}                                                                                
\lambda^\ast  = \frac{w_q}{\e_q^\top \hessinv \e_q} = \frac{w_q}{\lbrack\hessinv\rbrack_{qq}}.                     
\end{equation}

The corresponding optimal perturbation, ${\delw}^\ast$, so obtained is as follows:
\begin{equation}\label{eq:opt_delw}
\delw^\ast  = \frac{-w_q \hessinv \e_q}{ \lbrack\hessinv\rbrack_{qq}}.
\end{equation}

Finally, the resulting change in loss corresponding to the optimal perturbation that removes parameter $w_q$ is, 
\begin{equation}\label{eq:opt_loss}
\dell^\ast  = \frac{w_q^2}{2 \, \lbrack\hessinv\rbrack_{qq}}.
\end{equation}

Going back to the problem in Eq.~\eqref{eq:prob_w_q}, the best choice of $q$ corresponds to removing that parameter $w_q$ which has the minimum value of the above change in loss. We refer to this change in loss as the pruning statistic $\rho$, see Eq.~\eqref{eq:prun_stat}, which we compute for all the parameters and then sort them in the descending order of its value. 
\begin{equation}\label{eq:prun_stat}
\boxed{\rho_q  = \frac{w_q^2}{2 \, \lbrack\hessinv\rbrack_{qq}}}.
\end{equation}
investigation of this direction is left for a future work. 

\subsection{Removing multiple parameters at once}\label{sec:app_mult_prune}
For brevity, consider that we are removing two distinct parameters, $q_1$ and $q_2$, without loss of generality. The constrained optimization corresponding to pruning can be then described as follows, 
\begin{equation}\label{eq:prob_w_q_q}
\min_{q_1  \,\in  \, [d],\; q_2 \, \in [d]} \;\;\; \bigg\lbrace \min _{\delw \, \in  \,\mathbb{R}^d} \;\; \left(\frac{1}{2} \delw^{\top} \, \hess \, \delw\right), \quad \text { s.t. } \quad \e_{q_1} ^\top \delw + w_{q_1} = 0, \; \e_{q_2} ^\top \delw + w_{q_2} = 0,  \; \bigg\rbrace.
\end{equation}

We can see how the search space for the best parameter choices ($q_1, q_2$) grows exponentially. In general, solving this problem optimally seems to be out of hand. Although, it could be possible that the analysis might lead to a tractable computation for the optimal solution. Unfortunately, as described below, this is not the case and we will have to resort to an approximation to make things practical. 
\begin{equation}\label{eq:lagr_q_q}
\lagr(\delw, \lambda_1, \lambda_2) = \frac{1}{2} \delw^{\top} \, \hess \, \delw + \lambda_1 \left( \e_{q_1} ^\top \delw + w_{q_1} \right) + \lambda_2 \left( \e_{q_2} ^\top \delw + w_{q_2} \right). 
\end{equation}

The Lagrange dual function $g^\prime(\lambda_1, \lambda_2)$, which is the infimum of the Lagrangian in Eq.~\eqref{eq:lagr_q_q} with respect to $\w$, can be then obtained by first differentiating Eq.~\ref{eq:lagr_q_q} and setting it to $0$, and then substituting the obtained value of $\delw$. These steps are indicated respectively in Eq.~\eqref{eq:lagr_derivative_q_q} and Eq.~\eqref{eq:lagr_dual_q_q} below.
\begin{equation}\label{eq:lagr_derivative_q_q}
\hess \delw + \lambda_1 \e_{q_1} + \lambda_2 \e_{q_2}= 0 \implies \delw = - \lambda_1 {\hessinv} e_{q_1} - \lambda_2 {\hessinv} e_{q_2}.
\end{equation}

As a remark, we denote the Lagrange dual function here by $g^\prime$ instead of $g$ to avoid confusion with the notation for Lagrange dual function in case of a single multiplier.
\begin{equation}\label{eq:lagr_dual_q_q}
g^\prime(\lambda_1, \lambda_2)\;\; = \;\; - \frac{\lambda_1^2}{2} \e_{q_1}^\top \hessinv \e_{q_1} + \lambda_1 w_{q_1}  -\frac{\lambda_2^2}{2} \e_{q_2}^\top \hessinv \e_{q_2} + \lambda_2 w_{q_2} - \lambda_1 \lambda_2 \e_{q_1}^\top \hessinv \e_{q_2}.                                              
\end{equation} 
Comparing this with the case when a single parameter is removed, c.f. Eq.~\eqref{eq:lagr_dual}, we can rewrite Lagrange dual function as follows, 
\begin{equation}\label{eq:lagr_dual_q_q_simplified}
g^\prime(\lambda_1, \lambda_2)\;\; = \;\; g(\lambda_1) + g(\lambda_2) - \lambda_1 \lambda_2\, \e_{q_1}^\top \hessinv \e_{q_2}.                                              
\end{equation} 
We note that dual function is not exactly separable in terms of the dual variables, $\lambda_1$ and $\lambda_2$, unless the off-diagonal term in the hessian inverse corresponding to $q_1, q_2$ is zero, i.e.,  $\lbrack\hessinv\rbrack_{q_1 q_2} = 0$.

To maximize the dual function in Eq.~\eqref{eq:lagr_dual_q_q_simplified} above, we need to solve a linear system with the Lagrange multipliers $\lambda_1, \lambda_2$ as variables. The equations for this system program correspond to setting the respect partial derivatives to zero, as described in Eq.~\eqref{eq:lagr_mult_lp} below, 
\begin{equation}\label{eq:lagr_mult_lp}
\left.
\begin{aligned}
\frac{\partial g^\prime}{\partial \lambda_1} & \; =  \; -\lambda_1 \e_{q_1}^\top \hessinv \e_{q_1} - \lambda_2 \e_{q_1}^\top \hessinv \e_{q_2} + w_{q_1} = 0 \\
\frac{\partial g^\prime}{\partial \lambda_2}  \; & =  \; -\lambda_1 \e_{q_1}^\top \hessinv \e_{q_2} - \lambda_2 \e_{q_2}^\top \hessinv \e_{q_2} + w_{q_2} = 0
\end{aligned}
\quad 
\right\}
\quad
\text{Solve to obtain $\lambda_1^\ast, \lambda_2^\ast$}
\end{equation}

Hence, it is evident that exactly solving this resulting linear system will get intractable when we consider the removal of many parameters at once.

\subsection{Discussion}

\paragraph{Pruning Direction.} As a practical approximation to this combinatorially hard problem, when removing multiple parameters, we  sort the parameters by the pruning statistic $\rho_q$, removing those with the smallest values. The overall weight perturbation in such a scenario is computed by adding the optimal weight update, Eq.~\eqref{eq:opt_delw}, for each parameter that we decide to prune. (We mask the weight update at the indices of removed parameters to zero, so as to adjust for adding the weight updates separately.) We call this resulting weight update \emph{the pruning direction}. 

If the Hessian is assumed to be diagonal,  we recover the pruning statistic of optimal brain damage~\citep{NIPS1989_250}, $\dell^\ast_{\text{OBD}}  = \frac{1}{2} w_q^2 \lbrack\hess\rbrack_{qq}$. Further, if we let the Hessian be isotropic, we obtain the case of (global) \emph{magnitude pruning}, as the  statistic amounts to $\dell^\ast_{\text{Mag}}  = \frac{1}{2} w_q^2$.  Note, magnitude pruning based methods, when combined with intermittent retraining, form one of the leading practical methods~\cite{gale2019state}.

\paragraph{Pruning using WoodFisher.}
We use WoodFisher to get estimates of the Hessian inverse required in Eq.~\eqref{eq:opt_delw}. Next, the decision to remove parameters based on their pruning statistic can be made either independently for every layer, or jointly across the whole network. The latter option allows us to automatically adjust the sparsity distribution across the various layers given a global sparsity target for the network. As a result, we do not have to perform a sensitivity analysis for the layers or use heuristics such as skipping the first or the last layers, as commonly done in prior work. We refer to the latter as \textit{joint (or global)}-WoodFisher and the former as \textit{independent (or layerwise)}-WoodFisher.

\section{Experimental Results}

We now apply WoodFisher to compress commonly used CNNs for image classification. 
We consider both the \emph{one-shot} pruning case, and the \emph{gradual} case, 
as well as investigate how the block-wise assumptions and the number of samples used for estimating Fisher affect the quality of the approximation, and whether this can lead to more accurate pruned models.

\subsection{One-shot pruning}\label{sec:one-shot-pruning}
Assume that we are given a pre-trained neural network which we would like to sparsify in a single step, without any re-training. 
This scenario might arise when having access to limited data, making re-training infeasible. Plus, it allows us to directly compare approximation quality. 

\begin{figure}[h!]
	\centering
	\begin{subfigure}{0.48\textwidth}
		\centering
		\includegraphics[width=\linewidth]{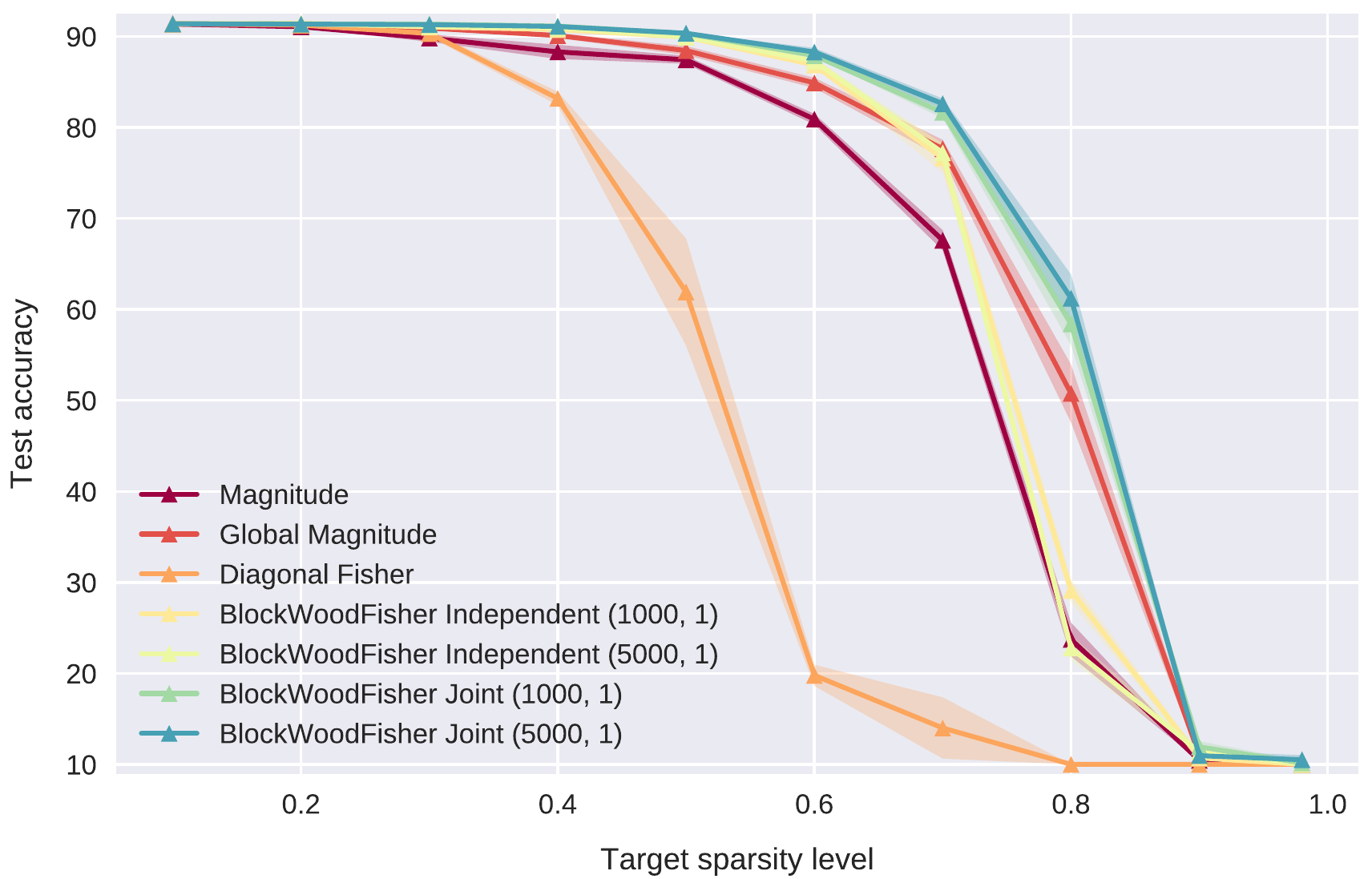}
		\caption{}
		\label{fig:one-shot-cifarx}
	\end{subfigure}
	\begin{subfigure}{0.48\textwidth}
		\centering
		\includegraphics[width=\linewidth]{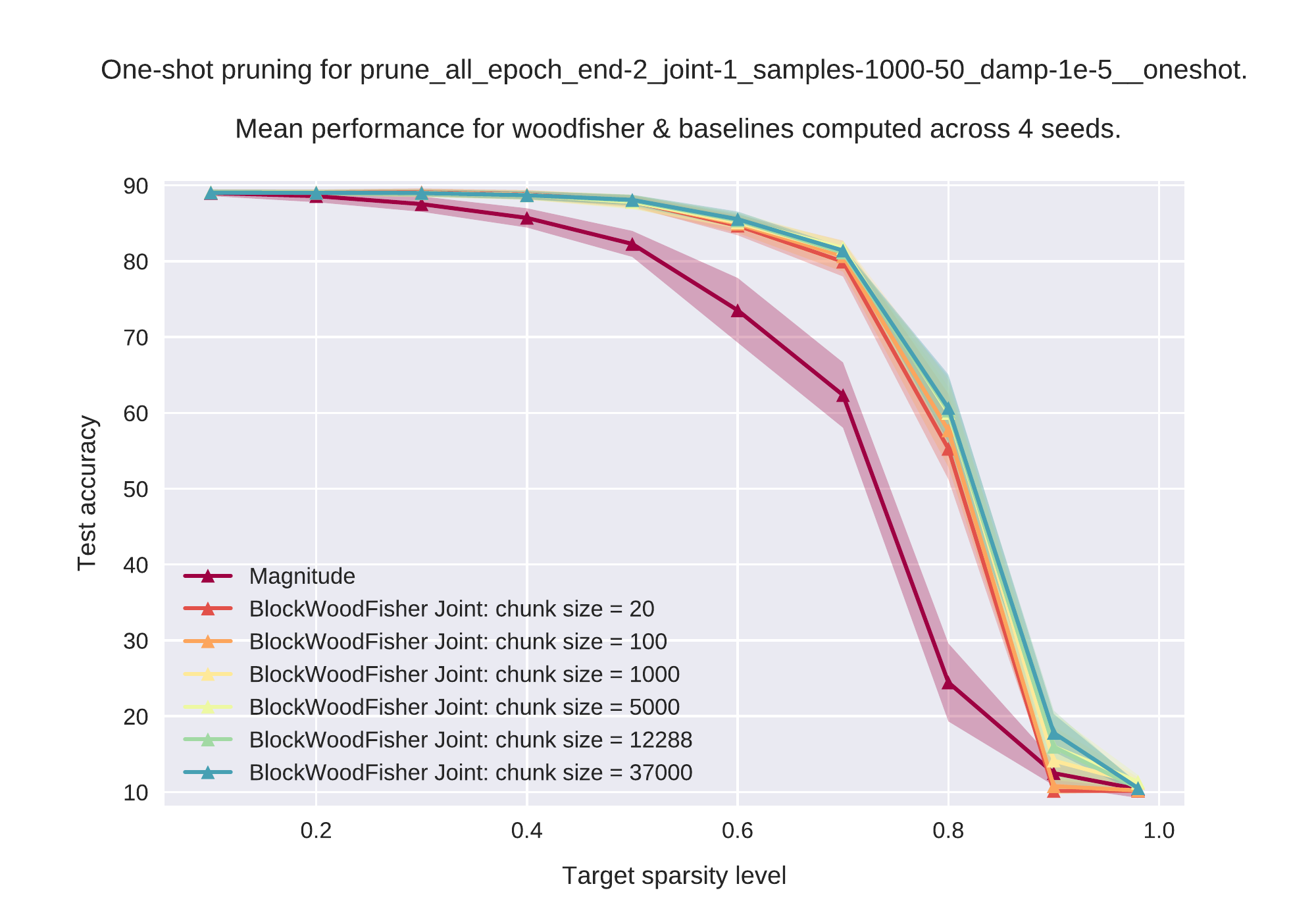}
		\caption{}
		\label{fig:ablation-joint}
	\end{subfigure}
	\caption{\small (a) Comparing the one-shot pruning performance of WoodFisher on \textsc{ResNet-20} and \textsc{CIFAR10}. (b) An ablation showing the effect of chunk size used for WoodFisher.}
\end{figure}

\paragraph{\textsc{ResNet-20}, \textsc{CIFAR10}.} First, we consider a pre-trained \textsc{ResNet-20} \citep{He_2016} network on \textsc{CIFAR10} with $\sim 300K$ parameters. We compute the inverse of the diagonal blocks corresponding to each layer. Figure~\ref{fig:one-shot-cifarx} contains the test accuracy results for one-shot pruning in this setting, averaged across four seeds, as we increase the percentage of weights pruned. 
Despite the block-wise approximation, we observe that both independent- and joint-WoodFisher variants significantly outperform magnitude pruning and diagonal-Fisher based pruning. 

We also compare against the global version of magnitude pruning, which can re-adjust sparsity across layers. 
Still, we find that the global magnitude pruning is worse than WoodFisher-independent until about $60\%$ sparsity, beyond which it is likely that adjusting layer-wise sparsity is essential. WoodFisher-joint performs the best amongst all the methods, and is better than the top baseline of global magnitude pruning - by about $5\%$ and $10\%$ in test accuracy at the $70\%$  and  $80\%$ sparsity levels. 
Finally, diagonal-Fisher performs worse than magnitude pruning for sparsity levels higher than $30\%$. This finding was consistent, and so we omit it in the sections ahead. (We used 16,000 samples to estimate the diagonal Fisher, whereas WoodFisher performs well even with 1,000 samples.)

\paragraph{\textsc{Effect of chunk size.}} For networks which are much larger than \textsc{ResNet-20}, we  also need to split the layerwise blocks into smaller chunks along the diagonal, to maintain efficiency. So, here we discuss the effect of this chunk-size on the performance of WoodFisher. We take the above setting of \textsc{ResNet-20} on \textsc{CIFAR10} and evaluate the performance for chunk-sizes in the set, $\lbrace20, \,100, \,1000, \,5000, \,12288, \,37000\rbrace$. Note that, $37000$ corresponds to the complete blocks across all the layers. Figure~\ref{fig:ablation-joint}  illustrate these results for WoodFisher in the joint mode. We observe that performance of WoodFisher increases monotonically as the size of the blocks (or chunk-size) is increased. This fits well with our expectation that a large chunk-size would lead to a more accurate estimation of the inverse. Similar trend is observed for the independent mode, and is presented in Figure~\ref{fig:app-ablation-indep} of the Appendix~\ref{sec:app_one_shot_detail}.

\begin{figure}[h]
	\centering
	\includegraphics[width=0.7\linewidth]{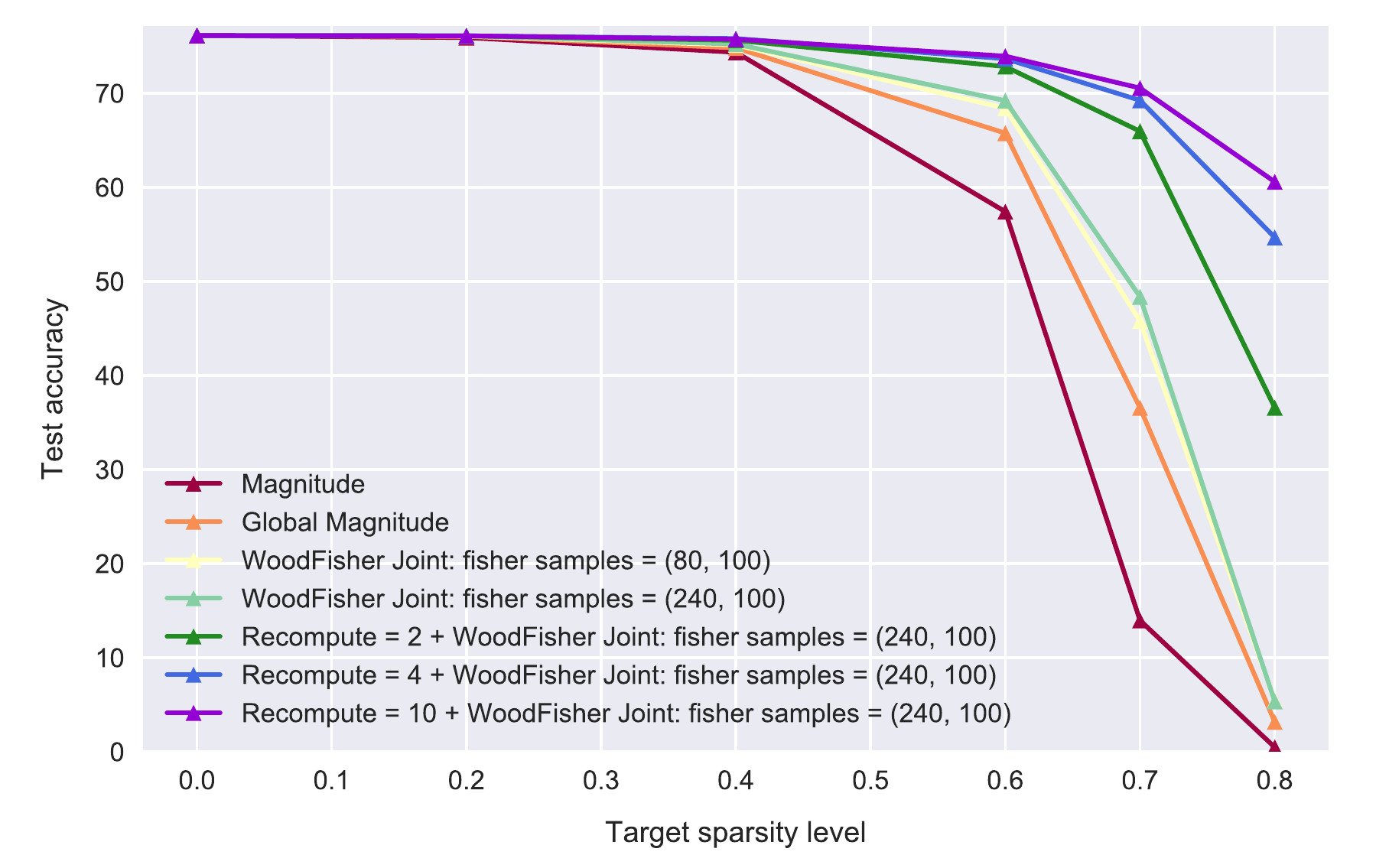}
	\caption{\small One-shot sparsity results for \textsc{ResNet-50} on \textsc{ImageNet}. In addition, we show here the effect of fisher subsample size as well as how the performance is improved if we allow for recomputation of the Hessian (still no retraining). 
		The numbers corresponding to tuple of values called fisher samples refers to (fisher subsample size, fisher mini-batch size). A chunk size of $1000$ was used for this experiment. }
	\label{fig:one-shot-imagenet-main}
\end{figure}

\paragraph{\textsc{ResNet-50}, \textsc{ImageNet}.} Likewise, we performed a one-shot pruning experiment for the larger \textsc{ResNet-50} model ($25.5$M parameters) on ImageNet, which for efficiency we break into layer-wise blocks (chunks) of size 1K. 
We found that this suffices for significant performance gain over layer-wise and global magnitude pruning, c.f. Figure~\ref{fig:one-shot-imagenet-main}. 
In addition, we observed that it is useful to replace individual gradients, in the definition of the empirical Fisher, with gradients averaged over a mini-batch of samples. 
Typically, we use 80 or 240 (i.e., `fisher subsample size') such averaged gradients over a mini-batch of size 100 (i.e., `fisher mini-batch size'). This allows us to exploit mini-batch times more samples, without affecting the cost of  WoodFisher, as only the fisher subsample size dictates the number of Woodbury iterations. 

Besides, we found that accuracy can be further improved if we allow recomputation of the Hessian inverse estimate during pruning (but without retraining), as the local quadratic model is valid otherwise only in a small neighbourhood (or trust-region). These results correspond to curves marked with `Recompute' in Figure~\ref{fig:one-shot-imagenet-main}, where the effect of increasing the recomputation steps is also shown.

Further results can be found in Appendix~\ref{sec:app_one_shot_detail}, including one-shot pruning results of \textsc{MobileNetV1} on \textsc{ImageNet}
, as well as ablation for the effect of chunk-size and number of samples used for Fisher computations.

\begin{figure}[h!]
	\centering
\begin{subfigure}{0.46\textwidth}
	\centering
	\includegraphics[width=\linewidth]{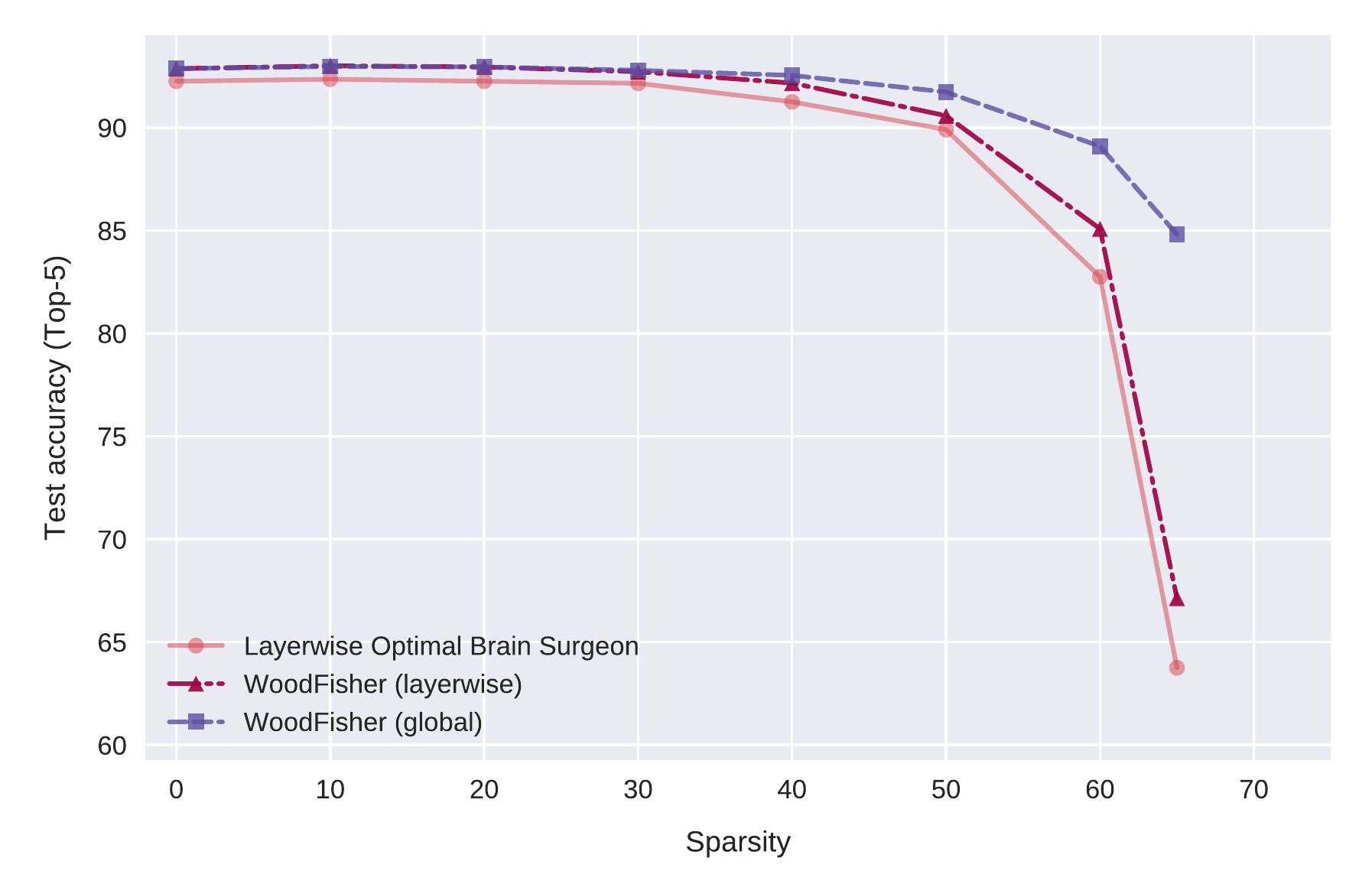}
	\caption{}
	\label{fig:one-shot-lobs}
\end{subfigure}
\begin{subfigure}{0.48\textwidth}
	\centering
	\vspace{-3mm}
	\includegraphics[width=0.95\linewidth]{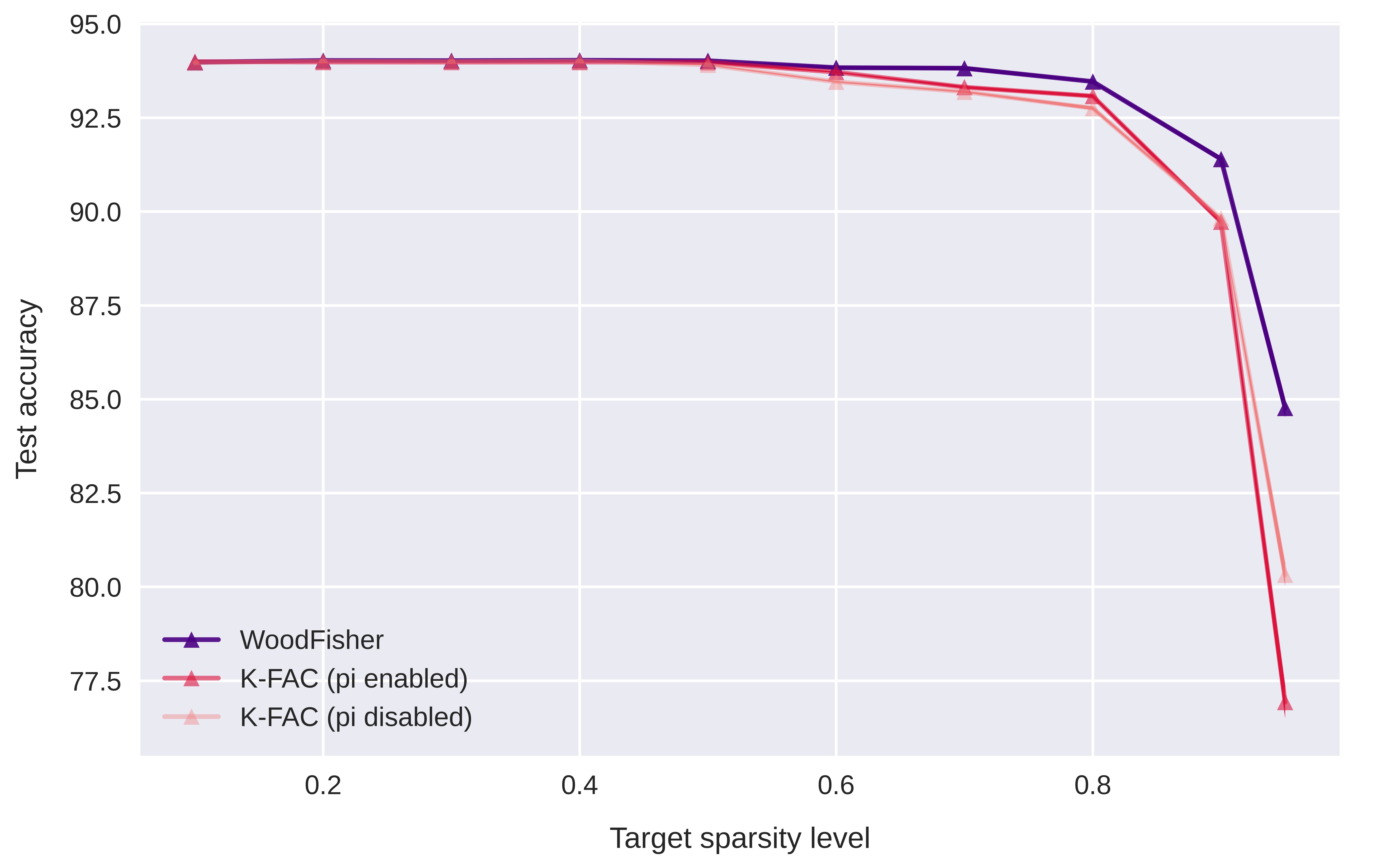}
	\caption{}
	\label{fig:kfac_comparison}
	\vspace{-1mm}
\end{subfigure}
	\caption{\footnotesize{\textbf{(a)} Top-5 test accuracy comparison of\textbf{ L-OBS and WoodFisher} on \textsc{ImageNet} for \textsc{ResNet-50}. \textbf{(b)} \textbf{WoodFisher vs K-FAC:} one-shot sparsity results for \textsc{MLPNet} on \textsc{MNIST}. K-FAC results are shown for both when the dampening $\pi$ is taken into account and otherwise.}}
\end{figure}

\paragraph{Comparison against L-OBS.} 

To facilitate a consistent comparison with L-OBS~\cite{dong2017learning}, we consider one-shot pruning of \textsc{ResNet-50} on \textsc{ImageNet}, and evaluate the performance in terms of top-5 accuracy as reported by the authors. (Besides this figure, all other results for test-accuracies are top-1 accuracies.) Here, all the layers are pruned to equal amounts, and so we first compare it with WoodFisher independent (layerwise). 
Further, in comparison to L-OBS, our approach also allows to automatically adjust the sparsity distributions. Thus, we also report the performance of WoodFisher joint (global)
The resulting plot is illustrated in Figure~\ref{fig:one-shot-lobs}, where we find that both independent and joint WoodFisher outperform L-OBS at all sparsity levels, and yield improvements of up to $\sim 3.5\%$ and $ 20\%$ respectively in test accuracy over L-OBS at the same sparsity level of $65\%$. 

\paragraph{Comparison against K-FAC based pruner.}\label{sec:kfac_comparison}

To quantitatively compare the approximation quality of the inverses (or IHVPs) yielded by K-FAC and WoodFisher, we consider the scenario of one-shot pruning of \textsc{MLPNet} on \textsc{MNIST}. For both, we utilize a block-wise estimate of the Fisher (with respect to the layers, i.e., no further chunking).

Figure~\ref{fig:kfac_comparison} illustrates these results for the `joint' pruning mode (however, similar results can observed in the `independent' mode as well). The number of samples used for estimating the inverse is the same across K-FAC and WoodFisher (i.e., $50,000$ samples)\footnote{For the sake of efficiency, in the case of WoodFisher, we utilize $1000$ averaged gradients over a mini-batch size of $50$. But, even if no mini-batching is considered and say, we considered $1000$ samples for both, we notice similar gains over K-FAC.}. This highlights the better approximation quality provided by WoodFisher, which unlike K-FAC does not make major assumptions. Note that, for convolutional layers, K-FAC needs to make additional approximations, so we can expect WoodFisher results to further improve over K-FAC.

\subsection{Gradual Pruning}\label{sec:gradual}

\paragraph{Setup.} So far, the two best methods we identified in one-shot tests are WoodFisher (joint/global) and global magnitude pruning. 
We now compare these methods extensively against several previous methods for unstructured pruning. 
To facilitate consistent comparison, we  demonstrate our results on the pre-trained \textsc{ResNet-50} and \textsc{MobileNetV1} models of the STR method~\cite{Kusupati2020SoftTW}, which claims state-of-the-art results. As in~\cite{Kusupati2020SoftTW}, all our \textsc{ImageNet} experiments are run for 100 epochs on 4 NVIDIA V100 GPUs (i.e., $\sim 2.5$ days for \textsc{ResNet-50} and $\sim 1$ day for \textsc{MobileNetV1}).  In terms of the pruning schedule, we follow the polynomial scheme of \cite{zhu2017prune} (see illustration in Figures~\ref{fig:gradual-mobilenet} and ~\ref{fig:gradual-imagenet}), and run WoodFisher and global magnitude in identical settings. 

Note, all the sparsity percentages are with respect to the weights of all the layers present, as none of the methods prune batch-norm parameters. Also, see Appendix~\ref{sec:app_exp_details} for further details on the pruning schedule, hyperparameters, and runtime costs for WoodFisher.

\begin{table}[t!] \centering\ra{1.1}
	\centering
	
	\resizebox{0.75\textwidth}{!}{
		\begin{tabular}{@{}lccccc@{}}
			\toprule
			
			\multirow{1}{*}{} & \multicolumn{2}{c}{Top-1 accuracy (\%)} &Relative Drop  & \multicolumn{1}{c}{Sparsity} & \multirow{1}{*}{Remaining }  \\
			\cmidrule(l{3pt}r{3pt}){2-3}
			\multirow{1}{*}{Method} & Dense {\small($D$)}  & Pruned  {\small($P$)}& {\small${100 \times \frac{ (P-D)}{D}}$} &   (\%) & \# of params\\
			\midrule
			DSR \citep{mostafa2019parameter} & 74.90 & 71.60  & -4.41 & 80.00 & 5.10 M\\
			Incremental \citep{zhu2017prune} & 75.95 & 74.25 &  -2.24 & 73.50 & 6.79 M \\
			DPF \citep{Lin2020Dynamic} & 75.95 & 75.13  & -1.08 & 79.90 & 5.15 M \\
			GMP + LS \citep{gale2019state} & 76.69 & 75.58  & -1.44 & 79.90 & 5.15 M  \\
			Variational Dropout \citep{molchanov2017variational} & 76.69 & 75.28  & -1.83  & 80.00 & 5.12 M  \\
			RIGL + ERK \citep{Evci2019RiggingTL} & 76.80 & 75.10 & -2.21 & 80.00 & 5.12 M \\
			SNFS + LS \citep{dettmers2019sparse}  & 77.00 & 74.90  & -2.73 &  80.00 &  5.12 M\\
			STR \citep{Kusupati2020SoftTW} & 77.01 & 76.19 & -1.06 & 79.55 & 5.22 M \\
			Global Magnitude & 77.01 & 76.59 & -0.55 & 80.00 & 5.12 M \\
			DNW \citep{wortsman2019discovering} & 77.50 & 76.20  & -1.67 & 80.00 & 5.12 M\\
			\textbf{\WF{WoodFisher}} & 77.01 & \WF{\textbf{76.76}} &  \WF{\textbf{-0.32}} & 80.00 & 5.12 M \\			
			\midrule
			
			GMP + LS \citep{gale2019state} & 76.69 & 73.91  & -3.62 & 90.00 & 2.56 M  \\
			Variational Dropout \citep{molchanov2017variational} & 76.69 & 73.84 & -3.72 & 90.27 & 2.49 M   \\
			RIGL + ERK \citep{Evci2019RiggingTL} & 76.80 & 73.00 & -4.94 & 90.00 & 2.56 M \\
			SNFS + LS \citep{dettmers2019sparse}  & 77.00 & 72.90  & -5.32 &  90.00 &  2.56 M\\
			STR \citep{Kusupati2020SoftTW} & 77.01 & 74.31 & -3.51 & 90.23 & 2.49 M \\
			Global Magnitude & 77.01 & 75.15 & -2.42 & 90.00 & 2.56 M \\
			DNW \citep{wortsman2019discovering} & 77.50 & 74.00  & -4.52   & 90.00 & 2.56 M\\
			\textbf{\WF{WoodFisher}} & 77.01 & \WF{\textbf{75.21}} & \WF{\textbf{-2.34}}  & 90.00 & 2.56 M \\			
			\midrule
			
			GMP  \citep{gale2019state} & 76.69 & 70.59  & -7.95 & 95.00 & 1.28 M \\
			Variational Dropout \citep{molchanov2017variational} & 76.69 & 69.41 & -9.49  & 94.92 & 1.30 M  \\
			Variational Dropout \citep{molchanov2017variational} & 76.69 & 71.81 & -6.36 & 94.94 & 1.30 M  \\
			RIGL + ERK \citep{Evci2019RiggingTL} & 76.80 & 70.00 & -8.85 & 95.00 & 1.28 M \\
			DNW \citep{wortsman2019discovering} & 77.01 & 68.30  & -11.31 & 95.00 & 1.28 M \\
			STR \citep{Kusupati2020SoftTW} & 77.01 & 70.97 & -7.84  & 94.80 & 1.33 M \\
			STR \citep{Kusupati2020SoftTW} & 77.01 & 70.40 & -8.58  & 95.03 & 1.27 M \\
			Global Magnitude & 77.01 &71.72 & -6.87  & 95.00 & 1.28 M \\
			\textbf{\WF{WoodFisher}} & 77.01 & \WF{\textbf{72.12}} & \WF{\textbf{-6.35}}  & 95.00 & 1.28 M \\
			\midrule
			
			GMP + LS \citep{gale2019state} & 76.69 & 57.90  &-24.50  &98.00  & 0.51 M \\
			Variational Dropout \citep{molchanov2017variational} & 76.69 & 64.52 & -15.87 &98.57 & 0.36 M   \\
			DNW \citep{wortsman2019discovering} & 77.01 & 58.20  & -24.42 & 98.00 & 0.51 M \\
			STR \citep{Kusupati2020SoftTW} & 77.01 & 61.46 & -20.19 & 98.05 & 0.50 M \\
			STR \citep{Kusupati2020SoftTW} & 77.01 & 62.84 & -18.40 & 97.78 & 0.57 M \\
			Global Magnitude & 77.01 & 64.28 & -16.53 & 98.00 & 0.51 M \\
			\textbf{\WF{WoodFisher}} & 77.01 & \WF{\textbf{65.55}} & \WF{\textbf{-14.88}} & 98.00 & 0.51 M \\
			
			\bottomrule
		\end{tabular}
	}\vspace{-1mm}
	\caption{Comparison of WoodFisher gradual pruning results with the state-of-the-art approaches. WoodFisher and Global Magnitude \textbf{results are averaged over two runs}. For their best scores, please refer to Table~\ref{tab:imagenet_gradual_sota_best}. LS denotes label smoothing, and ERK denotes the Erd\H{o}s-Renyi Kernel.}\label{tab:imagenet_gradual_sota}
\end{table}

\paragraph{\textsc{ResNet-50}.} Table~\ref{tab:imagenet_gradual_sota} presents our results with comparisons against numerous baselines for pruning \textsc{ResNet-50} at the sparsity levels of $80\%$, $90\%$, $95\%$ and $98\%$. To take into account that some prior work uses different dense baselines, we also report the relative drop. WoodFisher outperforms all baselines, across both gradual and dynamic pruning approaches, in every sparsity regime. Compared to STR~\citep{Kusupati2020SoftTW}, WoodFisher improves accuracy at all sparsity levels, with a Top-1 test accuracy gain of $\sim 1\%$, $1.5\%$, and $3.3\%$  respectively at the $90\%$, $95\%$, and $98\%$ sparsity levels. 

We also find that global magnitude (GM) is quite effective, surpassing many recent dynamic pruning methods, that can adjust the sparsity distribution across layers \citep{Evci2019RiggingTL,Lin2020Dynamic,Kusupati2020SoftTW}. Comparing GM and WoodFisher, the latter outperforms at all sparsity levels, with larger gain at higher sparsities, e.g.,  $\sim1.3\%$ boost in accuracy at $98\%$ sparsity. 

WoodFisher also outperforms Variational Dropout (VD) \citep{molchanov2017variational}, the top-performing regularization-based method, on all sparsity targets, with a margin of $\sim 1.5\%$ at $80\%$ and $90\%$ sparsity. VD is also known to be quite sensitive to initialization and hyperparameters~\citep{gale2019state}. This can be partly seen from its results in the $95\%$ regime, where a slight change in hyperparameters amounts to $0.02\%$ difference in sparsity and affects the obtained accuracy by over $2\%$.

Besides, WoodFisher significantly outperforms gradual magnitude pruning (GMP) \citep{gale2019state}, in all the sparsity regimes. Specifically, this amounts to a gain in test accuracy of $\sim 1 - 2\%$ at sparsity levels of $80\%$, $90\%$, and $95\%$, with a gain of $\ge 7\%$ at higher sparsities such as $98\%$. In fact, many of these GMP \citep{gale2019state} results employ label smoothing (LS) which we do not need, but this should only improve WoodFisher's performance further.

Note, here the authors also report the results when GMP is run for an extended duration ($\sim 2 \times$ longer compared to other methods). However, to be fair we do not compare them with other methods presented in Table~\ref{tab:imagenet_gradual_sota}, as followed in prior work~\citep{Kusupati2020SoftTW}. Nevertheless, even under such an extended pruning scenario, the final test accuracy of their models is often less than that obtained via WoodFisher, where we additionally pruned the first convolutional layer and our whole procedure took about half the number of epochs. Lastly, WoodFisher does not require industry-scale extensive hyperparameter tuning, unlike~\citep{gale2019state}.

\paragraph{\textsc{Additional comparisons}} 

\paragraph{(a) With DPF.} Besides the comparison in Table~\ref{tab:imagenet_gradual_sota}, we further compare against another recent state-of-the-art DPF~\cite{Lin2020Dynamic} in a more commensurate setting by starting from a similarly trained baseline. We follow their protocol and prune all layers except the last: see Table~\ref{tab:dpf}. In this setting as well, WoodFisher significantly outperforms DPF and the related baselines. 

\begin{SCtable} [1][h] \centering\ra{1.1}
	\centering
	\resizebox{0.7\textwidth}{!}{
		\begin{tabular}{@{}lccccc@{}}
			\toprule
			\multirow{1}{*}{} & \multicolumn{2}{c}{Top-1 accuracy (\%)} & Relative Drop & \multicolumn{1}{c}{Sparsity} & \multirow{1}{*}{Remaining }  \\
			\cmidrule(l{3pt}r{3pt}){2-3}
			\multirow{1}{*}{Method} & Dense {\small($D$)} & Pruned {\small($P$)} &   {\small${100 \times \frac{ (P-D)}{D}}$} &   (\%) & \# of params\\
			\midrule
			Incremental  & 75.95 & 73.36 &  -3.41 & 82.60 &4.45 M \\
			SNFS & 75.95 & 72.65  & -4.34 &  82.00 & 4.59 M \\
			DPF & 75.95 & 74.55  & -1.84 & 82.60  & 4.45 M \\
			\textbf{\WF{WoodFisher}} & 75.98 & \WF{\textbf{75.20}} & \WF{\textbf{-1.03}} & 82.70 &  4.41 M\\
			\bottomrule
		\end{tabular}
	}
	\caption{\footnotesize{Comparison with state-of-the-art DPF~\citep{Lin2020Dynamic} in a more commensurate setting by starting from a similarly trained dense baseline. }The numbers for Incremental \& SNFS are taken from \cite{Lin2020Dynamic}.}\label{tab:dpf}
\end{SCtable}

\paragraph{(b) With GMP.}  In ~\citet{gale2019state} (GMP), the authors empirically find that by keeping the first convolutional layer dense, pruning the last fully-connected layer to $80\%$, and then pruning rest of the layers in the network equally to the desired amount, the performance of magnitude pruning is significantly improved and serves as a state-of-the-art. To show that there is further room for improvement even when such sparsity profiles are utilized, we consider WoodFisher and Global Magnitude for automatically adjusting the sparsity distribution for the intermediate layers while pruning them. Table~\ref{tab:wf-indep} shows the results for this setting where the intermediate layers are pruned to an overall sparsity of $90\%$.

\begin{table}[h!]\centering\ra{1.1}
	\centering
	
	\resizebox{0.75\textwidth}{!}{
		\begin{tabular}{@{}lccccc@{}}
			\toprule
			
			\multirow{1}{*}{} & \multicolumn{2}{c}{Top-1 accuracy (\%)} & Relative Drop  & \multicolumn{1}{c}{Sparsity} & \multirow{1}{*}{Remaining }  \\
			\cmidrule(l{3pt}r{3pt}){2-3}
			\multirow{1}{*}{Method} & Dense {\small($D$)} & Pruned  {\small($P$)} & {\small${100 \times \frac{ (P-D)}{D}}$} &   (\%) & \# of params\\
			\midrule
			GMP  & 77.01 & 75.13  & -2.45 & 89.1 & 2.79 M \\
			Global Magnitude & 77.01 & 75.45  & -2.03 & 89.1 & 2.79 M \\
			\textbf{\WF{WoodFisher}} & 77.01 & \WF{\textbf{75.64}} & \WF{\textbf{-1.78}} & 89.1 & 2.79 M \\
			\bottomrule
		\end{tabular}
	}
	\caption{\small Comparison of WoodFisher, Global Magnitude and Magnitude (GMP) pruning for \textsc{ResNet-50} on \textsc{ImageNet}. Namely, this involves skipping the (input) first convolutional layer, and pruning the last fully connected layer to $80\%$. The rest of the layers are pruned to an overall target of $90\%$. So, here GMP prunes all of them uniformly to $90\%$, while WoodFisher and Global Magnitude use their obtained sparsity distributions for intermediate layers at the same overall target of $90\%$.}\label{tab:wf-indep}
	
\end{table}

\paragraph{\textsc{MobileNetV1}.}  MobileNets~\citep{Howard2017MobileNetsEC} are a class of parameter-efficient networks designed for mobile applications, and so is commonly used as a test bed to ascertain generalizability of unstructured pruning methods. In particular, we consider the gradual pruning setting as before on \textsc{MobileNetV1} which has $\sim 4.2 M$ parameters. Following STR~\citep{Kusupati2020SoftTW}, we measure the performance on two sparsity levels: $75\%$ and $90\%$  and utilize their pre-trained dense model for fair comparisons. 

\begin{table}[h!]\centering\ra{1.1}
	\centering
	
	\resizebox{0.75\textwidth}{!}{
		\begin{tabular}{@{}lcccccc@{}}
			\toprule
			
			\multirow{1}{*}{} & \multicolumn{2}{c}{Top-1 accuracy (\%)} & Relative Drop  & \multicolumn{1}{c}{Sparsity}  & \multirow{1}{*}{Remaining }  \\
			\cmidrule(l{3pt}r{3pt}){2-3}
			\multirow{1}{*}{Method} & Dense {\small($D$)} & Pruned  {\small($P$)} & {\small${100 \times \frac{ (P-D)}{D}}$} &   (\%) & \# of params\\
			\midrule
			Incremental \citep{zhu2017prune} & 70.60 & 67.70 & -4.11 & 74.11$^\alpha$ &1.09 M \\
			STR \citep{Kusupati2020SoftTW} & 72.00 & 68.35  & -5.07 & 75.28 & 1.04 M \\
			Global Magnitude & 72.00 & 69.90  & -2.92 & 75.28 & 1.04 M \\
			\textbf{\WF{WoodFisher}} & 72.00 & \WF{\textbf{70.09}} & \WF{\textbf{-2.65}}& 75.28 &  1.04 M \\
			\midrule
			Incremental \citep{zhu2017prune} & 70.60 & 61.80 & -12.46 & 89.03$^\alpha$ & 0.46 M \\
			STR \citep{Kusupati2020SoftTW} & 72.00 & 62.10  & -13.75  & 89.01 & 0.46 M \\
			Global Magnitude & 72.00 & 63.02 & -12.47 & 89.00 & 0.46 M \\
			\textbf{\WF{WoodFisher}} & 72.00 & \WF{\textbf{63.87}} & \WF{\textbf{-11.29}} & 89.00  &
			0.46 M\\
			\bottomrule
		\end{tabular}
	}
	\caption{\footnotesize{Comparison of WoodFisher gradual pruning results for \textbf{MobileNetV1 on ImageNet} in $75\%$ and $90\%$ sparsity regime.} $(^\alpha)$ next to Incremental~\citep{zhu2017prune} is to highlight that the first convolutional and all depthwise convolutional layers are kept dense, unlike the other shown methods. The obtained sparsity distribution and other details can be found in the section~\ref{sec:spar-mn50}.}\label{tab:mobilenet}
	
\end{table}

Table~\ref{tab:mobilenet} shows the results for WoodFisher and global magnitude along with the methods mentioned in STR. Note that the Incremental baseline from ~\citep{zhu2017prune} keeps the first convolutional and the (important) depthwise convolutional layers dense. However, in an aim to be network-agnostic, we let the global variant of WoodFisher to automatically adjust the sparsity distributions across the layers. Nevertheless, we observe that WoodFisher outperforms~\citep{zhu2017prune} as well as the other baselines: STR and global magnitude, in each of the sparsity regimes. 

Further, it can be argued that the results in Table~\ref{tab:mobilenet} are from just single runs. Hence, in order to check the statistical significance of the results, we run both Global Magnitude and WoodFisher for 5 times\footnote{Doing 5 runs for all experiments is feasible in the \textsc{MobileNetV1} setting in contrast to \textsc{ResNet50}, as the overall gradual pruning time is much shorter for the former.} (at each of the sparsity levels) and carry out a two-sided student's t-test. We observe that WoodFisher results in a statistically significant gain over Global Magnitude (as well as other baselines) for both the sparsity regimes. The results for this comparison can be found in Table~\ref{tab:mobilenet-test}.

\begin{table}[h!]\centering\ra{1.2}
	\centering
	
	\resizebox{0.9\textwidth}{!}{
		\begin{tabular}{@{}lccccc|cc@{}}
			\toprule
			
			\multirow{1}{*}{} & \multicolumn{2}{c}{Top-1 accuracy (\%)} & Relative Drop  & \multicolumn{1}{c}{Sparsity}  & \multirow{1}{*}{Remaining } & \multicolumn{1}{c}{p-value} & \multicolumn{1}{c}{Significant}\\
			\cmidrule(l{3pt}r{3pt}){2-3}
			\multirow{1}{*}{Method} & Dense {\small($D$)} & Pruned  {\small($P$)} & {\small${100 \times \frac{ (P-D)}{D}}$} &   (\%) & \multicolumn{1}{c}{\# of params} & \multicolumn{1}{|c}{} & (at $\alpha=0.05$)\\
			\midrule
			Global Magnitude & 72.00 & 69.93 $\pm$ 0.07  & -2.88 & 75.28 & 1.04 M & \multirow{2}{*}{0.02979} & \multirow{2}{*}{\cmark}\\
			\textbf{\WF{WoodFisher}} & 72.00 & \WF{\textbf{70.04 $\pm$ 0.06}}  & \WF{\textbf{-2.72}}& 75.28 &  1.04 M \\
			\midrule
			Global Magnitude & 72.00 & 63.08 $\pm$ 0.12 & -12.39 & 89.00 & 0.46 M & \multirow{2}{*}{0.00003} & \multirow{2}{*}{\cmark}\\
			\textbf{\WF{WoodFisher}} & 72.00 & \WF{\textbf{63.69 $\pm$ 0.11}} & \WF{\textbf{-11.54}} & 89.00  &
			0.46 M\\
			\bottomrule
		\end{tabular}
	}
	\caption{\footnotesize{WoodFisher and Global Magnitude gradual pruning results, reported with \textbf{mean and standard deviation across 5 runs}, for \textbf{MobileNetV1 on ImageNet} in $75\%$ and $90\%$ sparsity regime. We also run a two-sided student's t-test to check if WoodFisher significantly outperforms Global Magnitude, which we find to be true at a significance level of $\alpha=0.05$.}}\label{tab:mobilenet-test}
	
\end{table}

\paragraph{\textsc{Summary}.} To sum up, results show that WoodFisher outperforms state-of-the-art approaches, from each class of pruning methods, in all the considered sparsity regimes, for both \textsc{ResNet-50} and \textsc{MobileNetV1} setting a new state-of-the-art in unstructured pruning for these common benchmarks.

\subsection{What goes on during gradual pruning?} 
\begin{figure}[h!]
	\centering
	\begin{subfigure}{0.49\textwidth}
		\centering 
		\includegraphics[width=\linewidth]{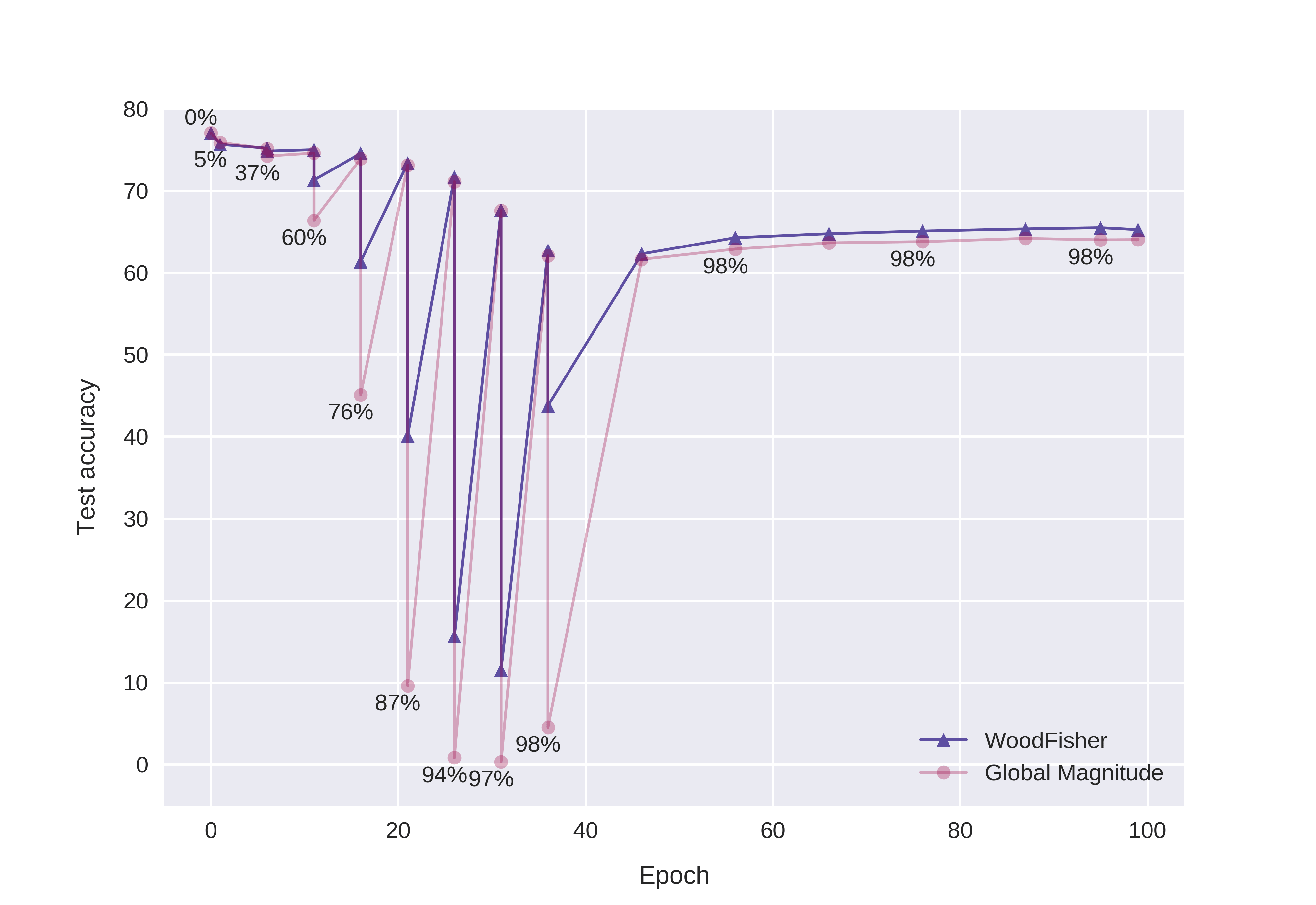}
				\caption{\textsc{ResNet-50}, \textsc{ImageNet}}
		\label{fig:gradual-imagenet}
	\end{subfigure}
	\begin{subfigure}{0.49\textwidth}
		\centering
		\includegraphics[width=\linewidth]{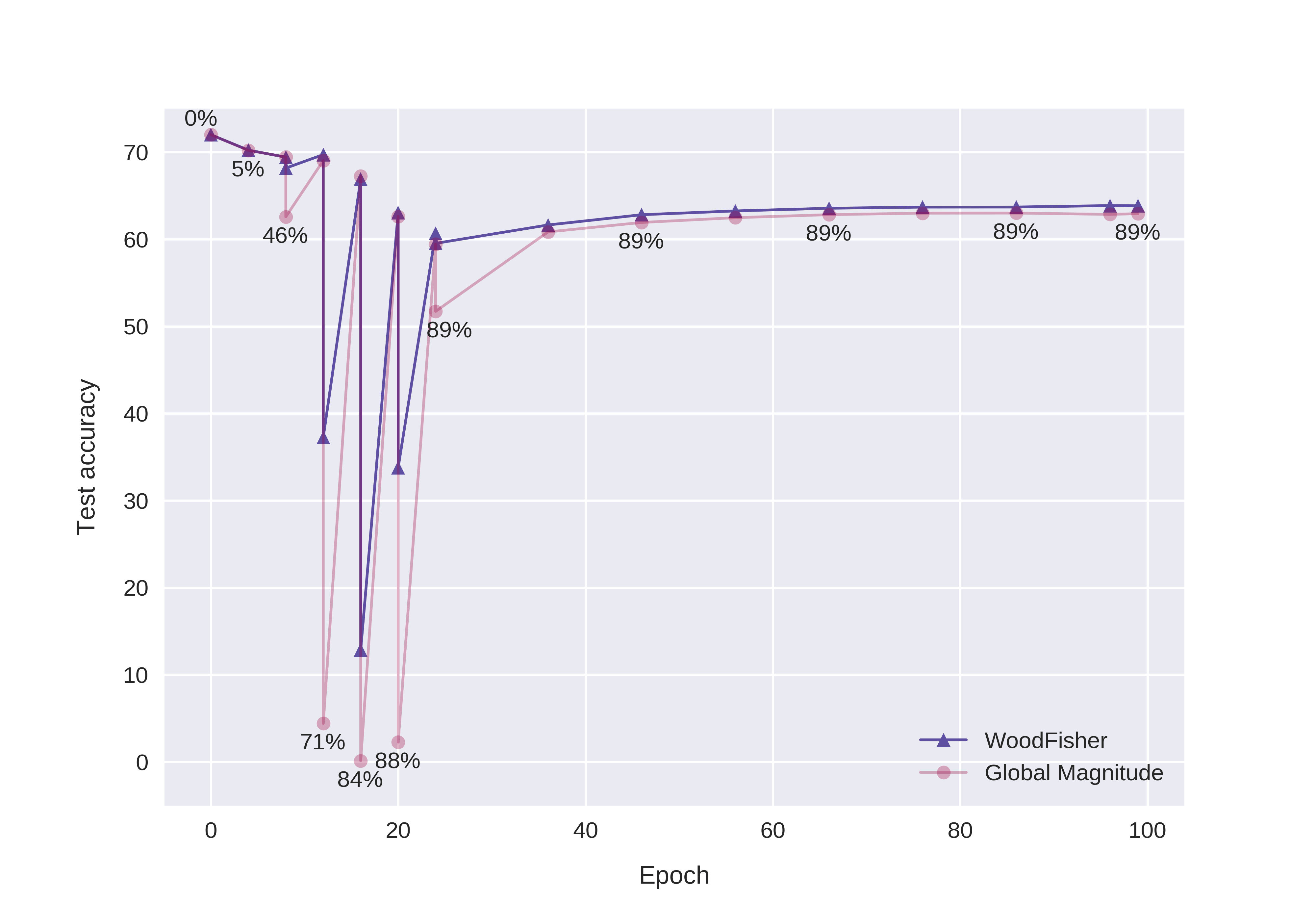}
		\caption{\textsc{MobileNetV1}, \textsc{ImageNet}}
		\label{fig:gradual-mobilenet}
	\end{subfigure}
	\caption{\small The course of gradual pruning with points annotated by the corresponding sparsity amounts.}
\end{figure}

Next, to give some further insights into these results, we illustrate, in Figures~\ref{fig:gradual-mobilenet} and \ref{fig:gradual-imagenet}, how WoodFisher and global magnitude behave during the course of gradual pruning, for \textsc{ResNet-50} and \textsc{MobileNetV1} respectively. Here, we see the rationale behind the improved performance of WoodFisher. After almost 
every pruning step, WoodFisher provides a better pruning direction, and even with substantial retraining in between and after, global magnitude fails to catch up in terms of accuracy. The trend holds for other sparsity levels as well, and these figures can be found in the Apendix~\ref{sec:app_gradual_detail}.  This shows the benefit of using the second order information via WoodFisher to perform superior pruning steps.

Magnitude pruning, which prunes all layers equally, performs even worse, and Figure~\ref{fig:gradual-mobilenet-all} in the Apendix~\ref{sec:app_gradual_detail} showcases the comparison between pruning steps for all the three: WoodFisher, Global Magnitude, and Magnitude.

\subsection{Actual inference speed-up} Further, it is interesting to consider the actual speed-up which can be obtained via these methods, as the theoretical FLOP counts might not reveal the complete picture. And, to test the speed-up at inference, we use the inference framework of~\cite{NM},  which supports efficient execution of unstructured sparse convolutional models on CPUs. 
We execute their framework on an Amazon EC2 c4.8xlarge instance with an 18-core Intel Haswell CPU. Sparse models are exported and executed through a modified version of the ONNX Runtime~\cite{ONNX}. Experiments are averaged over 10 runs, and have low variance. 
The full results are given in Table~\ref{tab:inference}.

\begin{table}[h] \centering\ra{1.1}
	\centering
	\resizebox{0.65\textwidth}{!}{
		\begin{tabular}{@{}lcccc@{}}
			\toprule
			
			\multirow{1}{*}{} & \multicolumn{2}{c}{Inference Time (ms)} & Top-1 Acc. \\
			\cmidrule(l{3pt}r{3pt}){2-3}
			\multirow{1}{*}{Compression} & Batch 1 & Batch 64 \\
			\midrule
			Dense  & 7.1 & 296 &  77.01\%  \\ 			\midrule
			STR-81.27\% & 5.6 & 156  & 76.12\% \\
			\textbf{\WF{WF-Joint-80\%}} & 6.3 & 188  & 76.73\% \\
			\midrule
			STR-90.23\% & 3.8 & 144  & 74.31\% \\
			\textbf{\WF{WF-Independent-89.1\%}} & 4.3 & 157  & 75.23\% \\
			\textbf{\WF{WF-Joint-90\%}} & 5.0 & 151  & 75.26\% \\
			\bottomrule
		\end{tabular}
	}
	
	\caption{\footnotesize{Comparison of inference times at batch sizes 1 and 64 for various sparse models, executed on the framework of~\citep{NM}, on an 18-core Haswell CPU. The table also contains the Top-1 Accuracy for the model on the ILSVRC validation set}.}\label{tab:inference}
\end{table}

We briefly summarize the results as follows. 
First, note that all methods obtain speed-up relative to the dense baseline, with speed-up being correlated with increase in sparsity. 
At the same time, the standard WoodFisher variants (WF) tend to have higher inference time in comparison to STR, but also a higher accuracy for the same sparsity budget. 
A good comparison point is at $\sim$90\% global sparsity, where WF-Joint has 0.95\% higher Top-1 accuracy, for a 1.2ms difference in inference time at batch size 1. 
However, here the WF-Independent-89.1\% model\footnote{This refers to the WoodFisher version, where following \cite{gale2019state}, the first convolutional layer is not pruned and the last fully-connected layer is pruned to $80\%$, while the rest of layers are pruned to $90\%$. We note that this latter model has lower overall sparsity than WF-Joint, since it does not prune the first and last layers, following the recipe from~\cite{gale2019state}. } offers a better trade-off, with similar accuracy improvement, but only 0.5ms difference in inference time with respect to STR. Further, in Section~\ref{sec:app_flops}, we also consider a variant that can optimize for FLOPs as well. 

\section{Extensions}\label{sec:extensions}

\subsection{WoodFisher for structured pruning}\label{sec:structured}
Let's start by reformulating the pruning statistic in Eq.~\eqref{eq:prun_stat} for removing a single parameter as follows,
\begin{equation}\label{eq:prun_stat-equiv}
\rho_q  = \frac{w_q^2}{2 \, \lbrack\hessinv\rbrack_{qq}} = \frac{(\w^\top \\e_q)^{\top} (\w^\top \,\e_q)}{2 \,  \e_q^\top \hessinv \e_q} = \frac{\e_q^{\top} \, \w \w^\top \, \e_q}{2 \,  \e_q^\top \hessinv \e_q}.
\end{equation}

Then to remove a parameter group/block (given by the set $Q$), we can either just consider the sum of pruning statistics over the parameters in $Q$, i.e., 
\begin{equation}\label{eq:prun_stat-struct1}
\rho_Q  = \sum_{q \in Q} \frac{\e_q^{\top} \, \w \w^\top \, \e_q}{2 \,  \e_q^\top \hessinv \e_q}.
\end{equation}

Or, we can take into account the correlation, which might be helpful for the case of structured pruning (like when removing filters) via the formulation given below:
\begin{equation}\label{eq:prun_stat-struct2}
\rho_Q  = \frac{\tile_Q^{\top}\, \w \w^\top  \,\tile_Q}{2 \,  {\tile_Q}^\top \hessinv \tile_Q},
\end{equation}

where, $\tile_Q = \1\lbrace q \in Q\rbrace$, i.e., it is the one-hot vector denoting which parameters are present in the set $Q$. 
Depending on the choice of the pruning statistic, we can modify the optimal perturbation $\delw^\ast$ accordingly.

More rigorously, the above formulation can be obtained by relaxing the problem of removing multiple parameters in Eq.~\eqref{eq:prob_w_q_q} via summing the individual constraints into one combined constraint. 
\begin{equation}\label{eq:prob_w_Q}
\min_{Q} \;\;\; \bigg\lbrace \min _{\delw \, \in  \,\mathbb{R}^d} \;\; \left(\frac{1}{2} \delw^{\top} \, \hess \, \delw\right), \quad \text { s.t. } \quad \tile_Q ^\top \delw + \w^\top \tile_Q = 0 \; \bigg\rbrace.
\end{equation}

Then using the procedure of Lagrange multipliers as done earlier, we obtain the pruning statistic as in Eq.~\eqref{eq:prun_stat-struct2}, and the optimal weight perturbation is given by the following, 
\begin{equation}\label{eq:opt_delw_struct}
\delw^\ast  = \frac{- (\w^\top \tile_Q) \hessinv \tile_Q}{ \tile_Q^\top \hessinv \tile_Q}.
\end{equation}

Lastly, we leave a detailed empirical investigation of structured pruning for future work.

\subsection{WoodTaylor: Pruning at a general point}\label{sec:woodtaylor}

\paragraph{Pruning analysis.} Incorporating the first-order gradient term in the Optimal Brain Damage framework should result in a more faithful estimate of the pruning direction, as many times in practice, the gradient is not exactly zero. Or it might be that pruning is being performed during training like in dynamic pruning methods. Hence, we redo the analysis by accounting for the gradient (see Appendix \ref{sec:app_woodtaylor}) and we refer to this resulting method as \textit{`WoodTaylor'.}

Essentially, this modifies the problem in Eq.~\eqref{eq:prob_w} to as follows:
\begin{equation}\label{eq:grad_prob_w}
\min _{\delw \, \in  \,\mathbb{R}^d} \;\; \left( {\gradw}^{\top} \delta \mathbf{w}+ \frac{1}{2} \delw^{\top} \, \hess \, \delw\right), \quad \text { s.t. } \quad \e_q ^\top \delw + w_q = 0.
\end{equation}

The corresponding Lagrangian can be then written as:
\begin{equation}\label{eq:grad_lagr}
\lagr(\delw, \lambda) = {\gradw}^{\top} \delta \mathbf{w}+ \frac{1}{2} \delw^{\top} \, \hess \, \delw + \lambda \left( \e_q ^\top \delw + w_q \right). 
\end{equation}

Taking the derivative of which with respect to $\delw$ yields,
\begin{equation}\label{eq:lagr_grad_derivative}
\gradw + \hess \delw + \lambda \e_q = 0 \implies \delw = - \lambda {\hessinv} e_q - \hessinv \gradw.
\end{equation}

The Lagrange dual function $g(\lambda)$ can be then computed by putting the above value for $\delw$ in the Lagrangian in Eq.~\eqref{eq:grad_lagr} as follows:
\begin{align}\label{eq:grad_lagr_dual}
g(\lambda)\;\; &= \;\;   -\lambda \gradw^{\top} \hessinv \e_q - \gradw^\top \hessinv \gradw \notag\\
&+ \; \frac{1}{2} {\big(\lambda \hessinv \e_q + \hessinv \gradw\big)}^\top \hess \; \big(\lambda \hessinv \e_q + \hessinv \gradw\big) \notag\\
&+ \;  \lambda \, \big(-\lambda \e_q^\top \hessinv \e_q \, - \, \e_q^\top \hessinv \gradw + w_q \big) \notag\\
& = - \frac{\lambda^2}{2} \e_q^\top \hessinv \e_q - \lambda \e_q^\top \hessinv \gradw + \lambda w_q - \frac{1}{2} \gradw^\top \hessinv \gradw                                          \, .
\end{align}                                                                                                     

As before, maximizing with respect to $\lambda$, we obtain that the optimal value $\lambda^\ast$ of this Lagrange multiplier as follows:
\begin{equation}\label{eq:grad_opt_mult}                                                                                
\lambda^\ast  = \frac{w_q \, - \, \e_q^\top \hessinv \gradw}{\e_q^\top \hessinv \e_q} = \frac{w_q \, - \, \e_q^\top \hessinv \gradw}{\lbrack\hessinv\rbrack_{qq}}.                     
\end{equation}

Note, if the gradient was 0, then we would recover the same $\lambda^\ast$ as in Eq.~\eqref{eq:opt_mult}. Next, the corresponding optimal perturbation, ${\delw}^\ast$, so obtained is as follows:
\begin{equation}\label{eq:grad_opt_delw}
\delw^\ast  = \frac{-\,\big(w_q \, - \, \e_q^\top \hessinv \gradw \,\big)\, \hessinv \e_q}{ \lbrack\hessinv\rbrack_{qq}} - \hessinv \gradw \,.
\end{equation}

In the end, the resulting change in loss corresponding to the optimal perturbation that removes parameter $w_q$ can be written as (after some calculations\footnote{It's easier to put the optimal value of $\lambda^\ast$ in the dual function (Eq.~\eqref{eq:grad_lagr_dual}) and use duality, than substituting the optimal perturbation $\delw^\ast$ in the primal objective.}),
\begin{equation}\label{eq:grad_opt_loss}
\dell^\ast  = \frac{w_q^2}{2 \, \lbrack\hessinv\rbrack_{qq}} + \frac{1}{2} \frac{{\big(\e_q^\top \hessinv \gradw\big)}^2}{\hessqq} - w_q \frac{\e_q^\top \hessinv \gradw}{\hessqq} -\frac{1}{2} \gradw^\top \hessinv \gradw\,.
\end{equation}

Lastly, choosing the best parameter $\w_q$ to be removed, corresponds to one which leads to the minimum value of the above change in loss. As in Section~\ref{sec:zerograd}, our pruning statistic $\rho$ in this setting can be similarly defined, in addition by excluding the last term in the above Eq.~\eqref{eq:grad_opt_loss} since it does not involved the choice of removed parameter $q$. This is indicated in the Eq.~\eqref{eq:grad_prun_stat} below.

\begin{equation}\label{eq:grad_prun_stat}
\boxed{\rho_q  = \frac{w_q^2}{2 \, \lbrack\hessinv\rbrack_{qq}} + \frac{1}{2} \frac{{\big(\e_q^\top \hessinv \gradw\big)}^2}{\hessqq} - w_q \frac{\e_q^\top \hessinv \gradw}{\hessqq}.}
\end{equation}

\paragraph{Results.} We present some initial results for the case when the model is far from the optimum, and hence the gradient is not close to zero. This setting will allow us to clearly see the effect of incorporating the first-order gradient term, considered in the WoodTaylor analysis. In particular, we consider an \mlp $\,$ on \mnist, which has only been trained for 2 epochs. 

\begin{figure}[h]
	\centering
	\includegraphics[width=0.6\linewidth]{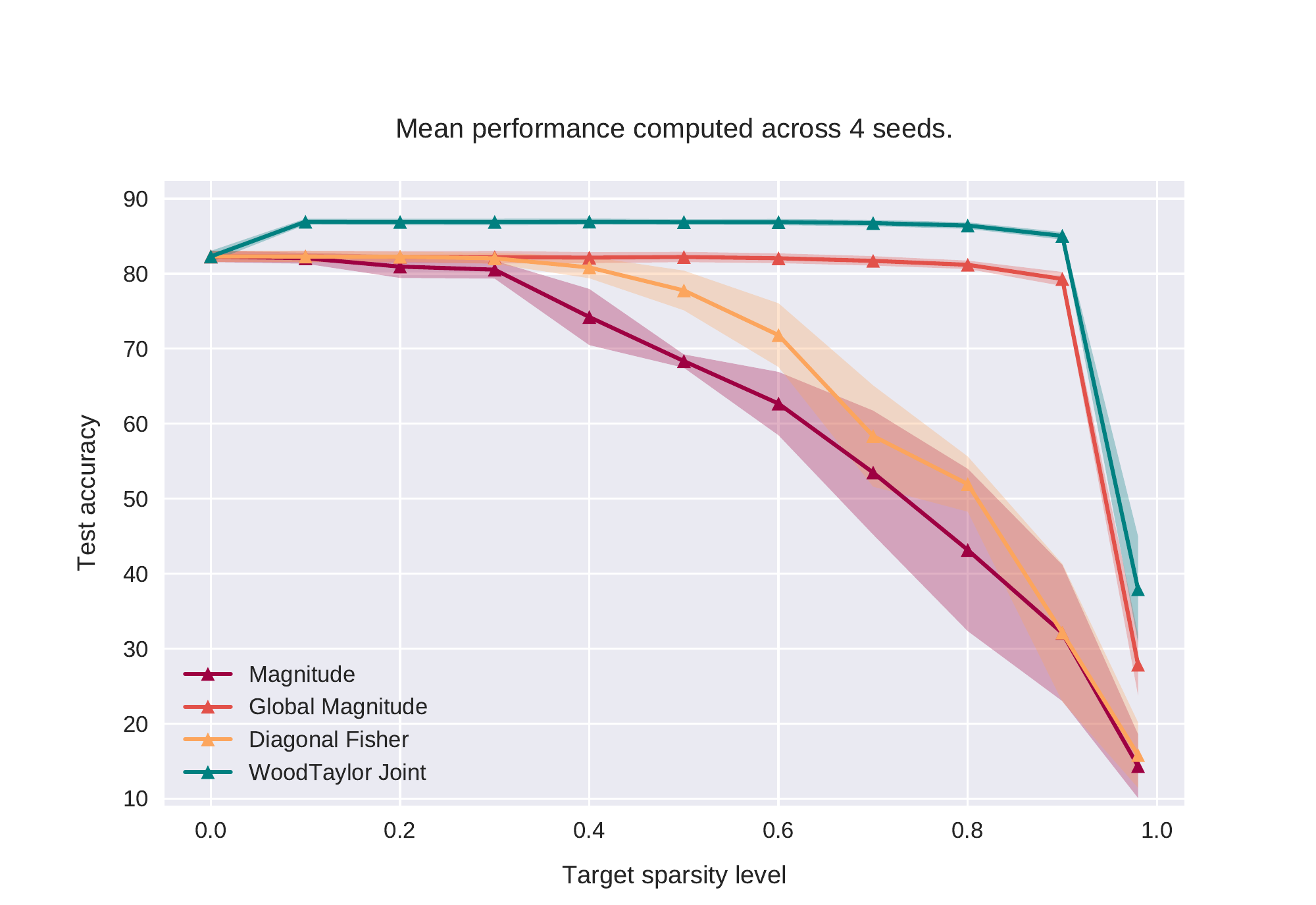}
	\caption{Comparing one-shot sparsity results for WoodTaylor and baselines on the partially trained \mlp $\,$ on \mnist. }
	\label{fig:one-shot-woodtaylor-mnist}
\end{figure}

Figure~\ref{fig:one-shot-woodtaylor-mnist} presents the results for performing one-shot compression for various sparsity levels at this stage in the training. Similar to the results in past, we find that WoodTaylor is significantly better than magnitude or diagonal Fisher based pruning as well as the global version of magnitude pruning. But the more interesting aspect is the improvement brought about by WoodTaylor, which in fact also improves over the accuracy of the initial dense model by about $5\%$, up to sparsity levels of $\sim90\%$. Here, the number of samples used for Fisher in both WoodTaylor and Diagonal Fisher is 8000. A dampening of $1e-1$ was used for WoodTaylor, and Figures~\ref{fig:ablation-woodtaylor-simplified},\ref{fig:ablation-woodtaylor} of the Appendix~\ref{sec:app_woodtaylor_pre} present relevant ablation studies. 

Further, even on pre-trained models such as \textsc{ResNet-20} on \textsc{CIFAR-10}, we find that WoodTaylor can also outperform WoodFisher, since the gradient is usually not exactly zero (c.f. Appendix ~\ref{sec:app_woodtaylor_pre}).

\paragraph{Applicability with dynamic methods.} Note, an advantage of dynamic pruning methods is that the pruning is performed during the training itself, although we have seen in Table~\ref{tab:imagenet_gradual_sota}, better results are obtained when pruning is performed post-training. Current dynamic pruning methods like DPF \citep{Lin2020Dynamic} prune via global magnitude, and a possible future work would be to use WoodTaylor instead.

\subsection{Sampled Fisher: The case of unlabelled data}\label{sec:sampled_fisher}

While empirical Fisher inherently uses the label information when computing gradients, it is possible to avoid that and instead use a single sample from the model distribution, thus making it applicable to unlabeled data. (Appendix ~\ref{sec:sampled_fisher} shows this does not impact the results much). 

\begin{table}[h]\centering\ra{1.1}
	\centering
	
	\resizebox{0.6\textwidth}{!}{
		\begin{tabular}{@{}lccc@{}}
			\toprule
			
			\multirow{1}{*}{} & \multicolumn{3}{c}{Top-1 accuracy (\%)} \\
			\cmidrule(l{3pt}r{3pt}){2-4}
			\multirow{1}{*}{Sparsity} & Dense & empirical WoodFisher & sampled WoodFisher \\
			\midrule
			$20\%$  & \multirow{4}{*}{76.13} & 76.10 $\pm$ 0.04  & 76.16 $\pm$ 0.02\\
			$40\%$  &  & 75.22 $\pm$ 0.07 & 75.31 $\pm$ 0.05 \\
			$60\%$  & & 69.21 $\pm$ 0.05 & 69.29 $\pm$0.20\\
			$70\%$  & & 48.35 $\pm$ 0.22 & 48.74 $\pm$1.03\\
			$80\%$  & & 5.48 $\pm$ 0.45 & 5.77 $\pm$ 0.34\\
			\bottomrule
		\end{tabular}
	}\vspace{-1mm}
	\caption{\footnotesize{Comparison of \textbf{one-shot pruning performance} of WoodFisher, when the considered Fisher matrix is empirical Fisher or one-sample approximation to true Fisher, for \textsc{ResNet-50} on \textsc{ImageNet}. The results are averaged over three seeds.}}\label{tab:sample_fisher}
	
\end{table}

The `empirical' WoodFisher denotes the usual WoodFisher used throughout the paper. All the presented results in the paper are based on this setting. Whereas, the `sampled WoodFisher' refers to sampling the output from the model's conditional distribution instead of using the labels, in order to compute the gradients. As a result, the latter can be utilized in an unsupervised setting. 

Note that, ideally we would want to use the true Fisher, and none of the above approximations to it. But, computing the true Fisher would require $m$ additional back-propagation steps for each sampled $\vy$ or $k$ more steps to compute the Jacobian. This would make the whole procedure $m \times$ or $k \times$ more expensive. Hence, the common choices are to either use $m=1$ samples as employed in K-FAC~\cite{martens2015optimizing} or simply switch to empirical Fisher (which we choose). The presented results are based on the following hyperparameters: 
chunk size $= 1000$, fisher subsample size $=240$, fisher mini batch-size = $100$.

In Table~\ref{tab:sample_fisher}, we contrast the performance of these two options: `empirical' and `sampled' WoodFisher, when performing one-shot pruning\footnote{As in the one-shot experiments, we use the \textsc{ResNet-50} model from \textsc{torchvision} as the dense baseline.} of \textsc{ResNet-50} on \textsc{ImageNet} in the \textit{joint} mode of WoodFisher.  We find that results for both types of approximations to Fisher (or whether we use labels or not) are in the same ballpark. The sampled WoodFisher does, however, have slightly higher variance which is expected since it is based on taking one sample from the model's distribution. Nevertheless, it implies that we can safely switch to this sampled WoodFisher when labels are not present.

\section{Discussion}
\paragraph{Future Work.} Some of the many interesting directions to apply WoodFisher, include, e.g., structured pruning which can be facilitated by the OBD framework (as in e.g., Section~\ref{sec:structured}), compressing popular models used in NLP like Transformers~\citep{vaswani2017attention}, providing efficient IHVP estimates for influence functions~\citep{koh2017understanding}, etc.

\paragraph{Conclusion.} In sum, our work revisits the theoretical underpinnings of neural network pruning, and shows that foundational work can be successfully extended to large-scale settings, yielding state-of-the-art results. 
We hope that our findings can provide further momentum to the investigation of second-order properties of neural networks, and be extended to applications beyond compression.  
  
\subsection*{Acknowledgements}
This project has received funding from the European Research Council (ERC) under the European Union's Horizon 2020 research and innovation programme (grant agreement No 805223 ScaleML). Also, we would like to thank Alexander Shevchenko, Alexandra Peste, and other members of the group for fruitful discussions.

\begin{small}
\bibliographystyle{unsrtnat}
\bibliography{references}
\end{small}

%

\onecolumn
\clearpage
\newpage{}
\appendix
\appendixpage

\renewcommand{\thesection}{S\arabic{section}}
\renewcommand{\thetable}{S\arabic{table}}
\renewcommand{\thefigure}{S\arabic{figure}}
\renewcommand{\thefootnote}{S\arabic{footnote}}
\setcounter{figure}{0}
\setcounter{table}{0}
\setcounter{footnote}{0}

\setlength\cftparskip{2pt}
\setlength\cftbeforesecskip{2pt}
\setlength\cftaftertoctitleskip{3pt}
\addtocontents{toc}{\protect\setcounter{tocdepth}{2}}
\setcounter{tocdepth}{1}

\tableofcontents

\clearpage

\hypersetup{linkcolor=red}
\section{Experimental Details}\label{sec:app_exp_details}

\subsection{Pruning schedule}
We use Stochastic Gradient Descent (SGD) as an optimizer during gradual pruning, with a learning rate $=0.005$, momentum $= 0.9$, and weight decay $=0.0001$. We run the overall procedure for 100 epochs, similar to other works~\citep{Kusupati2020SoftTW}.Retraining happens during this procedure, i.e., in between the pruning steps and afterwards, as commonly done when starting from a pre-trained model. 

For \textsc{ResNet-50}, we carry out pruning steps from epoch $1$, at an interval of $5$ epochs, until epoch $40$\footnote{All the epoch numbers are based on zero indexing.}. Once the pruning steps are over, we decay the learning rate in an exponential schedule from epoch $40$ to $90$ by a factor of $0.6$ every $6$ epochs.

For \textsc{MobileNetV1}, we carry out pruning steps from epoch $4$, at an interval of $4$ epochs, until epoch $24$. Once the pruning steps are over, we decay the learning rate in an exponential schedule from epoch $30$ to $100$ by a factor of $0.92$ every epoch. 

The amount by which to prune at every step is calculated based on the polynomial schedule, suggested in~\citep{zhu2017prune}, with the initial sparsity percentage set to $0.05$.

The same pruning schedule is followed for all: WoodFisher, Global Magnitude, and Magnitude. In fact, this whole gradual pruning procedure was originally made for magnitude pruning, and we simply replaced the pruner by WoodFisher.

\subsection{WoodFisher Hyperparameters}

The hyperparameters for WoodFisher are summarized in the Table \ref{tbl:hyper_params}. Fisher subsample size refers to the number of outer products considered for empirical Fisher. Fisher mini-batch size means the number of samples over which the gradients are averaged, before taking an outer product for Fisher. This was motivated from computation reasons, as this allows us to see a much larger number of data samples, at a significantly reduced cost. 

The chunk size refers to size of diagonal blocks based on which the Hessian (and its inverse) are approximated. For \textsc{ResNet-50}, we typically use a chunk size of $2000$, while for \textsc{MobileNetV1} we use larger chunk size of $10,000$ since the total number of parameters is less in the latter case (allowing us to utilize a larger chunk size).

A thorough ablation study is carried out with respect to these hyperparameters for WoodFisher in Section~\ref{sec:app_one_shot_detail}.

\begin{table}[h!]\centering\ra{1.2}
	\small
	\caption{Detailed hyperparameters for the gradual pruning results presented in Tables~\ref{tab:imagenet_gradual_sota}, \ref{tab:mobilenet}. }\label{tbl:hyper_params}
	\vspace{1em}
	
	\resizebox{0.9\linewidth}{!}{		\begin{tabular}[h]{lccccc}
			\toprule
			
			Model & Sparsity ($\%$) & \multicolumn{2}{c}{Fisher} & Chunk size & Batch Size \\
			\cmidrule{3-4}
			& & subsample size &mini-batch size  &  \\\midrule
			
			\multirow{4}{*}{\textsc{ResNet-50}} & 80.00 & 400   & 400 & 2000  & 256 \\
			& 90.00 & 400   & 400 & 2000  & 180 \\
			& 95.00 & 400   & 400 & 2000  & 180 \\
			& 98.00 & 400   & 400 & 1000  & 180 \\\midrule
			\multirow{2}{*}{\textsc{MobileNetV1}}	 & 75.28 & 400   & 2400 & 10,000  & 256 \\
			& 89.00 & 400   & 2400 & 10,000  & 180 \\
			\bottomrule
	\end{tabular}}
	
\end{table}

Note, we used a batch size of 256 or 180, depending upon whether the GPUs we were running on had 16GB memory or less. Anyhow, the same batch size was used at all the respective sparsity levels for Global Magnitude to ensure consistent comparisons. 

Besides, the dampening $\lambda$ used in WoodFisher, to make the empirical Fisher positive definite, is set to $1\text{e}-5$ in all the experiments.

\subsection{Run time costs for WoodFisher pruning steps}
WoodFisher indeed incurs more time during each pruning step. However, the additional time taken for these pruning steps (which are also limited in number, $\sim 6$ to $8$ in our experiments) pales in comparison to the overall 100 epochs on \textsc{ImageNet} in the gradual pruning procedure. 

The exact time taken in each pruning step, depends upon the value of the fisher parameters and chunk size. So more concretely, for \textsc{ResNet-50}  this time can vary as follows e.g., $\sim 15$ minutes for fisher subsample size $=80$, fisher subsample size $=400$, chunk size $=1000$ to $\sim 47$ minutes for fisher subsample size $=160$, fisher subsample size $=800$, chunk size $=2000$. However, as noted in Tables~\ref{tab:effect_resnet},~\ref{tab:effect_mobilenet}, the gains from increasing these hyperparameter values is relatively small (after a threshold), so one can simply trade-off the accuracy in lieu of time, or vice versa. 

Most importantly, one has to keep in mind that compressing a particular neural network will only be done once. The \textbf{pruning cost will be amortized} over its numerous runs in the future. As a result, the extra test accuracy provided via WoodFisher is worthwhile the slightly increased running cost.

Lastly, note that, currently in our implementation we compute the inverse of the Hessian sequentially across all the blocks (or chunks). But, this \textit{computation is amenable to parallelization} and a further speed-up can easily obtained for use in future work.

\section{More on the related work for IHVPs}\label{app:ihvp}

\subsection{K-FAC }\label{sec:kfac-problems}

In the recent years, an approximation called K-FAC \citep{Heskes2000OnNL, martens2015optimizing} has been made for the Fisher that results in a more efficient application when used as a pre-conditioner or for IHVPs. Consider we have a fully-connected network with $l$ layers.  If we denote the pre-activations of a layer $i$ by $\vs_i$, we can write them as $\vs_i = W_i \va_{i-1}$, where $W_i$ is the weight matrix at the $i^\text{th}$ layer and $a_{i-1}$ denotes the activations from the previous layer (which the $i^\text{th}$ layer receives as input). 

By chain rule, the gradient of the objective $L$ with respect to the weights in layer $i$, is the following:  $\nabla_{W_{i}} L = \text{vec} (\,\vg_i \va_{i-1}^\top\,)$. Here, $\vg_i$ is the gradient of the objective with respect to the pre-activations $s_i$ of this layer, so $\vg_i = \nabla_{s_{i}} L$. Using the fact that $\text{vec}(\vu \vv^\top) = \vv \otimes \vu$, where $\otimes$ denotes the Kronecker product, we can simplify our expression of the gradient with respect to $W_i$ as $\nabla_{W_{i}} L = \va_{i-1}^\top \otimes \vg_i $.

Then, we can then write the Fisher block corresponding to layer $i$ and $j$ as follows, 

\begin{equation}
\begin{aligned}
F_{i, j}=\mathrm{E}\left[\nabla_{W_{i}} L  \, \nabla_{W_{j}} L^{\top}\right]=\mathrm{E}\left[\left(\va_{i-1} \otimes \vg_{i}\right)\left(\va_{j-1} \otimes \vg_{j}\right)^{\top}\right] &\stackrel{(a)}{=}\mathrm{E}\left[\left(\va_{i-1} \otimes \vg_{i}\right)\left(\va_{j-1}^{\top} \otimes \vg_{j}^{\top}\right)\right] \\
&\stackrel{(b)}{=}\mathrm{E}\left[\va_{i-1} \va_{j-1}^{\top} \otimes \vg_{i} \vg_{j}^{\top}\right], 
\end{aligned}
\end{equation}

where, in (a) and (b) we have used the transpose and mixed-product properties of Kronecker product. The expectation is taken over the model's distribution as in the formulation of Fisher. 

The Kronecker Factorization (K-FAC) based approximation $\widetilde{F}$ thus used by the authors can be written as, 
\begin{equation}
\widetilde{F}_{i, j} = \mathrm{E}\left[\va_{i-1} \va_{j-1}^\top \right] \otimes \mathrm{E}\left[\vg_i \vg_{j}^\top\right] = \widetilde{A}_{i-1, j-1} \otimes \widetilde{G}_{i, j}
\end{equation}

Essentially, we have moved the expectation inside and do it prior to performing the Kronecker product. As mentioned by the authors, this is a major approximation since in general the expectation of a Kronecker product is not equal to the Kronecker product of the expectations. We will refer to $\widetilde{F}$ as the Fisher matrix underlying K-FAC or the K-FAC approximated Fisher.

The advantage of such an approximation is that it allows to compute the inverse of K-FAC approximated Fisher quite efficiently. This is because the inverse of a Kronecker product is equal to the Kronecker product of the inverses. This implies that instead of inverting one matrix of bigger size $n_{i-1} n_{i} \times n_{j-1} n_{j}$, we need to invert two smaller matrices $\widetilde{A}_{i, j}$ and $\widetilde{G}_{i, j}$ of sizes $n_{i-1} \times n_{j-1}$ and $n_{i} \times n_{j}$ respectively (here, we have denoted the number of neurons in layer $\ell$ by $n_\ell$).

As a result, K-FAC has found several applications in the last few years in: optimization \citep{ba2016distributed, Osawa_2019}, pruning \citep{zeng2019mlprune, wang2019eigendamage}, reinforcement-learning \citep{WuMGLB17}, etc. However, an aspect that has been ignored is the accuracy of this approximation, which we discuss in Section~\ref{sec:kfac_comparison} in the context of pruning. Besides, there are a couple more challenges associated with the Kronecker-factorization based approaches. 

\paragraph{Extending to different network types.}  Another issue with K-FAC is that it only naturally exists for fully-connected networks. When one proceeds to the case of convolutional or recurrent neural networks, the Kronecker structure needs to be specially designed by making further approximations \citep{grosse2016kroneckerfactored,martens2018kroneckerfactored}. Whereas, a WoodFisher based method would not suffer from such a problem. 

\paragraph{Application to larger networks.} Furthermore, when applied to the case of large neural networks like \textsc{ResNet-50}, often further approximations like the chunking of block size as we consider or channel-grouping as called by \cite{laurent2018an}, or assuming spatially uncorrelated activations are anyways required. 

Thus, in lieu of these aspects, we argue that WoodFisher, i.e., (empirical) Fisher used along with Woodbury-based inverse, is a better alternative (also see the quantitative comparison with K-FAC in Section~\ref{sec:one-shot-pruning}).

\subsection{Other methods}
\paragraph{Double back-propagation.} This forms the naive way of computing the entire Hessian matrix by explicitly computing each of its entries. However, such an approach is extremely slow and would require $\mathcal{O}(d^2)$ back-propagation steps, each of which has a runtime of $\mathcal{O}(md)$, where $m$ is the size of the mini-batch considered. Thus, this cubic time approach is out of the question. 

\paragraph{Diagonal Hessian.} If we assume the Hessian to be diagonal, this allows us to compute the inverse very easily by simply inverting the elements of the diagonal. But, even if we use the Pearlmutter's trick \citep{10.1162/neco.1994.6.1.147}, which lets us compute the exact Hessian-vector product in linear time, we need a total of $\mathcal{O}(d)$ such matrix-vector products to estimate the diagonal, which results in an overall quadratic time. 

\paragraph{Diagonal Fisher.} Diagonal estimate for the empirical Fisher is really efficient to build, since it just requires computing the average of the squared gradient across the training samples, for each dimension. If the mini-batch size is $m$, we just need $\mathcal{O}(md)$ time to build this estimate. This approach has been widely used in optimization by adaptive methods \citep{kingma2014adam,Duchi2010AdaptiveSM}, as well for model compression by the work called Fisher-pruning \citep{theis2018faster}. However as we show ahead, by simply paying a small additional factor of $c$ in the runtime, we can estimate the inverse and IHVPs more accurately. This leads to a performance which is significantly better than that obtained via diagonal Fisher.

\paragraph{Hessian-Free methods.}  Another line of work is to completely forgo the explicit computation of Hessians \citep{martens_free}, by posing the problem of computing IHVP with a vector $\vv$ as solving the linear system $\mathbf{H} \vx = \vv$ for $\vx$. Such methods rely on conjugate-gradients based linear-system  solvers that only require matrix-vector products, which for neural networks can be obtained via Pearlmutter's trick \citep{10.1162/neco.1994.6.1.147}. However, a big disadvantage of these methods is that they can require a lot of iterations to converge since the underlying Hessian matrix is typically very ill-conditioned. Further, this whole procedure would have to be  repeated at least $\mathcal{O}(d)$ times to build just the diagonal of the inverse, which is the minimum needed for application in model compression. 

\paragraph{Neumann series expansion.} 
These kind of methods \citep{krishnan2018neumann,agarwal2016secondorder} essentially exploit the following result in Eqn.~\eqref{eq:neumann} for matrices $A$ which have an eigen-spectrum bounded between 0 and 1, i.e., $0 < \lambda(A) < 1$. 
\begin{equation}\label{eq:neumann}
A^{-1}=\sum_{i=0}^{\infty}(I-A)^{i}
\end{equation}
This can be then utilized to build a recurrence of the following form, 
\begin{equation}
A_{n}^{-1} \triangleq I+(I-A) A_{n-1}^{-1}, 
\end{equation}
which allows us to efficiently estimate (unbiased) IHVP's via sampling. However, an important issue here is the requirement of the eigen-spectrum to be between 0 and 1, which is not true by default for the Hessian.  This implies that we further need to estimate the largest absolute eigenvalue (to scale) and the smallest negative eigenvalue (to shift). Hence, requiring the use of the Power method which adds to the cost. Further, the Power method might not be able to return the smallest negative eigenvalue at all, since when applied to the Hessian (or its inverse) it would yield the eigenvalue with the largest magnitude (or smallest magnitude).

\paragraph{Woodbury-based methods.} In prior work, Woodbury-based inverse has been considered for the case of a one-hidden layer neural network in Optimal Brain Surgeon (OBS, \cite{NIPS1992_647}), where the analytical expression of the Hessian can be written as an outer product of gradients. An extension of this approach to deeper networks, called L-OBS, was proposed in \cite{dong2017learning}, by defining separate layer-wise objectives, and was applied to carefully-crafted blocks at the level of neurons. Our approach via empirical Fisher is more general, and we show ahead experimentally that it yields better approximations at scale (Figure~\ref{fig:one-shot-lobs}). 

To facilitate a consistent comparison with L-OBS, we consider one-shot pruning of \textsc{ResNet-50} on \textsc{ImageNet}, and evaluate the performance in terms of top-5 accuracy as reported by the authors. (Besides this figure, all other results for test-accuracies are top-1 accuracies.) Here, all the layers are pruned to equal amounts, and so we first compare it with WoodFisher independent (layerwise). 
Further, in comparison to L-OBS, our approach also allows to automatically adjust the sparsity distributions. Thus, we also report the performance of WoodFisher joint (global)
The resulting plot is illustrated in Figure~\ref{fig:one-shot-lobs}, where we find that both independent and joint WoodFisher outperform L-OBS at all sparsity levels, and yield improvements of up to $\sim 3.5\%$ and $ 20\%$ respectively in test accuracy over L-OBS at the same sparsity level of $65\%$. 

\begin{figure}[h]
	\centering
	\includegraphics[width=0.5\linewidth]{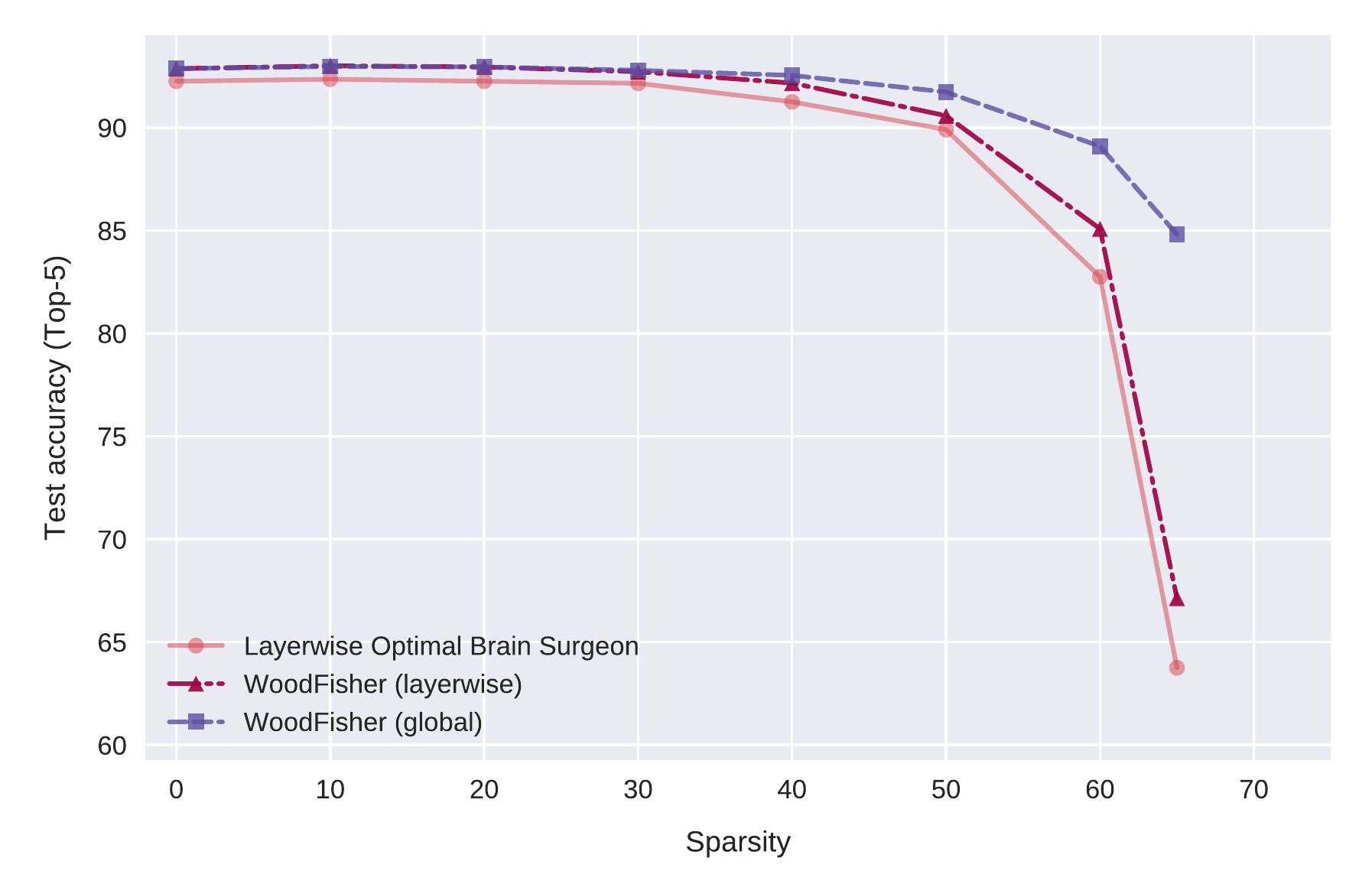}
	\caption{Top-5 test accuracy comparison of L-OBS and WoodFisher on \textsc{ImageNet} for \textsc{ResNet-50}.}
	\label{fig:one-shot-lobs}
\end{figure}

\clearpage
\section{Visual tour detailed}\label{app:hessian_pics_mnist}

\begin{figure}[h!]
	\centering
	
	\begin{subfigure}{0.9\textwidth}
		\centering
		\includegraphics[width=0.45\linewidth]{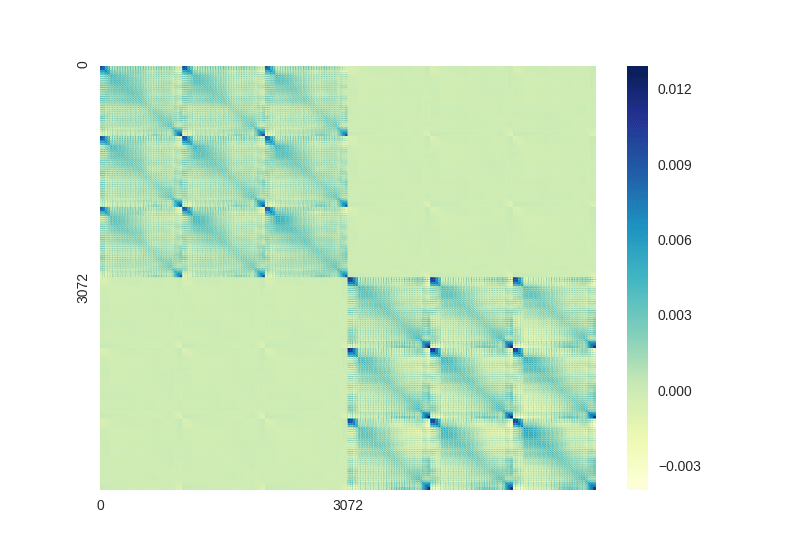}
		\includegraphics[width=0.45\linewidth]{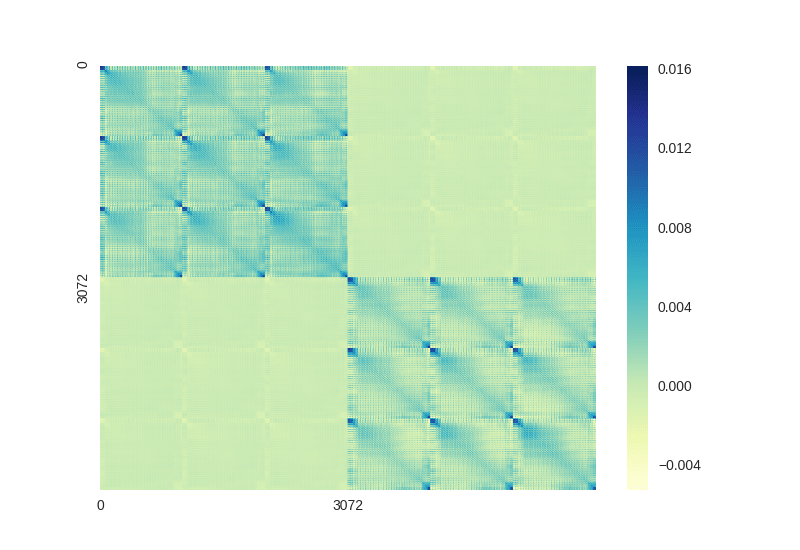}
		\caption{First-layer sub matrices averaged across over diagonal blocks of $6144 \times 6144$ for illustration purposes.}
		\label{fig:kfac_actual}
	\end{subfigure} 
	
	\begin{subfigure}{0.9\textwidth}
		\centering
		\includegraphics[width=0.45\linewidth]{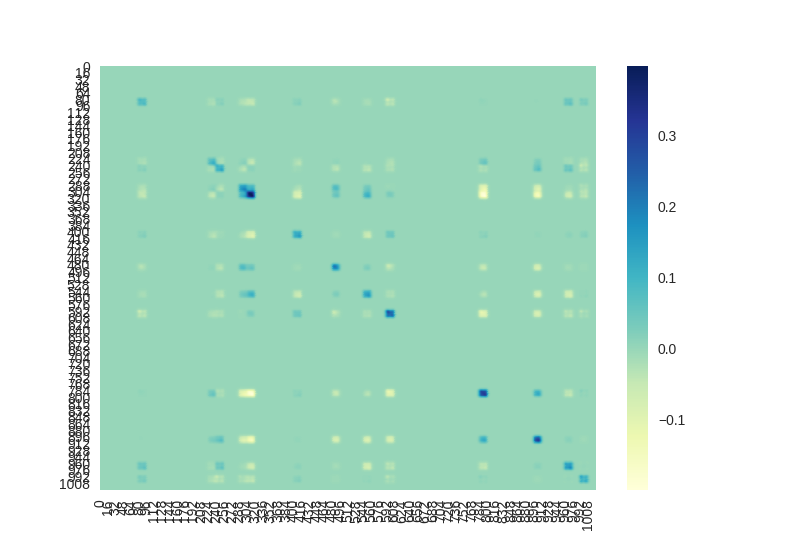}
		\includegraphics[width=0.45\linewidth]{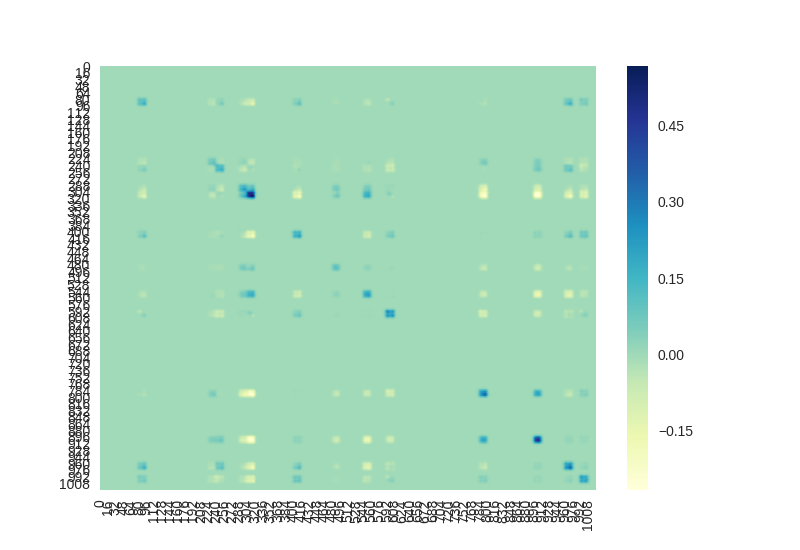}
		\caption{Second-layer sub-matrices.}
		\label{fig:kfac_actual}
	\end{subfigure} 
	
	\begin{subfigure}{0.9\textwidth}
		\centering
		\includegraphics[width=0.45\linewidth]{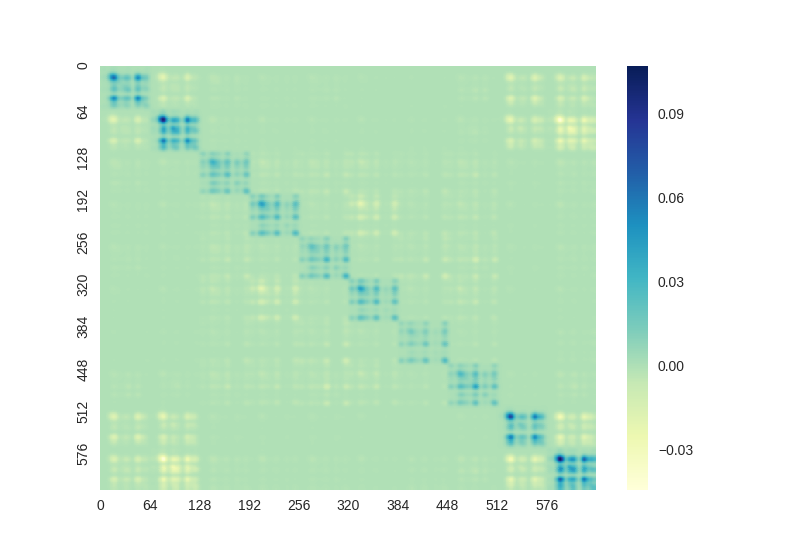}
		\includegraphics[width=0.45\linewidth]{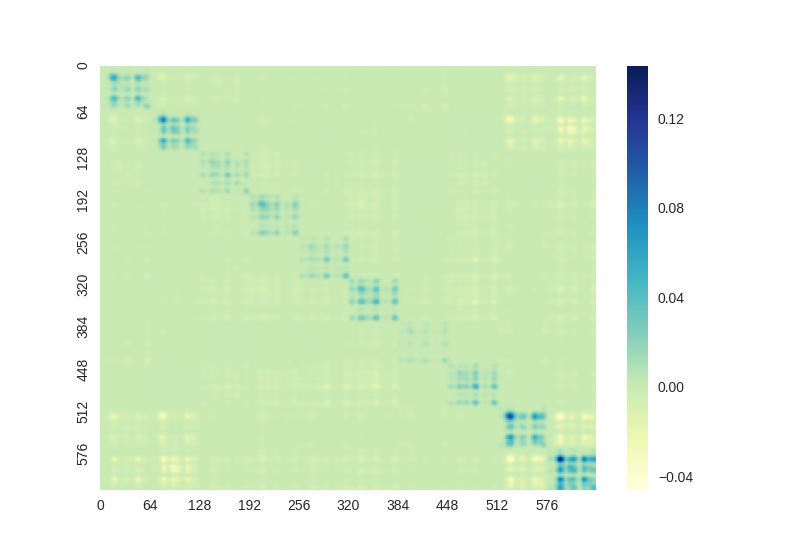}
		\caption{Third-layer sub matrices.}
		\label{fig:kfac_actual}
	\end{subfigure} 
	
	\caption{Hessian and empirical Fisher blocks for \textsc{CIFARNet} ($3072 \rightarrow 16 \rightarrow 64 \rightarrow10$) on the diagonal corresponding to different layers when trained on \cifar. Figures have been smoothened slightly with a Gaussian kernel for better visibility. Both Hessian and empirical Fisher have been estimated over a batch of $100$ examples in all the figures. Hessian blocks are in the left-column, while empirical Fisher blocks are displayed in the right-column. }
	\label{fig:cifar-figures-diag-all}
\end{figure}

\subsection{All figures for Hessian and empirical Fisher comparison on \textsc{CIFARNet}}\label{app:cifar_visual_all}
We consider a fully connected network, \textsc{CIFARNet}, with two hidden layers. Since, \textsc{CIFAR10} consists of $32 \times 32$ RGB images, so we adapt the size of network as follows: $3072 \rightarrow 16 \rightarrow 64 \rightarrow 10$. Such a size is chosen for computational reasons, as the full Hessian exactly is very expensive to compute. 

We follow a commonly used SGD-based optimization schedule for training this network on CIFAR10, with a learning rate $0.05$ which is decayed by a factor of 2 after every 30 epochs, momentum $0.9$, and train it for a total of 300 epochs. The checkpoint with best test accuracy is used as a final model, and this test accuracy\footnote{In fact, this \textsc{CIFARNet} model with $41.8 \%$ test accuracy, is far from ensuring that the model and data distribution match, yet the empirical Fisher is able to faithfully capture the structure of the Hessian.} is $41.8\%$. However, this low test accuracy is not a concern for us, as we are more interested in investigating the structures of the Hessian and the empirical Fisher matrices.

\begin{figure}[tb]
	\centering
		\begin{subfigure}{0.95\textwidth}
			\centering
			\includegraphics[width=0.45\linewidth]{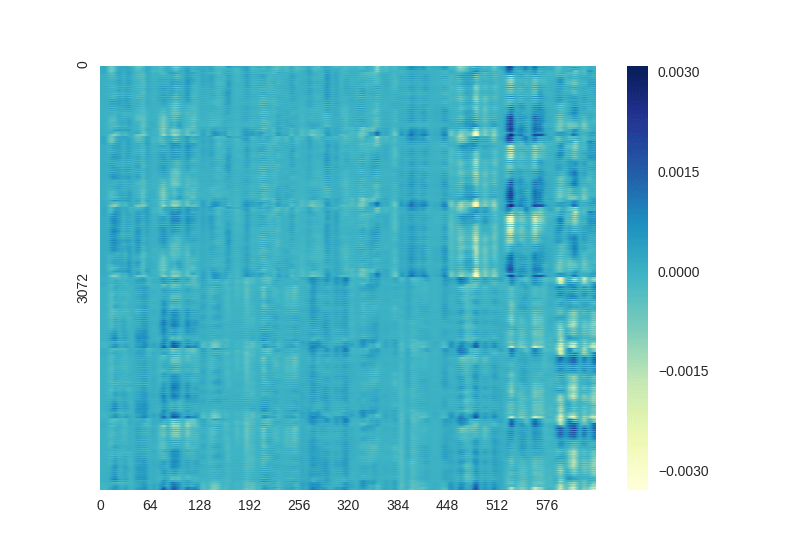}
			\includegraphics[width=0.45\linewidth]{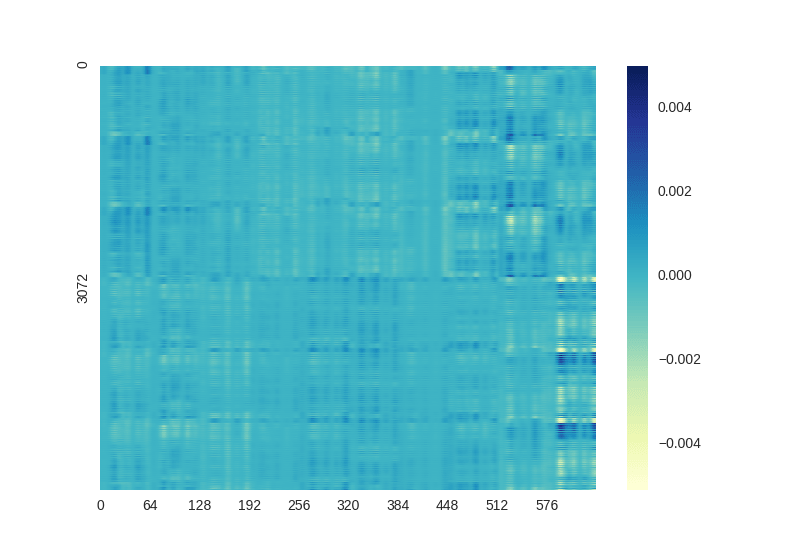}
			\caption{Cross-matrices between first-layer and third-layer averaged across over blocks of $6144 \times 640$ for illustration purposes. }
			\label{fig:kfac_actual}
		\end{subfigure} 
	
	\begin{subfigure}{0.95\textwidth}
		\centering
		\includegraphics[width=0.45\linewidth]{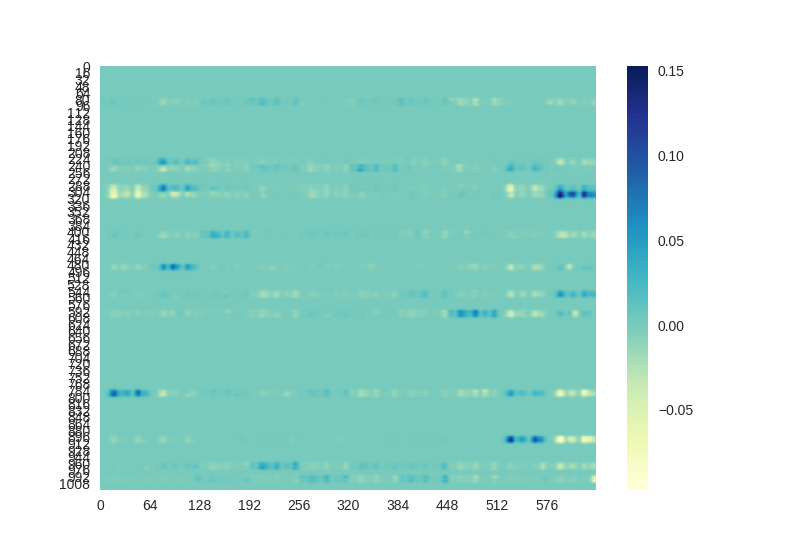}
		\includegraphics[width=0.45\linewidth]{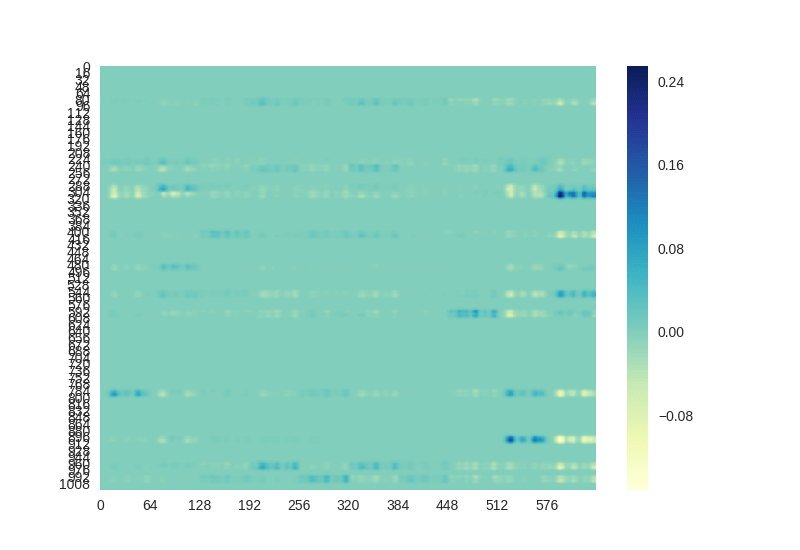}
		\caption{Cross-matrices between second-layer and third-layer. }
		\label{fig:kfac_actual}
	\end{subfigure} 

	\caption{\textbf{Off-Diagonal}  Hessian and empirical Fisher blocks for \textsc{CIFARNet} ($3072 \rightarrow 16 \rightarrow 64 \rightarrow10$) corresponding to different layers when trained on \cifar. Figures have been smoothened slightly with a Gaussian kernel for better visibility. Both Hessian and empirical Fisher have been estimated over a batch of $100$ examples in all the figures. Hessian blocks are in the left-column, while empirical Fisher blocks are displayed in the right-column. }
	\label{fig:cifar-figures-offdiag}
\end{figure}

The plots in the Figure~\ref{fig:cifar-figures-diag-all} illustrate the obtained matrices for the diagonal sub-matrices corresponding to the first, second, and the third layers. We observe that empirical Fisher possesses essentially the same structure as observed in the Hessian. Further, Figure~\ref{fig:cifar-figures-offdiag} presents the result for the off-diagonal or cross blocks of these two matrices, where also we find a similar trend. 

Thus, we conclude that the empirical Fisher shares the structure present in the Hessian matrix.

\subsection{Across different stages of training}\label{app:mnist_train_visual}

\begin{figure}[!h]
	\centering
	
	\begin{subfigure}{0.85\textwidth}
		\centering
		\includegraphics[width=\linewidth]{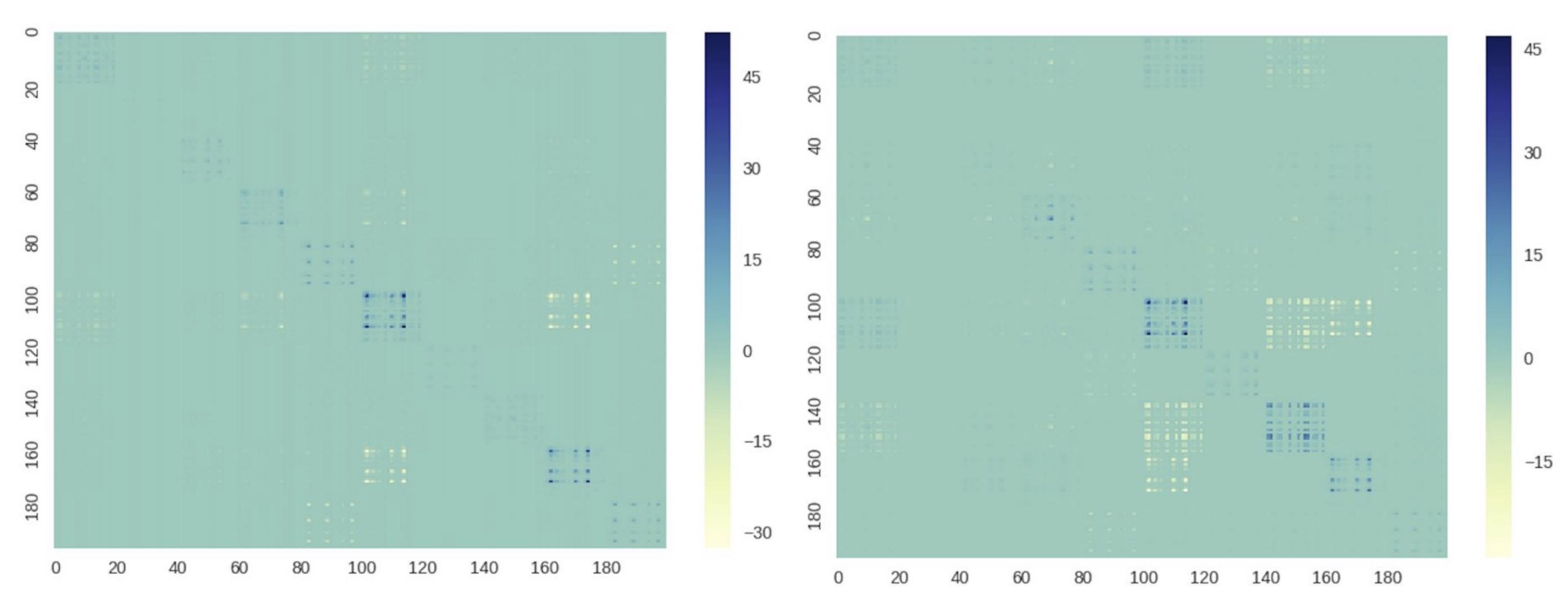}
		\caption{\textbf{At 0.5 epochs.} Test accuracy at this stage is $63.9\%$.}
		\label{fig:kfac_actual}
	\end{subfigure} 
	
	\begin{subfigure}{0.85\textwidth}
		\centering
		\includegraphics[width=\linewidth]{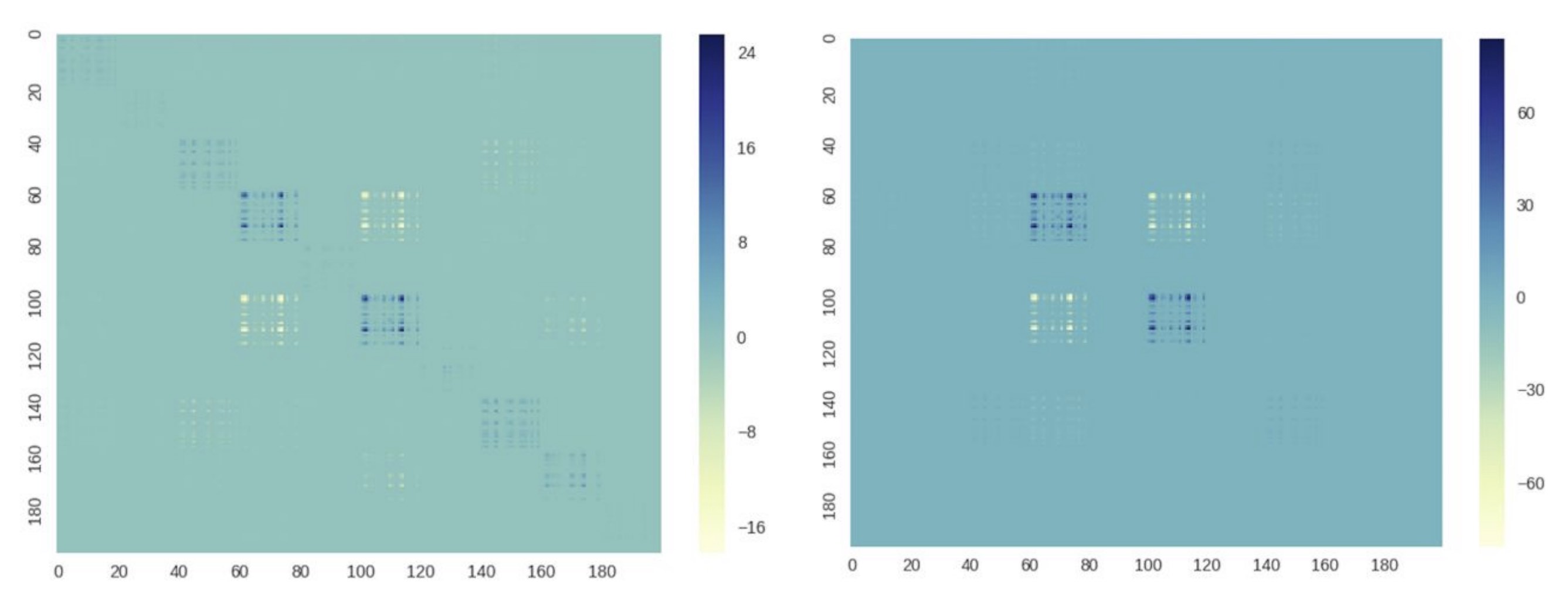}
		\caption{\textbf{At 5 epochs.} Test accuracy at this stage is $85.5\%$.}
		\label{fig:kfac_actual}
	\end{subfigure} 
	
	\begin{subfigure}{0.85\textwidth}
		\centering
		\includegraphics[width=\linewidth]{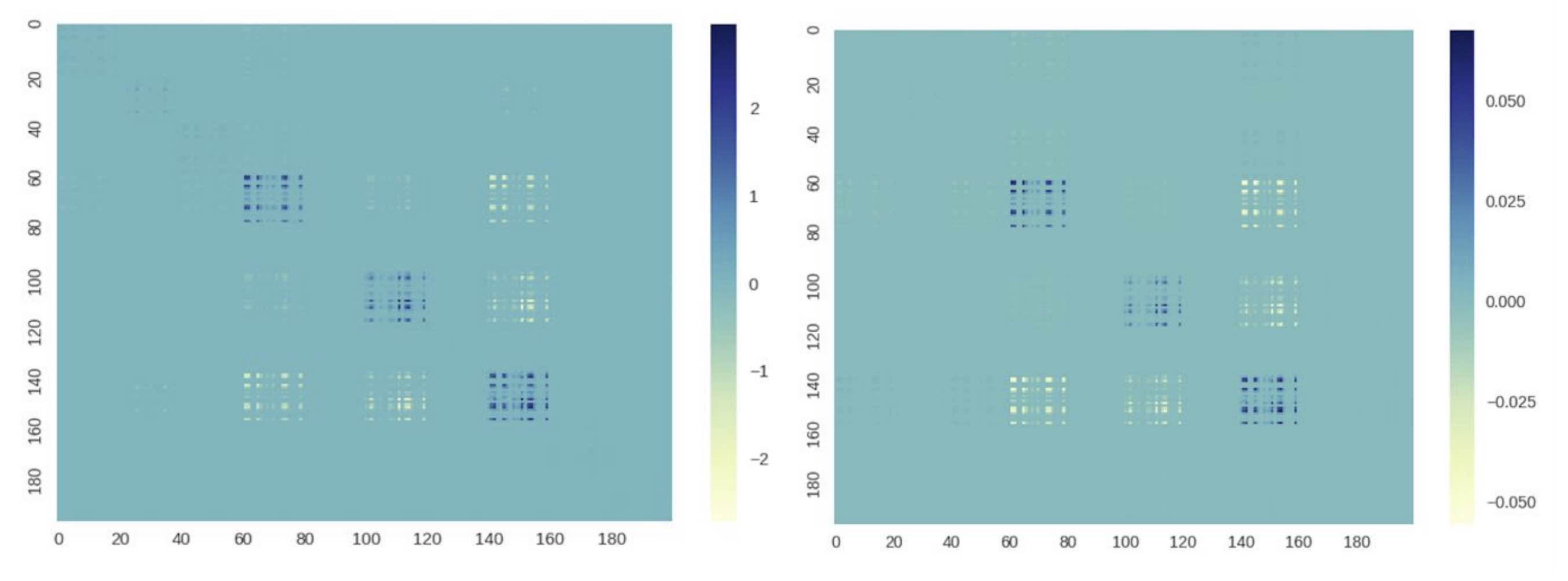}
		\caption{\textbf{At 50 epochs.} Test accuracy at this stage is $90.6\%$.}
		\label{fig:kfac_actual}
	\end{subfigure} 
	
	\caption{Last-layer Hessian and empirical Fisher blocks for \mlp $\, (784 \rightarrow 40 \rightarrow 20 \rightarrow 10)$ at different points of training on \mnist.  Both Hessian and empirical Fisher have been estimated over a batch of $64$ examples in all the figures. Hessian blocks are in the left-column, while empirical Fisher blocks are displayed in the right-column.}
	\label{fig:mnist-figures}
\end{figure}

While previously we compared the Hessian and the empirical Fisher at convergence, in this section our aim is to show that such a trend can be observed across various stages of training, including as early as $0.5$ epochs.

For our experimental setup, we first consider a fully-connected network trained on the \textsc{MNIST} digit recognition dataset. This fully-connected network has two hidden layers of size $40$ and $20$. \textsc{MNIST} consists of $28\times 28$ grayscale images for the digits $0-9$. Thus the overall shape of this network can be summarized as $784 \rightarrow 40 \rightarrow 20 \rightarrow 10$.

Note, here our purpose is not to get the best test accuracy, but rather we would like to inspect the structures of the Hessian and the empirical Fisher matrices. As a result, we choose the network with a relatively small size so as to exactly compute the full Hessian via double back-propagation. We use stochastic gradient descent (SGD) with a  constant learning rate of $0.001$ and a momentum of $0.5$ to train this network. The training set was subsampled to contain 5000 examples in order to prototype faster, and the batch size used during optimization was $64$. 

In Figure~\ref{fig:mnist-figures}, we compare the last-layer sub-matrices of sizes $200 \times 200$ for both Hessian and empirical Fisher at different stages of training. We see that both these matrices share a significant amount of similarities in their structure. In other words, if we were to compute say the correlation or cosine similarity between the two matrices, it would be quite high. In these plots, the number of samples used to build the estimates of Hessian and empirical Fisher was $16$, and similar trends can be observed if more samples are taken. 

Thus, we can establish that the empirical Fisher shares the same underlying structure as the Hessian, even at early stages of the training, where theoretically the model and data distribution still do not match.

\section{Detailed Results}

\subsection{One-shot Pruning}\label{sec:app_one_shot_detail}
In all the one-shot experiments, we use the \textsc{Torchvision} models for \textsc{ResNet-50} and \textsc{MobileNetV1} as dense baselines. 

\begin{figure}[h]
	\centering
	\includegraphics[width=0.8\linewidth]{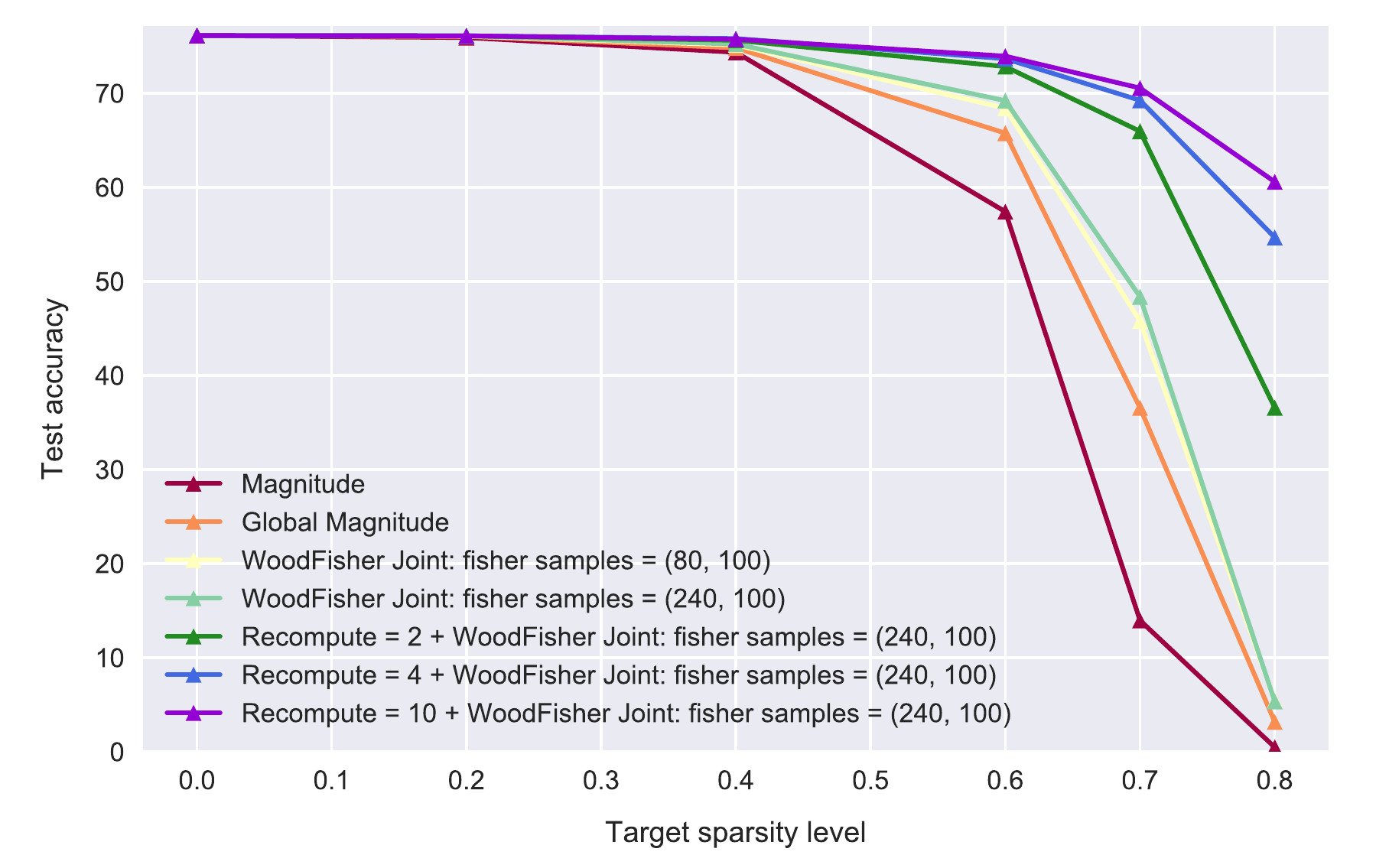}
	\caption{One-shot sparsity results for \textsc{ResNet-50} on \textsc{ImageNet}. In addition, we show here the effect of fisher subsample size as well as how the performance is greatly improved if we allow for recomputation of the Hessian (still no retraining). This is because the local quadratic model is only valid in a small neighbourhood (trust-region), beyond which it is not guaranteed to be accurate. The numbers corresponding to tuple of values called fisher samples refers to (fisher subsample size, fisher mini-batch size). A chunk size of $1000$ was used for this experiment. }
	\label{fig:one-shot-imagenet}
\end{figure}

\paragraph{\textsc{ResNet-50} on \textsc{ImageNet}.}
Here, all layers except the first convolutional layer are pruned to the respective sparsity values in a single shot, i.e., without any re-training. Figure~\ref{fig:one-shot-imagenet} shows how WoodFisher outperforms Global Magnitude and Magnitude with as few as 8,000 data samples. Increasing the fisher subsample size from $80$ to $240$ further helps a bit, and Table~\ref{tab:effect_resnet} properly investigates the effect of the fisher parameters, namely fisher subsample size and fisher mini-batch size, on the performance.

\begin{table}[h]\centering\ra{1.2}
	\centering
	
	\resizebox{0.8\textwidth}{!}{
		\begin{tabular}{@{}ccccc@{}}
			\toprule
			\multirow{1}{*}{Sparsity ($\%$)} & Fisher mini-batch size  & \multicolumn{1}{c}{Dense: Top-1 accuracy (\%)} & \multicolumn{2}{c}{Pruned: Top-1 accuracy (\%)}  \\
			\cmidrule(l{3pt}r{3pt}){4-5}
			& & & \multicolumn{2}{c}{Fisher subsample size} \\
			& & & 80 & 160 \\
			\midrule
			\multirow{6}{*}{30}	& 1  & \multirow{6}{*}{76.13}	& $55.81	\pm 3.28$ &  $ 57.53	\pm 1.62$ \\
			& 30  & 	& $75.17	\pm 0.09$ &  $ 75.26	\pm 0.17$ \\
			& 100  & 	& $75.73	\pm 0.03$ &  $ 75.77	\pm 0.08$ \\
			& 400  &	& $75.80	\pm 0.01$ &  $ 75.80	\pm 0.04$ \\
			& 800  &	& $75.74	\pm 0.04$ &  $ 75.71	\pm 0.08$ \\
			& 2400  & 	& $75.76	\pm 0.04$ &  $ 75.72	\pm 0.06$ \\ \midrule
			\multirow{6}{*}{50}	& 1  & \multirow{6}{*}{76.13}	& $48.76	\pm 4.95$ &  $ 51.23	\pm 3.11$ \\
			& 30  &	& $73.02	\pm 0.09$ &  $ 73.27	\pm 0.25$ \\
			& 100  & 	& $73.70	\pm 0.13$ &  $ 73.80	\pm 0.08$ \\
			& 400  &	& $73.66	\pm 0.02$ &  $ 73.73	\pm 0.02$ \\
			& 800  &	& $73.43	\pm 0.08$ &  $ 73.47	\pm 0.06$ \\
			& 2400  & 	& $73.32	\pm 0.10$ &  $ 73.30	\pm 0.04$ \\ \bottomrule
		\end{tabular}
	}\vspace{-1mm}
	\caption{\footnotesize{Effect of fisher subsample size and fisher mini-batch size on  one-shot pruning performance of WoodFisher, for \textsc{ResNet-50} on \textsc{ImageNet}. A chunk size of $1000$ was used for this experiment. The resutlts are averaged over three seeds.}}\label{tab:effect_resnet}
	
\end{table}

Importantly, in Figure~\ref{fig:one-shot-imagenet}, we observe that if we allow \textit{recomputing the Hessian inverse estimate} during pruning (but without retraining), it leads to significant improvements, since the local quadratic model is valid otherwise only in a small neighbourhood or trust-region. 

Note, this kind of recomputation can also be applied in the settings discussed below. However, having shown here for the primary example of \textsc{ResNet-50} on \textsc{ImageNet}, we skip the recomputation results for other secondary settings detailed below. 

\paragraph{\textsc{MobileNetV1} on\textsc{ ImageNet}.} Similarly, we perform one-shot pruning experiments for \textsc{MobileNetV1} on \textsc{ImageNet}. Here, also we find that WoodFisher outperforms both Global Magnitude and Magnitude. 

\begin{figure}
	\centering
	\includegraphics[width=0.8\linewidth]{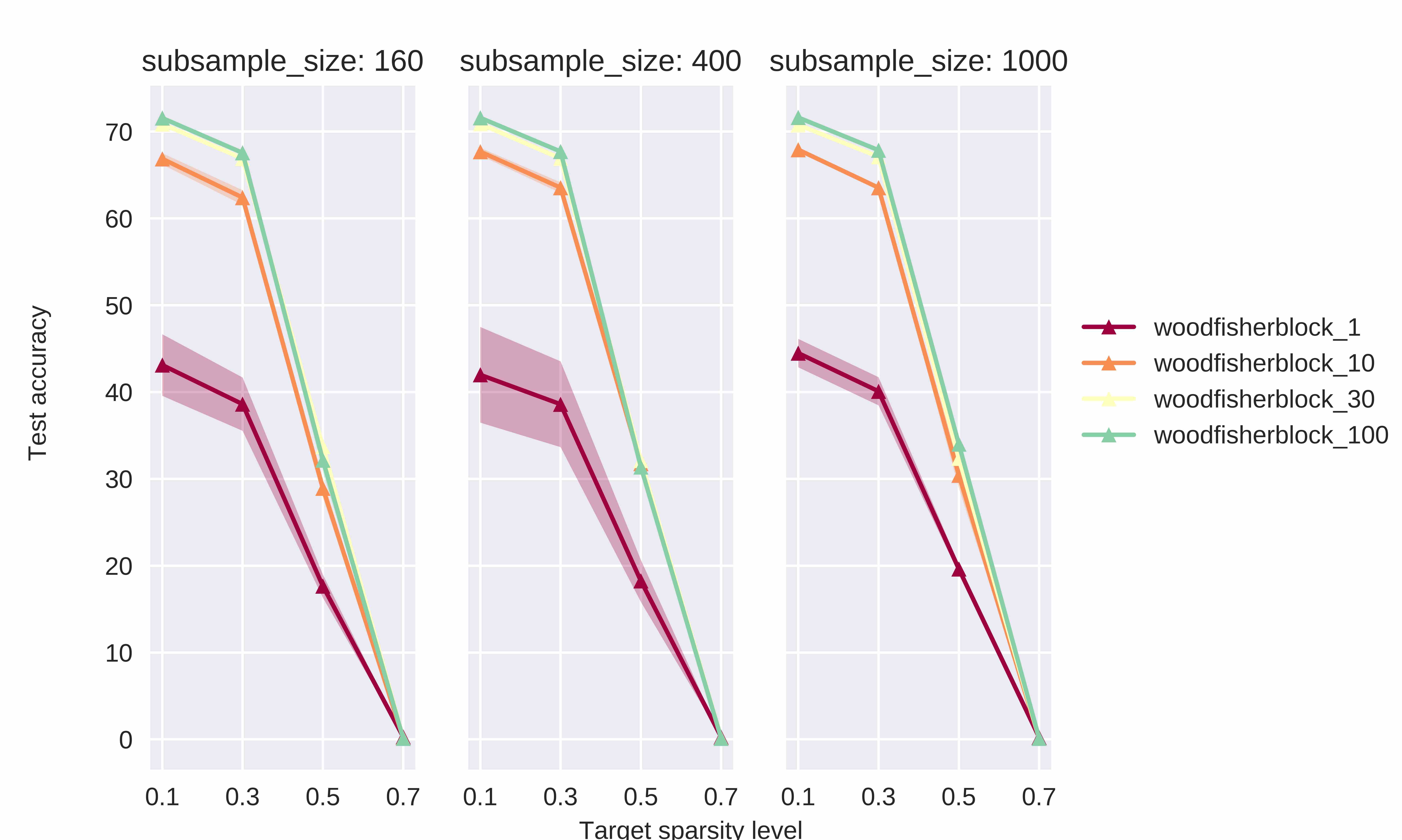}
	\caption{Effect of fisher mini batch size, across various values of fisher subsample size, on the one-shot pruning performance of WoodFisher for \textsc{MobileNetV1} on \textsc{ImageNet}.}
	\label{fig:effect-mini-bsz}
\end{figure}

Next, Figure~\ref{fig:effect-mini-bsz} shows the effect of fisher mini batch size, across various values of fisher subsample size, for this scenario. We notice that fisher mini-batch serves as a nice trick, which helps us take advantage of larger number of samples in the dataset at a much less cost. Further, in Table~\ref{tab:effect_mobilenet} presents the exact numbers for these experiments

\begin{table}[h]\centering\ra{1.2}
	\centering
	
	\resizebox{0.8\textwidth}{!}{
		\begin{tabular}{@{}ccccc@{}}
			\toprule
			\multirow{1}{*}{Sparsity ($\%$)} & Fisher mini-batch size  & \multicolumn{1}{c}{Dense: Top-1 accuracy (\%)} & \multicolumn{2}{c}{Pruned: Top-1 accuracy (\%)}  \\
			\cmidrule(l{3pt}r{3pt}){4-5}
			& & & \multicolumn{2}{c}{Fisher subsample size} \\
			& & & 160 & 400 \\
			\midrule
			\multirow{6}{*}{10}	& 1  & \multirow{6}{*}{71.76}	& $43.11 \pm 4.34$ & $41.99	\pm 6.75$	\\
			& 10  & 	& $66.86 \pm 0.79$ & $67.69	\pm 0.53$	\\
			& 30  & & $70.88 \pm 0.12$ & $70.89	\pm 0.13$	\\
			& 100  & & $71.56 \pm 0.05$ & $71.59	\pm 0.14$	\\
			& 400  & 	& $71.75 \pm 0.04$ & $71.79	\pm 0.05$	\\
			& 2400  & 	& $71.79 \pm 0.01$ & $71.77	\pm 0.08$	\\ \midrule
			\multirow{6}{*}{30}	& 1  & \multirow{6}{*}{71.76}	& $38.60 \pm 3.76$ & $38.60	\pm 6.05$	\\
			& 10  & 	& $62.40 \pm 1.03$ & $63.54	\pm 0.74$	\\
			& 30  & 	& $66.90 \pm 0.11$ & $66.92	\pm 0.17$	\\
			& 100  & 	& $67.55 \pm 0.12$ & $67.71	\pm 0.08$	\\
			& 400  &	& $67.88 \pm 0.06$ & $67.96	\pm 0.12$	\\
			& 2400  & 	& $67.88 \pm 0.05$ & $67.99	\pm 0.06$	\\	\midrule
			\multirow{6}{*}{50}	& 1  & \multirow{6}{*}{71.76}	& $17.64 \pm 1.62$ & $18.25	\pm 2.96$	\\
			& 10  & 	& $28.91 \pm 1.30$ & $31.75	\pm 0.74$	\\
			& 30  &	& $33.71 \pm 0.42$ & $32.10	\pm 0.42$	\\
			& 100  & 	& $32.15 \pm 1.70$ & $31.37	\pm 0.34$	\\
			& 400  &	& $32.30 \pm 0.40$ & $31.46	\pm 0.13$	\\
			& 2400  & 	& $32.39 \pm 0.67$ & $32.06	\pm 0.65$	\\	\bottomrule
		\end{tabular}
	}\vspace{-1mm}
	\caption{\footnotesize{Effect of fisher subsample size and fisher mini-batch size on  one-shot pruning performance of WoodFisher, for \textsc{MobileNetV1} on \textsc{ImageNet}. A chunk size of $10,000$ was used for this experiment. The results are averaged over three seeds.}}\label{tab:effect_mobilenet}
	
\end{table}

\paragraph{Effect of chunk size.}
For networks which are much larger than \textsc{ResNet-20}, we  also need to split the layerwise blocks into smaller chunks along the diagonal. So, here we study the effect of this chunk-size on the performance of WoodFisher. We take the setting of \textsc{ResNet-20} on \textsc{CIFAR10} and evaluate the performance for chunk-sizes in the set, $\lbrace20, \,100, \,1000, \,5000, \,12288, \,37000\rbrace$. Note that, $37000$ corresponds to the size of the block for the layer with the most number of parameters. Thus, this would correspond to taking the complete blocks across all the layers. 

\begin{figure}[h]
	\centering
	\includegraphics[width=0.7\linewidth]{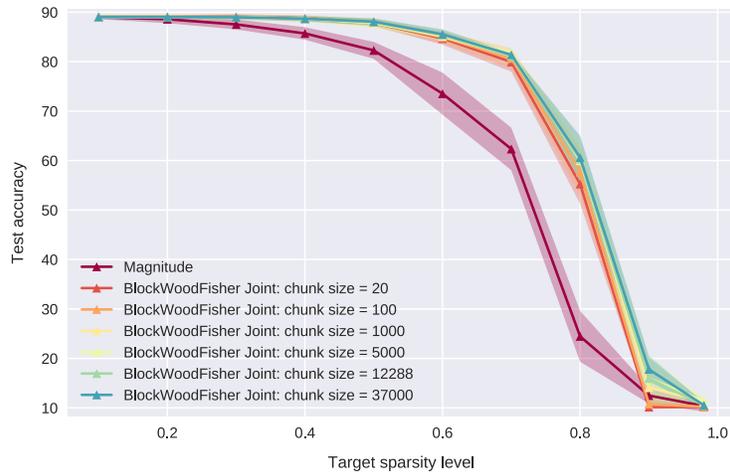}
	\caption{Effect of chunk size on one-shot sparsity results of WoodFisher \textbf{joint }for \textsc{ResNet-20} on \textsc{Cifar10}.}
	\label{fig:app-ablation-joint}
\end{figure}

\begin{figure}[h]
	\centering
	\includegraphics[width=0.7\linewidth]{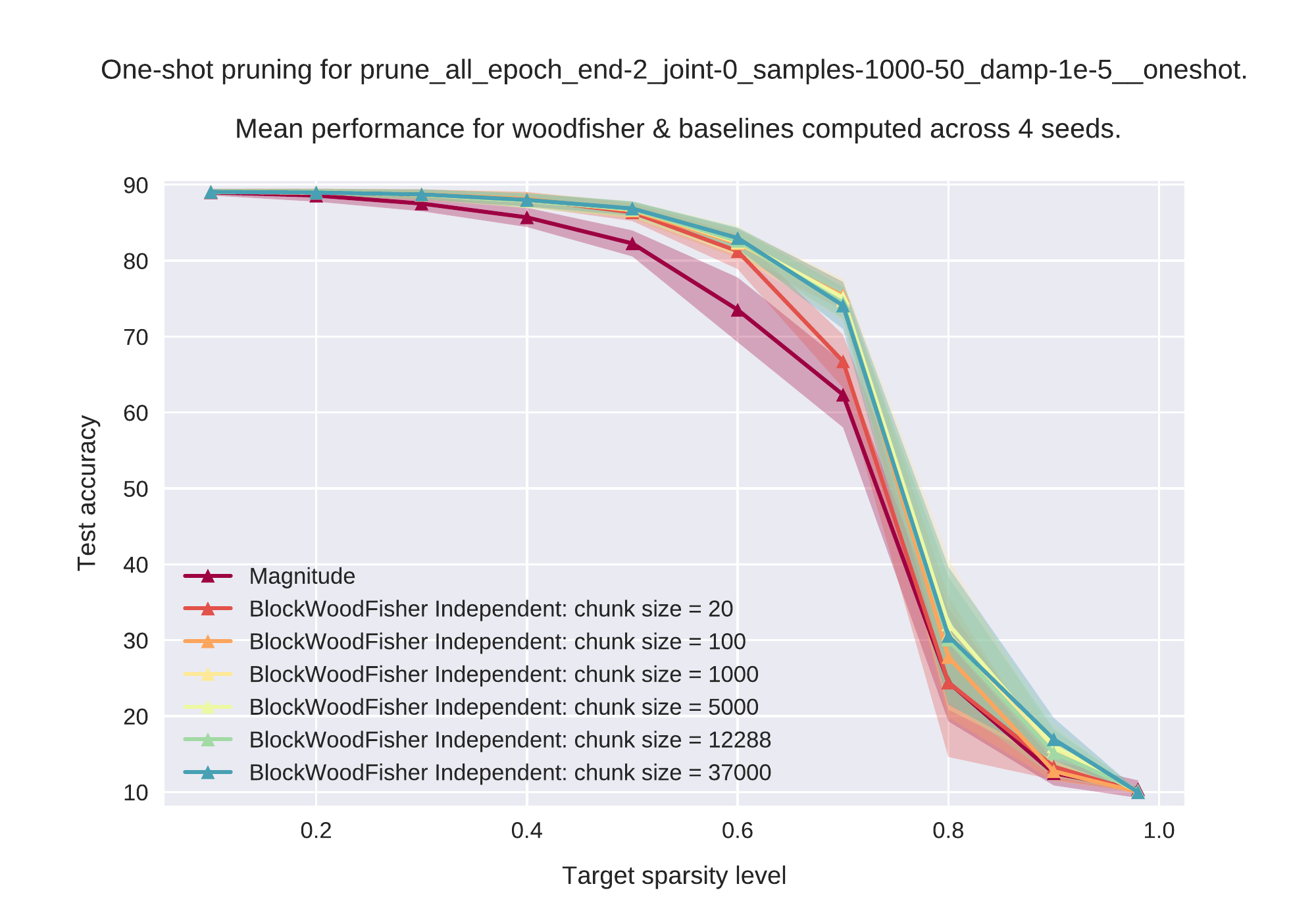}
	\caption{Effect of chunk size on one-shot sparsity results of WoodFisher \textbf{independent }for \textsc{ResNet-20} on \textsc{Cifar10}.}
	\label{fig:app-ablation-indep}
\end{figure}

Figures~\ref{fig:app-ablation-joint} and ~\ref{fig:app-ablation-indep} illustrate the impact of the block sizes used on the performance of WoodFisher in joint and independent mode respectively. We observe that performance of WoodFisher increases monotonically as the size of the blocks (or chunk-size) is increased, for both the cases. This fits well with our expectation that a large chunk-size would lead to a more accurate estimation of the inverse. However, it also tells us that even starting from blocks of size as small as $100$, there is a significant gain in comparison to magnitude pruning. 

\begin{figure}[h!]
	
	\centering
	\includegraphics[width=0.8\linewidth]{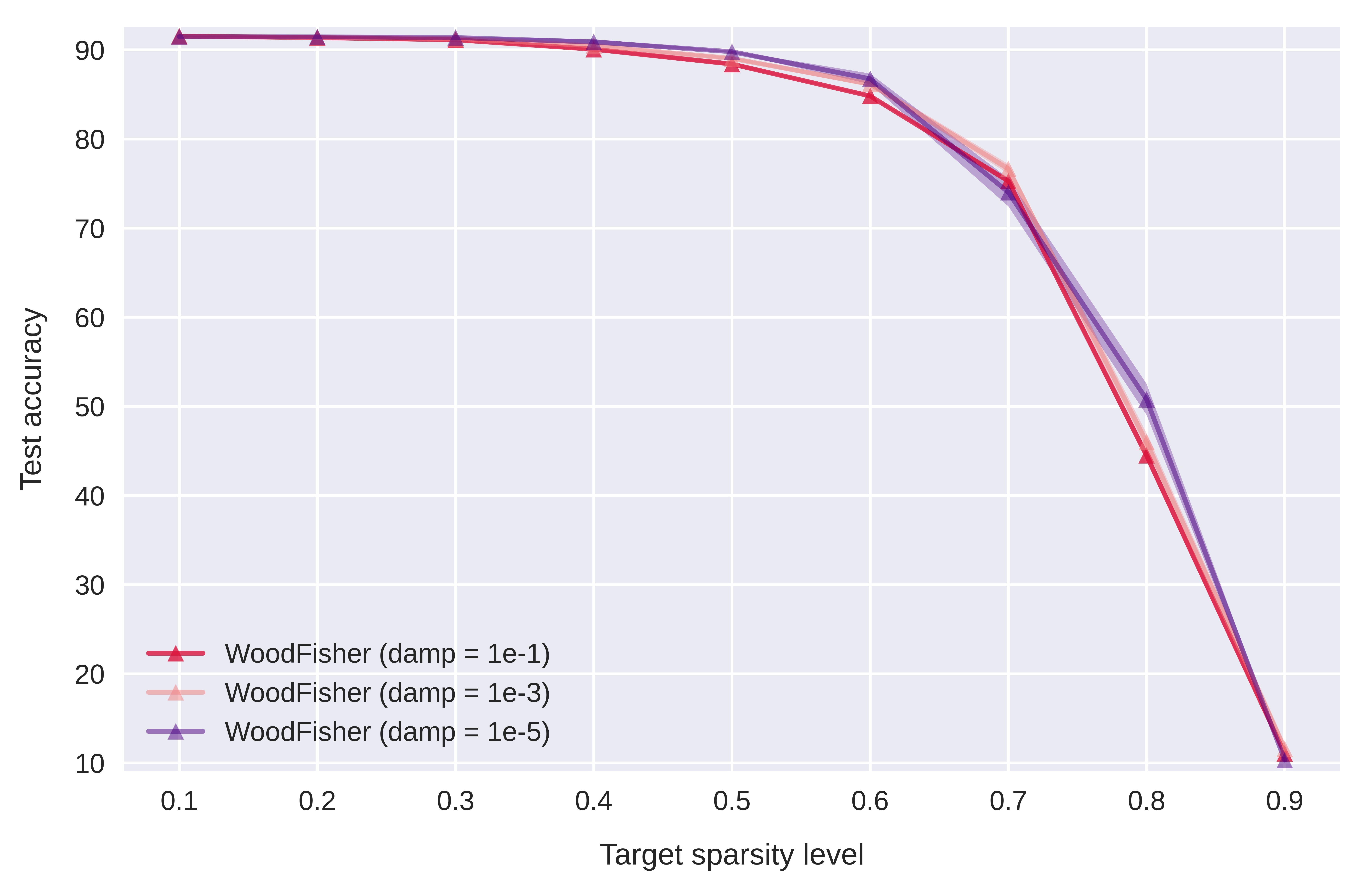}
	
	\caption{Effect of the dampening $\lambda$ on one-shot pruning results of WoodFisher (\textsc{ResNet-20}, \textsc{CIFAR10}) (avg over 4 seeds). As one would expect, the lower dampening value of $1e-5$ performs slightly better on average than the other values. This also highlights that the performance of WoodFisher is insensitive to the dampening $\lambda$.}
	\label{fig:damp}
	\vspace{-1mm}
\end{figure}

\paragraph{Effect of the dampening parameter.}
Regarding $\lambda$, we selected a small value so that the Hessian is not dominated by the dampening. 
We note that the algorithm is largely insensitive to this dampening value, Fig~\ref{fig:damp}. 

\clearpage
\subsection{Gradual Pruning}\label{sec:app_gradual_detail}

\begin{figure}[h!]
	\centering
	\begin{subfigure}{0.49\textwidth}
		\centering 
		\includegraphics[width=\linewidth]{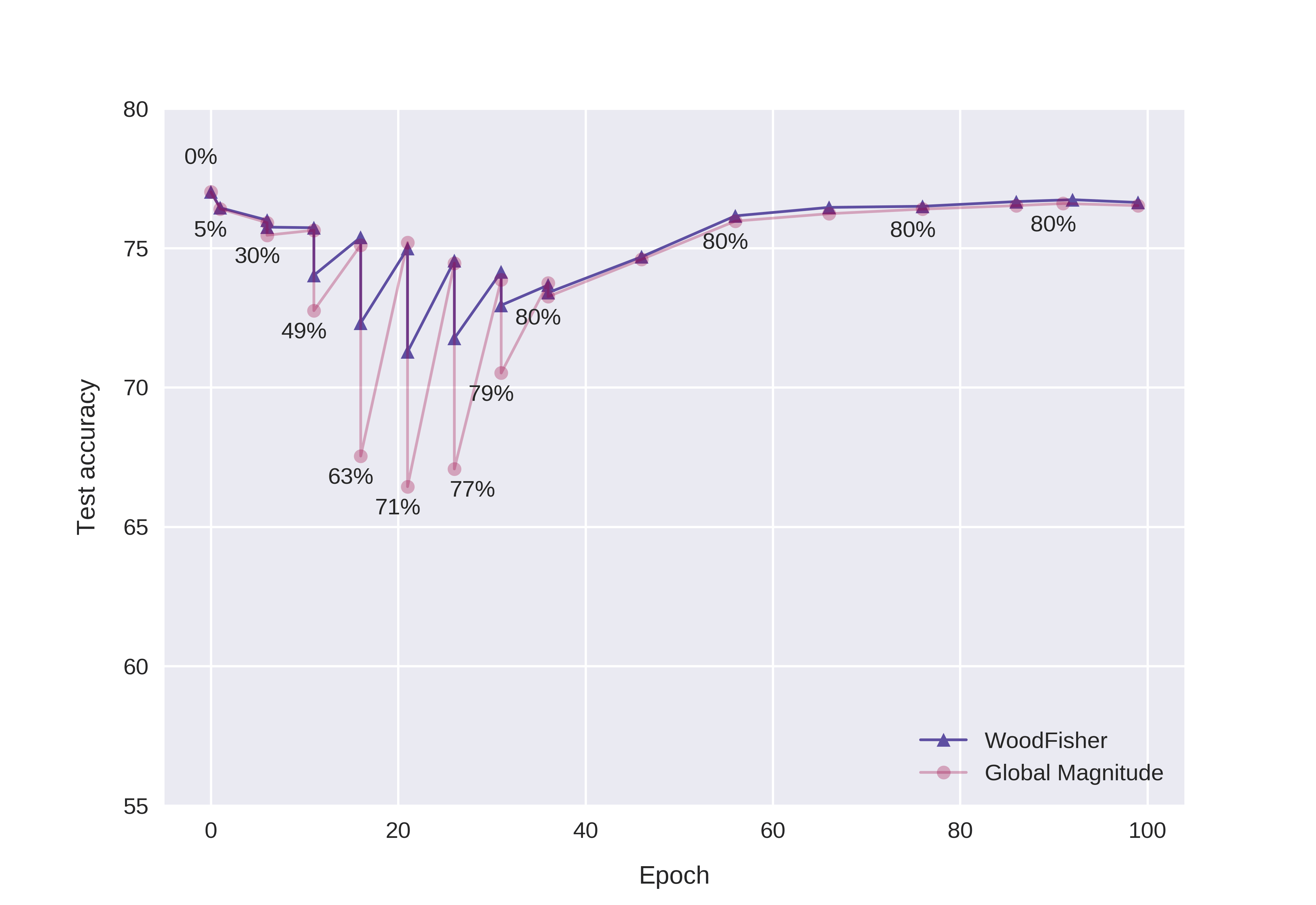}
		\caption{Target sparsity: $80\%$}
	\end{subfigure}
	\begin{subfigure}{0.49\textwidth}
		\centering
		\includegraphics[width=\linewidth]{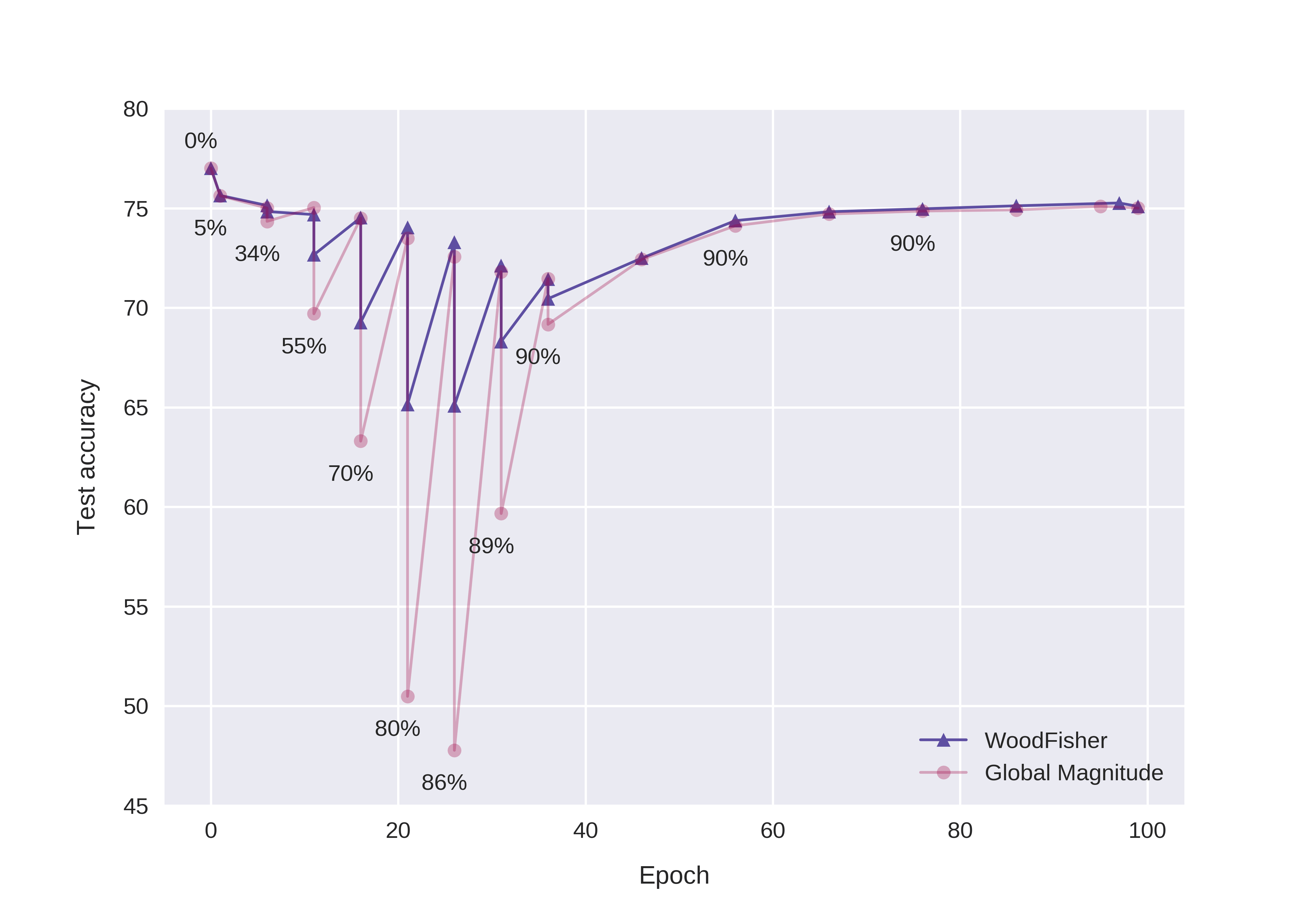}
		\caption{Target sparsity: $90\%$}
	\end{subfigure}
	\begin{subfigure}{0.49\textwidth}
		\centering 
		\includegraphics[width=\linewidth]{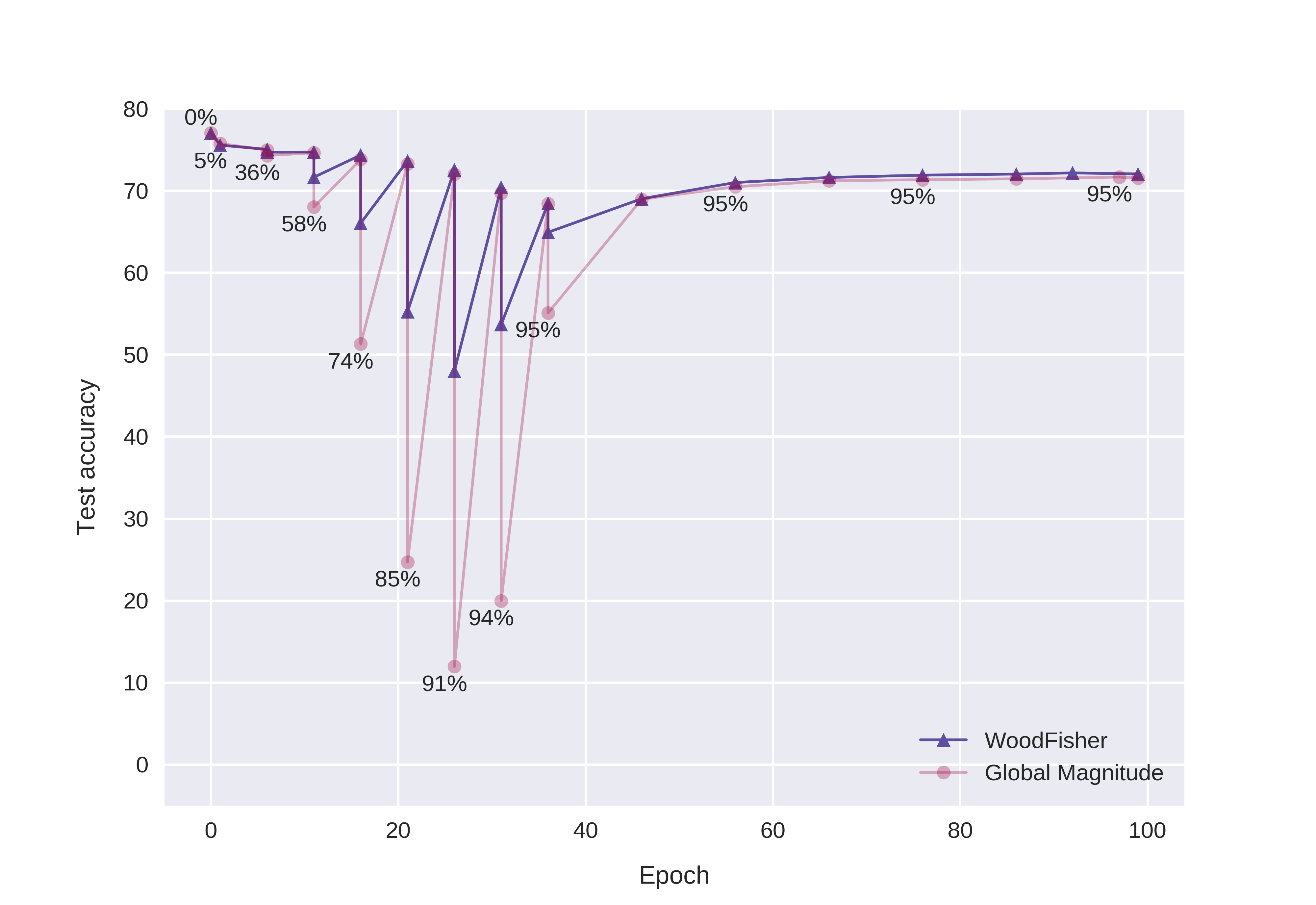}
		\caption{Target sparsity: $95\%$}
	\end{subfigure}
	\begin{subfigure}{0.49\textwidth}
		\centering
		\includegraphics[width=\linewidth]{images/str_imagenet_98_grad_fig_final_thin.png}
		\caption{Target sparsity: $98\%$}
	\end{subfigure}
	\label{fig:gradual-resnet50-regimes}
	\caption{\small The course of gradual pruning with points annotated by the corresponding sparsity amounts, for\textbf{ \textsc{ResNet-50} on \textsc{ImageNet}} across the different sparsity regimes.}
\end{figure}

\begin{figure}[h!]
	\centering
	\begin{subfigure}{0.49\textwidth}
		\centering 
		\includegraphics[width=\linewidth]{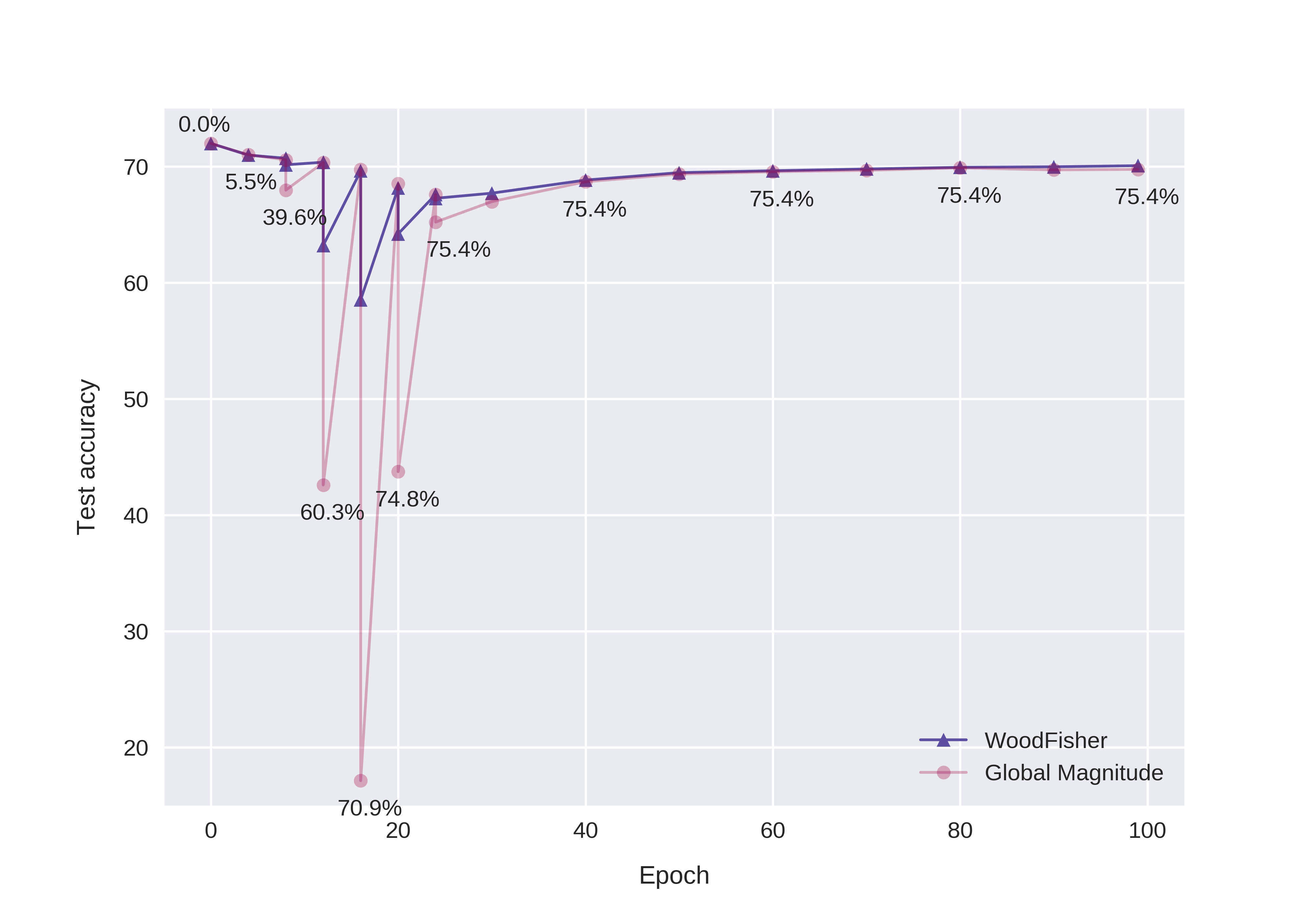}
		\caption{Target sparsity: $75.3\%$}
	\end{subfigure}
	\begin{subfigure}{0.49\textwidth}
		\centering
		\includegraphics[width=\linewidth]{images/str_mobilenet_89_grad_fig_final_thin.png}
		\caption{Target sparsity: $89\%$}
	\end{subfigure}
	\label{fig:gradual-mobilenet-regimes}
	\caption{\small The course of gradual pruning with points annotated by the corresponding sparsity amounts, for\textbf{ \textsc{MobileNetV1} on \textsc{ImageNet} }across the different sparsity regimes.}
\end{figure}

\clearpage

\begin{figure}
	\centering
	\includegraphics[width=0.8\linewidth]{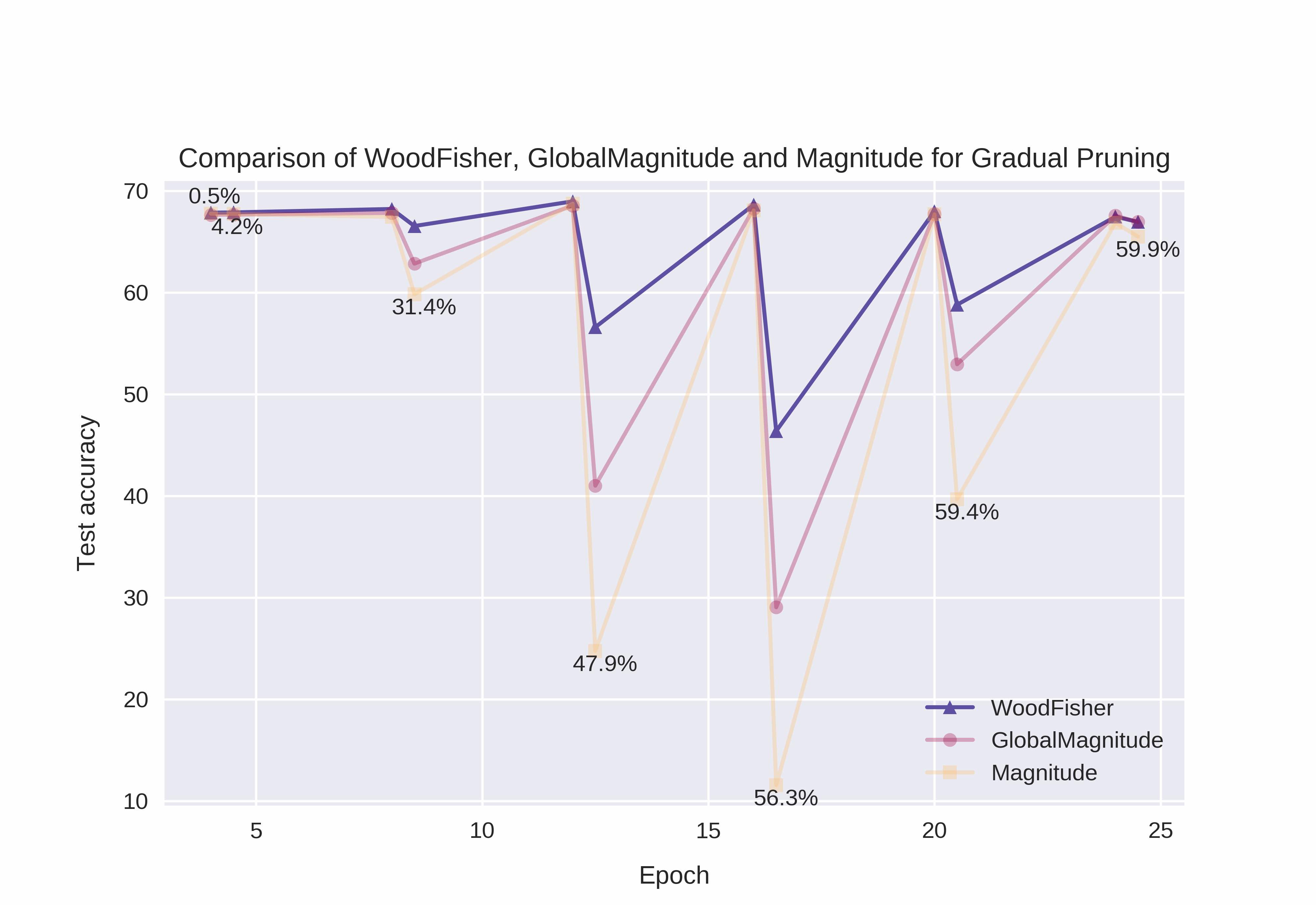}
	\caption{Comparison of the pruning phase during gradual pruning for WoodFisher, Global Magnitude, and Magnitude when compressing \textsc{MobileNetV1} to $60\%$ on \textsc{ImageNet}. The labels on the line plot indicate the corresponding sparsity level. We observe that after each pruning step WoodFisher outperforms both Global Magnitude and Magnitude.}
	\label{fig:gradual-mobilenet-all}
\end{figure}

Besides, magnitude pruning, which prunes all layers equally performs even worse, and Figure~\ref{fig:gradual-mobilenet-all} showcases the comparison between pruning steps for all the three: WoodFisher, Global Magnitude, and Magnitude. Such a trend is consistent and this is why we omit the results for magnitude pruning, and instead compare mostly with global magnitude.

\clearpage
\section{Gradual pruning scores at best runs}\label{sec:app_best_run}
In the main text, we reported results which were averaged across two runs for both WoodFisher and Global Magnitude. Here, we present the best run results for each method in Tables~\ref{tab:imagenet_gradual_sota_best} and ~\ref{tab:wf-indep-best}. We can see that even here as well, Global Magnitude (or Magnitude pruning)  does not perform better than WoodFisher, across any of the pruning scenarios.

\begin{table}[h!] \centering\ra{1.1}
	\centering
	
	\resizebox{0.75\textwidth}{!}{
		\begin{tabular}{@{}lccccc@{}}
			\toprule
			
			\multirow{1}{*}{} & \multicolumn{2}{c}{Top-1 accuracy (\%)} &Relative Drop  & \multicolumn{1}{c}{Sparsity} & \multirow{1}{*}{Remaining }  \\
			\cmidrule(l{3pt}r{3pt}){2-3}
			\multirow{1}{*}{Method} & Dense {\small($D$)}  & Pruned  {\small($P$)}& {\small${100 \times \frac{ (P-D)}{D}}$} &   (\%) & \# of params\\
			\midrule
			DSR \citep{mostafa2019parameter} & 74.90 & 71.60  & -4.41 & 80.00 & 5.10 M\\
			Incremental \citep{zhu2017prune} & 75.95 & 74.25 &  -2.24 & 73.50 & 6.79 M \\
			DPF \citep{Lin2020Dynamic} & 75.95 & 75.13  & -1.08 & 79.90 & 5.15 M \\
			GMP + LS \citep{gale2019state} & 76.69 & 75.58  & -1.44 & 79.90 & 5.15 M  \\
			Variational Dropout \citep{molchanov2017variational} & 76.69 & 75.28  & -1.83  & 80.00 & 5.12 M  \\
			RIGL + ERK \citep{Evci2019RiggingTL} & 76.80 & 75.10 & -2.21 & 80.00 & 5.12 M \\
			SNFS + LS \citep{dettmers2019sparse}  & 77.00 & 74.90  & -2.73 &  80.00 &  5.12 M\\
			STR \citep{Kusupati2020SoftTW} & 77.01 & 76.19 & -1.06 & 79.55 & 5.22 M \\
			Global Magnitude & 77.01 & 76.60 & -0.53 & 80.00 & 5.12 M \\
			DNW \citep{wortsman2019discovering} & 77.50 & 76.20  & -1.67  & 80.00 & 5.12 M\\
			\textbf{\WF{WoodFisher}} & 77.01 & \WF{\textbf{76.78}} &  \WF{\textbf{-0.30}} & 80.00 & 5.12 M \\			
			\midrule
			
			GMP + LS \citep{gale2019state} & 76.69 & 73.91  & -3.62 & 90.00 & 2.56 M  \\
			Variational Dropout \citep{molchanov2017variational} & 76.69 & 73.84 & -3.72 & 90.27 & 2.49 M   \\
			RIGL + ERK \citep{Evci2019RiggingTL} & 76.80 & 73.00 & -4.94 & 90.00 & 2.56 M \\
			SNFS + LS \citep{dettmers2019sparse}  & 77.00 & 72.90  & -5.32 &  90.00 &  2.56 M\\
			STR \citep{Kusupati2020SoftTW} & 77.01 & 74.31 & -3.51 & 90.23 & 2.49 M \\
			Global Magnitude & 77.01 & 75.20 & -2.36  & 90.00 & 2.56 M \\
			DNW \citep{wortsman2019discovering} & 77.50 & 74.00  & -4.52   & 90.00 & 2.56 M\\
			\textbf{\WF{WoodFisher}} & 77.01 & \WF{\textbf{75.26}} & \WF{\textbf{-2.27}}  & 90.00 & 2.56 M \\			
			\midrule
			
			GMP  \citep{gale2019state} & 76.69 & 70.59  & -7.95 & 95.00 & 1.28 M \\
			Variational Dropout \citep{molchanov2017variational} & 76.69 & 69.41 & -9.49  & 94.92 & 1.30 M  \\
			Variational Dropout \citep{molchanov2017variational} & 76.69 & 71.81 & -6.36 & 94.94 & 1.30 M  \\
			RIGL + ERK \citep{Evci2019RiggingTL} & 76.80 & 70.00 & -8.85 & 95.00 & 1.28 M \\
			DNW \citep{wortsman2019discovering} & 77.01 & 68.30  & -11.31 & 95.00 & 1.28 M \\
			STR \citep{Kusupati2020SoftTW} & 77.01 & 70.97 & -7.84  & 94.80 & 1.33 M \\
			STR \citep{Kusupati2020SoftTW} & 77.01 & 70.40 & -8.58  & 95.03 & 1.27 M \\
			Global Magnitude & 77.01 &71.79 & -6.78 & 95.00 & 1.28 M \\
			\textbf{\WF{WoodFisher}} & 77.01 & \WF{\textbf{72.16}} & \WF{\textbf{-6.30}}  & 95.00 & 1.28 M \\
			\midrule
			
			GMP + LS \citep{gale2019state} & 76.69 & 57.90  &-24.50  &98.00  & 0.51 M \\
			Variational Dropout \citep{molchanov2017variational} & 76.69 & 64.52 & -15.87 &98.57 & 0.36 M   \\
			DNW \citep{wortsman2019discovering} & 77.01 & 58.20  & -24.42 & 98.00 & 0.51 M \\
			STR \citep{Kusupati2020SoftTW} & 77.01 & 61.46 & -20.19 & 98.05 & 0.50 M \\
			STR \citep{Kusupati2020SoftTW} & 77.01 & 62.84 & -18.40 & 97.78 & 0.57 M \\
			Global Magnitude & 77.01 & 64.39 & -16.38 & 98.00 & 0.51 M \\
			\textbf{\WF{WoodFisher}} & 77.01 & \WF{\textbf{65.63}} & \WF{\textbf{-14.78}} & 98.00 & 0.51 M \\
			
			\bottomrule
		\end{tabular}
	}\vspace{-1mm}
	\caption{\textbf{At best runs: }Comparison of WoodFisher gradual pruning results with the state-of-the-art approaches. LS denotes label smoothing, and ERK denotes the Erd\H{o}s-Renyi Kernel.}\label{tab:imagenet_gradual_sota_best}
\end{table}

\begin{table}[h!]\centering\ra{1.1}
	\centering
	
	\resizebox{0.75\textwidth}{!}{
		\begin{tabular}{@{}lccccc@{}}
			\toprule
			
			\multirow{1}{*}{} & \multicolumn{2}{c}{Top-1 accuracy (\%)} & Relative Drop  & \multicolumn{1}{c}{Sparsity} & \multirow{1}{*}{Remaining }  \\
			\cmidrule(l{3pt}r{3pt}){2-3}
			\multirow{1}{*}{Method} & Dense {\small($D$)} & Pruned  {\small($P$)} & {\small${100 \times \frac{ (P-D)}{D}}$} &   (\%) & \# of params\\
			\midrule
			GMP  & 77.01 & 75.18  & -2.38 & 89.1 & 2.79 M \\
			\textbf{\WF{WoodFisher (independent)}} & 77.01 & \WF{\textbf{75.23}} & \WF{\textbf{-2.31}} & 89.1 & 2.79 M \\
			\bottomrule
		\end{tabular}
	}
	\caption{\small \textbf{At best runs: }comparison of WoodFisher and magnitude pruning \textbf{with the same layer-wise sparsity targets }as used in~\citep{gale2019state} for \textsc{ResNet-50} on \textsc{ImageNet}. Namely, this involves skipping the first convolutional layer, pruning the last fully connected layer to $80\%$ and the rest of the layers equally to $90\%$.}
	\label{tab:wf-indep-best}
	
\end{table}

 \clearpage
 \section{FLOPs and Inference Costs}\label{sec:app_flops}
 In the discussion so far, our objective has been to prune the model parameters in such a way that the accuracy is preserved or decreases minimally. For several practical use-cases, another important criterion to consider while pruning is to reduce the inference costs. However, our current procedure~\ref{sec:compression} is solely designed to maintain the accuracy of the pruned model. 
 
 \paragraph{FLOPs-aware pruning statistic.} Thus, to take into account the inference costs, we normalize the pruning statistic by the respective theoretical FLOP (floating point operation) cost for that parameter. The FLOP cost for a parameter ($\text{FLOPs-per-param}_q$) is considered as the FLOP cost for the layer it is present in divided by the number of parameters in that layer. Overall, we can define the FLOPs-aware pruning statistic as follows, 
 \begin{equation}\label{eq:prun_stat_flop}
 	\boxed{\rho^{\text{FLOPs}}_q  = \frac{\rho_q}{\big(\text{FLOPs-per-param}_q\big)^\beta}}.
 \end{equation}
 
 where, $\rho_q$ refers to the usual pruning statistic (as in Eq.~\eqref{eq:prun_stat}) and $\beta \geq 0$ is a hyper-parameter that controls the influence of FLOP costs in the net pruning statistic (for $\beta=0$, we recover the pruning statistic $\rho_q$). Intuitively, we are normalizing the sensitivity (of the loss) to removal of parameter $q$ (in other words, the pruning statistic $\rho_q$) by its corresponding FLOP cost. Note, during the course of gradual pruning, both the FLOP costs and the number of parameters are measured with respect to the active (non-pruned) parameters in the corresponding layer. 
 
 The benefits of such a formulation are that we still do not need to manually design a FLOPs-aware sparsity distribution for the model and we can control the FLOPs-accuracy trade-off as required, plus the fact that the rest of pruning procedure remains exactly the same. 
 
 To illustrate the effect of this FLOPs-aware pruning statistic, we consider one of the sparsity regimes for both \textsc{ResNet50} and \textsc{MobileNetV1} and compare the FLOP costs in Table~\ref{tab:flop-aware}. 
 
 \begin{table}[h!]\centering\ra{1.1}
 	\centering
 	
 	\resizebox{\textwidth}{!}{
 		\begin{tabular}{@{}lcccccccc@{}}
 			\toprule
 			
 			\multirow{1}{*}{Method}  & \multirow{1}{*}{Model} & \multicolumn{2}{c}{Top-1 accuracy (\%)} & Relative Drop  & \multicolumn{1}{c}{Sparsity}  & \multirow{1}{*}{Remaining }  & \multirow{1}{*}{FLOPs}\\
 			\cmidrule(l{3pt}r{3pt}){3-4}
 			\multirow{1}{*}{}  & \multirow{1}{*}{} & Dense {\small($D$)} & Pruned  {\small($P$)} & {\small${100 \times \frac{ (P-D)}{D}}$} &   (\%) & \# of params & \\
 			\midrule
 			
 			STR \citep{Kusupati2020SoftTW} & \multirow{3}{*}{\textsc{ResNet50}} & 77.01 & 74.31 & -3.51 & 90.23 & 2.49 M  & 343 M\\
 			STR \citep{Kusupati2020SoftTW} & & 77.01 & 74.01 & -3.90 & 90.55 & 2.41 M  & 341 M\\
 			\textbf{\WF{WoodFisher ($\beta=0.00$)}} & & 77.01 & \WF{\textbf{75.26}} & \WF{\textbf{-2.27}}  & 90.00 & 2.56 M & 594 M\\	
 			\textbf{\WF{WoodFisher ($\beta=0.30$)}} & & 77.01 & 74.34 & -3.47 & 90.23 & 2.49 M & \WF{\textbf{335 M}} \\	
 			\midrule
 			STR \citep{Kusupati2020SoftTW} & \multirow{3}{*}{\textsc{MobileNetV1}} & 72.00 & 68.35  & -5.07 & 75.28 & 1.04 M & 101 M\\
 			\textbf{\WF{WoodFisher ($\beta=0.00$)}} & & 72.00 & \WF{\textbf{70.09}} & \WF{\textbf{-2.65}}& 75.28 &  1.04 M & 159 M \\
 			\textbf{\WF{WoodFisher ($\beta=0.30$)}} & &72.00 & 69.26 & -3.81 & 75.28 &  1.04 M & 101 M\\
 			\textbf{\WF{WoodFisher ($\beta=0.35$)}} & &72.00 & 68.69 & -4.60 & 75.28 &  1.04 M & \WF{\textbf{92 M}}\\
 			
 			\bottomrule
 		\end{tabular}
 	}
 	\caption{\footnotesize{Comparison of WoodFisher (WF) and STR across theoretical FLOP counts for \textsc{ResNet50} and \textsc{MobileNetV1} on \textsc{ImageNet} in $90\%$ and $75\%$ sparsity regime respectively.}}\label{tab:flop-aware}
 	
 \end{table}
 
 First, we observe that interestingly while STR~\citep{Kusupati2020SoftTW} performs significantly worse than WoodFisher in terms of the pruned model accuracy, the theoretical FLOP costs for STR are much better than WoodFisher (or Global Magnitude).  Roughly, this is because STR leads to sparsity profiles that are relatively more ``uniform'' across layers, whereas WoodFisher and Global Magnitude may in theory arbitrarily re-distribute sparsity across layers. (In practice, we note that the sparsity profiles generated by these methods do correlate layer sparsity with the number of parameters in the layer.) 
 
 However, if we incorporate the FLOPs-aware pruning statistic (with $\beta=0.30$) for WoodFisher, then it improves over STR simultaneously along the axis of accuracy and FLOP costs. E.g., we see that for both the settings of \textsc{ResNet50} ($\beta=0.30$)  and \textsc{MobileNetV1} ($\beta=0.35$), WoodFisher has a higher accuracy than STR as well as lower FLOP costs. Plus, for the same FLOP costs of 101 M in case of \textsc{MobileNetV1}, WoodFisher results in $\approx 1\%$ accuracy gain over STR. This resulting improvement over STR illustrates the benefit of using FLOPs-aware pruning statistic for WoodFisher, as STR is claimed to be a state-of-the-art method for unstructured pruning that in addition minimizes FLOP costs. 
 
 The sparsity profiles for the WoodFisher pruned models with lower FLOP counts can be found in Section~\ref{sec:app_spar_pattern_flops}.
 
\paragraph{Effect of FLOPs-Aware on actual inference times. }
 
 \begin{table}[h] \centering\ra{1.1}
 	\centering
 	\resizebox{0.65\textwidth}{!}{
 		\begin{tabular}{@{}lcccc@{}}
 			\toprule
 			
 			\multirow{1}{*}{} & \multicolumn{2}{c}{Inference Time (ms)} & Top-1 Acc. \\
 			\cmidrule(l{3pt}r{3pt}){2-3}
 			\multirow{1}{*}{Compression} & Batch 1 & Batch 64 \\
 			\midrule
 			Dense  & 7.1 & 296 &  77.01\%  \\ 			\midrule
 			STR-81.27\% & 5.6 & 156  & 76.12\% \\
 			\textbf{\WF{WF-Joint-80\%}} & 6.3 & 188  & 76.73\% \\
 			\midrule
 			STR-90.23\% & 3.8 & 144  & 74.31\% \\
 			\textbf{\WF{FLOPs-aware WF-Joint-90.23\%}} & 3.8 & 116  & 74.34\% \\
 			\textbf{\WF{WF-Independent-89.1\%}} & 4.3 & 157  & 75.23\% \\
 			\textbf{\WF{WF-Joint-90\%}} & 5.0 & 151  & 75.26\% \\
 			\bottomrule
 		\end{tabular}
 	}
 	
 	\caption{\footnotesize{Comparison of inference times at batch sizes 1 and 64 for various sparse models, executed on the framework of~\citep{NM}, on an 18-core Haswell CPU. The table also contains the Top-1 Accuracy for the model on the ILSVRC validation set}. The difference with respect to the Table~\ref{tab:inference} is that here we include the results of FLOPs-aware variant too.}\label{tab:inference_flops_aware}
 \end{table}
 In Table~\ref{tab:inference_flops_aware}, we see that by considering the FLOPs-aware pruning statistic from the discussion before (with $\beta=0.30$), this results in $20\%$ relative improvement over STR on inference time at batch size 64.

\clearpage
\section{Sparsity distributions}\label{sec:app_spar_pattern}
As followed in the literature~\cite{Lin2020Dynamic,Kusupati2020SoftTW}, we prune only the weights in fully-connected and convolutional layers. This means that none of the batch-norm parameters or bias are pruned if present. The sparsity percentages in our work and others like~\citep{Kusupati2020SoftTW} are also calculated based on this. 

\subsection{ResNet-50}\label{sec:spar-rn50}

\begin{table}[!ht]\ra{1.2}
\centering
\caption{\small The obtained distribution of sparsity across the layers by WoodFisher and Global Magnitude when sparsifying \textsc{ResNet-50} to $80\%, 90\%, 95\%, 98\%$ levels on \textsc{ImageNet}.}
\label{tab:res50big}
\resizebox{\columnwidth}{!}{
\begin{tabular}{@{}l|r|cc|cc|cc|cc@{}}
\toprule
Module                           & \multicolumn{1}{c|}{\begin{tabular}[c]{@{}c@{}}Fully Dense \\ Params\end{tabular}}  & \multicolumn{8}{c}{Sparsity (\%)}                                                                                           \\ \midrule
 &  & WoodFisher &  Global Magni & WoodFisher &  Global Magni & WoodFisher &  Global Magni & WoodFisher &  Global Magni  \\ 

\multirow{1}{*}{Overall}                          & \multirow{1}{*}{25502912}  & \multicolumn{2}{c|}{$80\%$} &  \multicolumn{2}{c|}{$90\%$} & \multicolumn{2}{c|}{$95\%$} & \multicolumn{2}{c}{$98\%$} \\ \midrule

Layer 1 - conv1                  & 9408 & 37.04 & 37.61 & 44.97 & 45.72  & 51.65 & 51.86  &  60.63 &  60.09  \\
Layer 2 - layer1.0.conv1         & 4096 & 46.31 & 49.12 & 58.30 & 60.69  & 66.02 & 68.70  &  75.39 &  77.83  \\
Layer 3 - layer1.0.conv2         & 36864 & 68.18 & 68.69 & 79.48 & 80.19  & 86.81 & 87.54  &  93.03 &  93.45  \\
Layer 4 - layer1.0.conv3         & 16384 & 61.43 & 62.77 & 72.16 & 73.77  & 79.68 & 81.51  &  86.91 &  89.32  \\
Layer 5 - layer1.0.downsample.0  & 16384 & 56.10 & 57.78 & 66.31 & 68.08  & 74.10 & 75.67  &  81.68 &  83.82  \\
Layer 6 - layer1.1.conv1         & 16384 & 66.12 & 66.82 & 77.44 & 78.33  & 85.05 & 85.53  &  91.47 &  92.35  \\
Layer 7 - layer1.1.conv2         & 36864 & 71.19 & 71.78 & 82.65 & 83.01  & 89.15 & 89.52  &  94.19 &  94.76  \\
Layer 8 - layer1.1.conv3         & 16384 & 71.69 & 72.98 & 80.57 & 82.43  & 86.60 & 88.04  &  91.19 &  93.15  \\
Layer 9 - layer1.2.conv1         & 16384 & 60.47 & 61.04 & 73.18 & 74.16  & 81.82 & 83.04  &  89.84 &  90.91  \\
Layer 10 - layer1.2.conv2        & 36864 & 59.54 & 60.04 & 73.37 & 74.00  & 82.92 & 83.04  &  90.80 &  91.47  \\
Layer 11 - layer1.2.conv3        & 16384 & 72.05 & 73.14 & 79.29 & 80.76  & 84.31 & 86.17  &  89.09 &  91.46  \\
Layer 12 - layer2.0.conv1        & 32768 & 58.69 & 59.70 & 71.70 & 73.05  & 80.73 & 81.65  &  88.53 &  90.12  \\
Layer 13 - layer2.0.conv2        & 147456 & 71.07 & 71.44 & 83.83 & 84.42  & 90.77 & 91.20  &  95.70 &  96.25  \\
Layer 14 - layer2.0.conv3        & 65536 & 73.68 & 74.59 & 82.89 & 83.84  & 88.41 & 89.45  &  93.03 &  94.33  \\
Layer 15 - layer2.0.downsample.0 & 131072 & 80.55 & 81.24 & 88.98 & 89.66  & 93.31 & 94.04  &  96.60 &  97.22  \\
Layer 16 - layer2.1.conv1        & 65536 & 80.91 & 81.47 & 89.38 & 89.68  & 93.74 & 94.32  &  96.84 &  97.35  \\
Layer 17 - layer2.1.conv2        & 147456 & 77.50 & 77.66 & 87.38 & 87.63  & 92.82 & 92.95  &  96.57 &  96.71  \\
Layer 18 - layer2.1.conv3        & 65536 & 76.52 & 77.78 & 85.39 & 86.41  & 90.42 & 91.59  &  94.53 &  95.68  \\
Layer 19 - layer2.2.conv1        & 65536 & 72.53 & 73.08 & 84.34 & 85.05  & 90.64 & 91.40  &  95.21 &  96.06  \\
Layer 20 - layer2.2.conv2        & 147456 & 75.94 & 76.14 & 87.02 & 87.27  & 92.55 & 92.98  &  96.48 &  96.88  \\
Layer 21 - layer2.2.conv3        & 65536 & 69.96 & 70.99 & 82.19 & 83.21  & 88.65 & 89.91  &  93.77 &  95.04  \\
Layer 22 - layer2.3.conv1        & 65536 & 70.29 & 70.74 & 82.43 & 83.14  & 89.11 & 89.94  &  94.29 &  95.17  \\
Layer 23 - layer2.3.conv2        & 147456 & 72.67 & 72.75 & 84.80 & 85.04  & 91.15 & 91.53  &  95.85 &  96.23  \\
Layer 24 - layer2.3.conv3        & 65536 & 73.87 & 74.68 & 84.12 & 85.37  & 89.84 & 90.91  &  94.23 &  95.49  \\
Layer 25 - layer3.0.conv1        & 131072 & 62.86 & 63.53 & 76.43 & 77.31  & 84.79 & 85.61  &  91.73 &  92.67  \\
Layer 26 - layer3.0.conv2        & 589824 & 81.91 & 82.22 & 91.43 & 91.80  & 95.57 & 95.95  &  98.10 &  98.41  \\
Layer 27 - layer3.0.conv3        & 262144 & 72.28 & 72.95 & 84.20 & 85.00  & 90.76 & 91.48  &  95.39 &  96.13  \\
Layer 28 - layer3.0.downsample.0 & 524288 & 87.11 & 87.26 & 94.23 & 94.43  & 97.15 & 97.38  &  98.85 &  99.04  \\
Layer 29 - layer3.1.conv1        & 262144 & 85.79 & 85.99 & 93.09 & 93.43  & 96.38 & 96.76  &  98.32 &  98.64  \\
Layer 30 - layer3.1.conv2        & 589824 & 85.63 & 85.73 & 93.25 & 93.37  & 96.61 & 96.79  &  98.52 &  98.72  \\
Layer 31 - layer3.1.conv3        & 262144 & 77.65 & 78.16 & 88.41 & 89.09  & 93.59 & 94.24  &  96.99 &  97.58  \\
Layer 32 - layer3.2.conv1        & 262144 & 83.75 & 83.92 & 92.21 & 92.51  & 95.95 & 96.23  &  98.17 &  98.50  \\
Layer 33 - layer3.2.conv2        & 589824 & 84.99 & 84.97 & 93.31 & 93.42  & 96.73 & 96.94  &  98.63 &  98.86  \\
Layer 34 - layer3.2.conv3        & 262144 & 78.29 & 78.65 & 88.91 & 89.40  & 94.06 & 94.57  &  97.31 &  97.85  \\
Layer 35 - layer3.3.conv1        & 262144 & 81.17 & 81.27 & 90.86 & 91.14  & 95.15 & 95.52  &  97.86 &  98.21  \\
Layer 36 - layer3.3.conv2        & 589824 & 85.06 & 84.94 & 93.32 & 93.42  & 96.77 & 96.96  &  98.68 &  98.88  \\
Layer 37 - layer3.3.conv3        & 262144 & 80.29 & 80.71 & 89.93 & 90.42  & 94.54 & 95.08  &  97.54 &  98.01  \\
Layer 38 - layer3.4.conv1        & 262144 & 80.07 & 80.17 & 90.20 & 90.44  & 94.87 & 95.19  &  97.73 &  98.04  \\
Layer 39 - layer3.4.conv2        & 589824 & 84.99 & 84.95 & 93.24 & 93.37  & 96.75 & 96.92  &  98.65 &  98.88  \\
Layer 40 - layer3.4.conv3        & 262144 & 79.24 & 79.66 & 89.26 & 89.77  & 94.23 & 94.87  &  97.47 &  97.93  \\
Layer 41 - layer3.5.conv1        & 262144 & 75.83 & 75.99 & 87.68 & 87.91  & 93.44 & 93.81  &  97.07 &  97.48  \\
Layer 42 - layer3.5.conv2        & 589824 & 84.07 & 84.07 & 92.72 & 92.86  & 96.45 & 96.67  &  98.52 &  98.75  \\
Layer 43 - layer3.5.conv3        & 262144 & 75.90 & 76.42 & 87.00 & 87.59  & 92.85 & 93.52  &  96.74 &  97.38  \\
Layer 44 - layer4.0.conv1        & 524288 & 68.48 & 68.82 & 82.43 & 82.75  & 90.41 & 90.78  &  95.93 &  96.27  \\
Layer 45 - layer4.0.conv2        & 2359296 & 87.47 & 87.64 & 94.87 & 95.04  & 97.77 & 97.96  &  99.20 &  99.36  \\
Layer 46 - layer4.0.conv3        & 1048576 & 75.56 & 75.85 & 87.88 & 88.12  & 94.33 & 94.56  &  97.90 &  98.14  \\
Layer 47 - layer4.0.downsample.0 & 2097152 & 90.08 & 89.97 & 96.30 & 96.29  & 98.60 & 98.66  &  99.54 &  99.63  \\
Layer 48 - layer4.1.conv1        & 1048576 & 79.00 & 79.16 & 90.34 & 90.39  & 95.69 & 95.80  &  98.40 &  98.58  \\
Layer 49 - layer4.1.conv2        & 2359296 & 87.10 & 87.27 & 94.85 & 94.97  & 97.92 & 98.05  &  99.32 &  99.43  \\
Layer 50 - layer4.1.conv3        & 1048576 & 76.30 & 76.64 & 88.78 & 88.75  & 95.11 & 95.19  &  98.37 &  98.51  \\
Layer 51 - layer4.2.conv1        & 1048576 & 69.19 & 69.42 & 84.27 & 84.19  & 92.98 & 92.85  &  97.63 &  97.69  \\
Layer 52 - layer4.2.conv2        & 2359296 & 87.68 & 87.73 & 95.85 & 95.92  & 98.53 & 98.63  &  99.56 &  99.64  \\
Layer 53 - layer4.2.conv3        & 1048576 & 78.33 & 77.79 & 91.36 & 91.17  & 96.56 & 96.57  &  98.85 &  99.01  \\
Layer 54 - fc                    & 2048000 & 54.95 & 53.28 & 70.49 & 68.55  & 83.24 & 80.79  &  93.17 &  90.49  \\ \bottomrule
\end{tabular}}
\end{table}
\clearpage
\subsection{MobileNetV1}\label{sec:spar-mn50}

\begin{table}[!ht]\ra{1.2}
\centering
\caption{\small The obtained distribution of sparsity across the layers by WoodFisher and Global Magnitude when sparsifying \textsc{MobileNetV1} to $75\%, 89\%$ levels on \textsc{ImageNet}.}
\label{tab:mbv1nub}
\resizebox{0.6\columnwidth}{!}{
\begin{tabular}{@{}l|r|cc|cc@{}}
\toprule
Module                           & \multicolumn{1}{c|}{\begin{tabular}[c]{@{}c@{}}Fully Dense \\ Params\end{tabular}}  & \multicolumn{4}{c}{Sparsity (\%)} \\  \midrule
\multicolumn{1}{l|}{}                        & \multicolumn{1}{c|}{}                                                                               & WoodFisher    & GlobalMagni    & WoodFisher    & GlobalMagni      \\ 

Overall                                      & 4209088                                & \multicolumn{2}{c|}{75.28}                  & \multicolumn{2}{c}{89.00}    \\ \midrule
Layer 1                                      & 864 & 50.93 & 51.16 & 55.56 & 57.99 \\
Layer 2 (dw)                                 & 288 & 47.57 & 50.00 & 52.08 & 55.56 \\
Layer 3                                      & 2048 & 74.02 & 75.93 & 81.20 & 83.40 \\
Layer 4 (dw)                                 & 576 & 18.75 & 21.01 & 26.04 & 30.21 \\
Layer 5                                      & 8192 & 60.05 & 60.79 & 73.44 & 74.34 \\
Layer 6 (dw)                                 & 1152 & 30.30 & 31.86 & 39.84 & 43.75 \\
Layer 7                                      & 16384 & 58.16 & 58.69 & 73.55 & 74.29 \\
Layer 8 (dw)                                 & 1152 & 07.64 & 07.90 & 15.02 & 17.45 \\
Layer 9                                      & 32768 & 65.53 & 65.94 & 80.06 & 80.71 \\
Layer 10 (dw)                                & 2304 & 33.64 & 35.68 & 45.70 & 48.13 \\
Layer 11                                     & 65536 & 67.88 & 68.36 & 82.99 & 83.45 \\
Layer 12 (dw)                                & 2304 & 16.02 & 15.41 & 25.43 & 27.26 \\
Layer 13                                     & 131072 & 76.40 & 76.71 & 88.93 & 89.28 \\
Layer 14 (dw)                                & 4608 & 38.26 & 38.85 & 51.22 & 51.58 \\
Layer 15                                     & 262144 & 80.23 & 80.33 & 92.20 & 92.40 \\
Layer 16 (dw)                                & 4608 & 49.87 & 51.65 & 64.11 & 65.84 \\
Layer 17                                     & 262144 & 79.29 & 79.58 & 91.92 & 92.04 \\
Layer 18 (dw)                                & 4608 & 49.80 & 51.19 & 64.37 & 66.43 \\
Layer 19                                     & 262144 & 77.42 & 77.66 & 90.90 & 91.14 \\
Layer 20 (dw)                                & 4608 & 43.40 & 44.88 & 60.31 & 61.98 \\
Layer 21                                     & 262144 & 74.51 & 74.67 & 89.47 & 89.65 \\
Layer 22 (dw)                                & 4608 & 30.71 & 31.62 & 50.11 & 51.89 \\
Layer 23                                     & 262144 & 71.09 & 71.15 & 87.93 & 88.18 \\
Layer 24 (dw)                                & 4608 & 17.12 & 18.12 & 41.71 & 44.34 \\
Layer 25                                     & 524288 & 80.30 & 80.42 & 92.62 & 92.70 \\
Layer 26 (dw)                                & 9216 & 62.96 & 64.45 & 79.37 & 82.56 \\
Layer 27                                     & 1048576 & 87.58 & 87.57 & 96.67 & 96.80 \\
Layer 28 (fc)                                & 1024000 & 61.11 & 60.69 & 79.91 & 79.27 \\ 

\bottomrule
\end{tabular}
}
\end{table}
\clearpage

\section{FLOPs-aware sparsity distributions}\label{sec:app_spar_pattern_flops}
\begin{table}[!ht]\ra{1.2}
\centering
\caption{\small The obtained distribution of sparsity across the layers by WoodFisher when sparsifying \textsc{ResNet50} to $~90\%$ sparsity level \textbf{with FLOPs-aware hyperparameter $\beta=0.00, 0.30$} on \textsc{ImageNet}.}
\label{tab:resnet50nub_flops}
\resizebox{0.6\columnwidth}{!}{
\begin{tabular}{@{}l|r|cc@{}}
\toprule
Module                           & \multicolumn{1}{c|}{\begin{tabular}[c]{@{}c@{}}Fully Dense \\ Params\end{tabular}}  & \multicolumn{2}{c}{FLOPs-aware  ($\beta$)} \\  \midrule

Overall                                      & 4209088                                &\multicolumn{1}{c}{$\beta=0.00$ (usual)}   &  \multicolumn{1}{c}{$\beta=0.30$}   \\ \midrule
Layer 1 - conv1                  & 9408 & 44.97 &  62.83 \\
Layer 2 - layer1.0.conv1         & 4096 & 58.30 &  75.15 \\
Layer 3 - layer1.0.conv2         & 36864 & 79.48 &  91.88 \\
Layer 4 - layer1.0.conv3         & 16384 & 72.16 &  86.17 \\
Layer 5 - layer1.0.downsample.0  & 16384 & 66.31 &  80.66 \\
Layer 6 - layer1.1.conv1         & 16384 & 77.44 &  90.70 \\
Layer 7 - layer1.1.conv2         & 36864 & 82.65 &  93.73 \\
Layer 8 - layer1.1.conv3         & 16384 & 80.57 &  90.96 \\
Layer 9 - layer1.2.conv1         & 16384 & 73.18 &  89.09 \\
Layer 10 - layer1.2.conv2        & 36864 & 73.37 &  89.84 \\
Layer 11 - layer1.2.conv3        & 16384 & 79.29 &  89.34 \\
Layer 12 - layer2.0.conv1        & 32768 & 71.70 &  87.52 \\
Layer 13 - layer2.0.conv2        & 147456 & 83.83 &  92.60 \\
Layer 14 - layer2.0.conv3        & 65536 & 82.89 &  90.05 \\
Layer 15 - layer2.0.downsample.0 & 131072 & 88.98 &  94.45 \\
Layer 16 - layer2.1.conv1        & 65536 & 89.38 &  94.91 \\
Layer 17 - layer2.1.conv2        & 147456 & 87.38 &  94.12 \\
Layer 18 - layer2.1.conv3        & 65536 & 85.39 &  91.99 \\
Layer 19 - layer2.2.conv1        & 65536 & 84.34 &  92.60 \\
Layer 20 - layer2.2.conv2        & 147456 & 87.02 &  94.08 \\
Layer 21 - layer2.2.conv3        & 65536 & 82.19 &  90.64 \\
Layer 22 - layer2.3.conv1        & 65536 & 82.43 &  91.20 \\
Layer 23 - layer2.3.conv2        & 147456 & 84.80 &  93.00 \\
Layer 24 - layer2.3.conv3        & 65536 & 84.12 &  91.62 \\
Layer 25 - layer3.0.conv1        & 131072 & 76.43 &  87.23 \\
Layer 26 - layer3.0.conv2        & 589824 & 91.43 &  94.65 \\
Layer 27 - layer3.0.conv3        & 262144 & 84.20 &  89.21 \\
Layer 28 - layer3.0.downsample.0 & 524288 & 94.23 &  96.50 \\
Layer 29 - layer3.1.conv1        & 262144 & 93.09 &  95.76 \\
Layer 30 - layer3.1.conv2        & 589824 & 93.25 &  95.89 \\
Layer 31 - layer3.1.conv3        & 262144 & 88.41 &  92.44 \\
Layer 32 - layer3.2.conv1        & 262144 & 92.21 &  95.19 \\
Layer 33 - layer3.2.conv2        & 589824 & 93.31 &  96.04 \\
Layer 34 - layer3.2.conv3        & 262144 & 88.91 &  92.97 \\
Layer 35 - layer3.3.conv1        & 262144 & 90.86 &  94.33 \\
Layer 36 - layer3.3.conv2        & 589824 & 93.32 &  96.13 \\
Layer 37 - layer3.3.conv3        & 262144 & 89.93 &  93.66 \\
Layer 38 - layer3.4.conv1        & 262144 & 90.20 &  94.02 \\
Layer 39 - layer3.4.conv2        & 589824 & 93.24 &  96.11 \\
Layer 40 - layer3.4.conv3        & 262144 & 89.26 &  93.31 \\
Layer 41 - layer3.5.conv1        & 262144 & 87.68 &  92.40 \\
Layer 42 - layer3.5.conv2        & 589824 & 92.72 &  95.83 \\
Layer 43 - layer3.5.conv3        & 262144 & 87.00 &  91.79 \\
Layer 44 - layer4.0.conv1        & 524288 & 82.43 &  88.34 \\
Layer 45 - layer4.0.conv2        & 2359296 & 94.87 &  95.23 \\
Layer 46 - layer4.0.conv3        & 1048576 & 87.88 &  88.61 \\
Layer 47 - layer4.0.downsample.0 & 2097152 & 96.30 &  96.64 \\
Layer 48 - layer4.1.conv1        & 1048576 & 90.34 &  90.97 \\
Layer 49 - layer4.1.conv2        & 2359296 & 94.85 &  95.23 \\
Layer 50 - layer4.1.conv3        & 1048576 & 88.78 &  89.41 \\
Layer 51 - layer4.2.conv1        & 1048576 & 84.27 &  85.22 \\
Layer 52 - layer4.2.conv2        & 2359296 & 95.85 &  96.25 \\
Layer 53 - layer4.2.conv3        & 1048576 & 91.36 &  92.18 \\
Layer 54 - fc                    & 2048000 & 70.49 &  50.50 \\ 
\midrule
Total Sparsity & & 90.00\% & 90.23\%\\
FLOPs & & 594 M & 335 M \\
\bottomrule
\end{tabular}
}
\end{table}

\begin{table}[!ht]\ra{1.2}
\centering
\caption{\small The obtained distribution of sparsity across the layers by WoodFisher when sparsifying \textsc{MobileNetV1} to $75.28\%$ sparsity level \textbf{with FLOPs-aware hyperparameter $\beta=0.00, 0.30, 0.35$} on \textsc{ImageNet}.}
\label{tab:mbv1nub_flops}
\resizebox{0.6\columnwidth}{!}{
\begin{tabular}{@{}l|r|ccc@{}}
\toprule
Module                           & \multicolumn{1}{c|}{\begin{tabular}[c]{@{}c@{}}Fully Dense \\ Params\end{tabular}}  & \multicolumn{3}{c}{FLOPs-aware  ($\beta$)} \\  \midrule

Overall                                      & 4209088                                &\multicolumn{1}{c}{$\beta=0.00$ (usual)}   &  \multicolumn{1}{c}{$\beta=0.30$}                  & \multicolumn{1}{c}{$\beta=0.35$}    \\ \midrule
Layer 1                                      & 864 & 50.93 & 62.15 & 64.35 \\
Layer 2 (dw)                                 & 288 & 47.57 & 57.29 & 58.33 \\
Layer 3                                      & 2048 & 74.02 & 85.01 & 86.57 \\
Layer 4 (dw)                                 & 576 & 18.75 & 31.25 & 33.16 \\
Layer 5                                      & 8192 & 60.05 & 77.81 & 80.57 \\
Layer 6 (dw)                                 & 1152 & 30.30 & 47.22 & 50.00 \\
Layer 7                                      & 16384 & 58.16 & 78.97 & 81.96 \\
Layer 8 (dw)                                 & 1152 & 07.64 & 16.32 & 18.92 \\
Layer 9                                      & 32768 & 65.53 & 79.57 & 81.77 \\
Layer 10 (dw)                                & 2304 & 33.64 & 47.61 & 50.35 \\
Layer 11                                     & 65536 & 67.88 & 82.74 & 84.92 \\
Layer 12 (dw)                                & 2304 & 16.02 & 20.36 & 21.18 \\
Layer 13                                     & 131072 & 76.40 & 83.70 & 84.96 \\
Layer 14 (dw)                                & 4608 & 38.26 & 46.18 & 47.31 \\
Layer 15                                     & 262144 & 80.23 & 87.68 & 88.87 \\
Layer 16 (dw)                                & 4608 & 49.87 & 58.79 & 60.76 \\
Layer 17                                     & 262144 & 79.29 & 87.11 & 88.49 \\
Layer 18 (dw)                                & 4608 & 49.80 & 59.64 & 61.09 \\
Layer 19                                     & 262144 & 77.42 & 85.76 & 87.17 \\
Layer 20 (dw)                                & 4608 & 43.40 & 55.73 & 57.44 \\
Layer 21                                     & 262144 & 74.51 & 83.81 & 85.29 \\
Layer 22 (dw)                                & 4608 & 30.71 & 47.57 & 50.24 \\
Layer 23                                     & 262144 & 71.09 & 81.08 & 82.82 \\
Layer 24 (dw)                                & 4608 & 17.12 & 24.11 & 24.74 \\
Layer 25                                     & 524288 & 80.30 & 81.96 & 82.39 \\
Layer 26 (dw)                                & 9216 & 62.96 & 60.45 & 61.56 \\
Layer 27                                     & 1048576 & 87.58 & 88.60 & 88.94 \\
Layer 28 (fc)                                & 1024000 &  61.11 & 45.05 & 42.13 \\
\midrule
FLOPs & & 159 M & 101 M & 92 M \\
\bottomrule
\end{tabular}
}
\end{table}
\clearpage

\section{WoodTaylor Results}\label{sec:app_woodtaylor}

\subsection{Pre-trained model}\label{sec:app_woodtaylor_pre}
\begin{figure}[h!]
	\centering
	\includegraphics[width=0.55\linewidth]{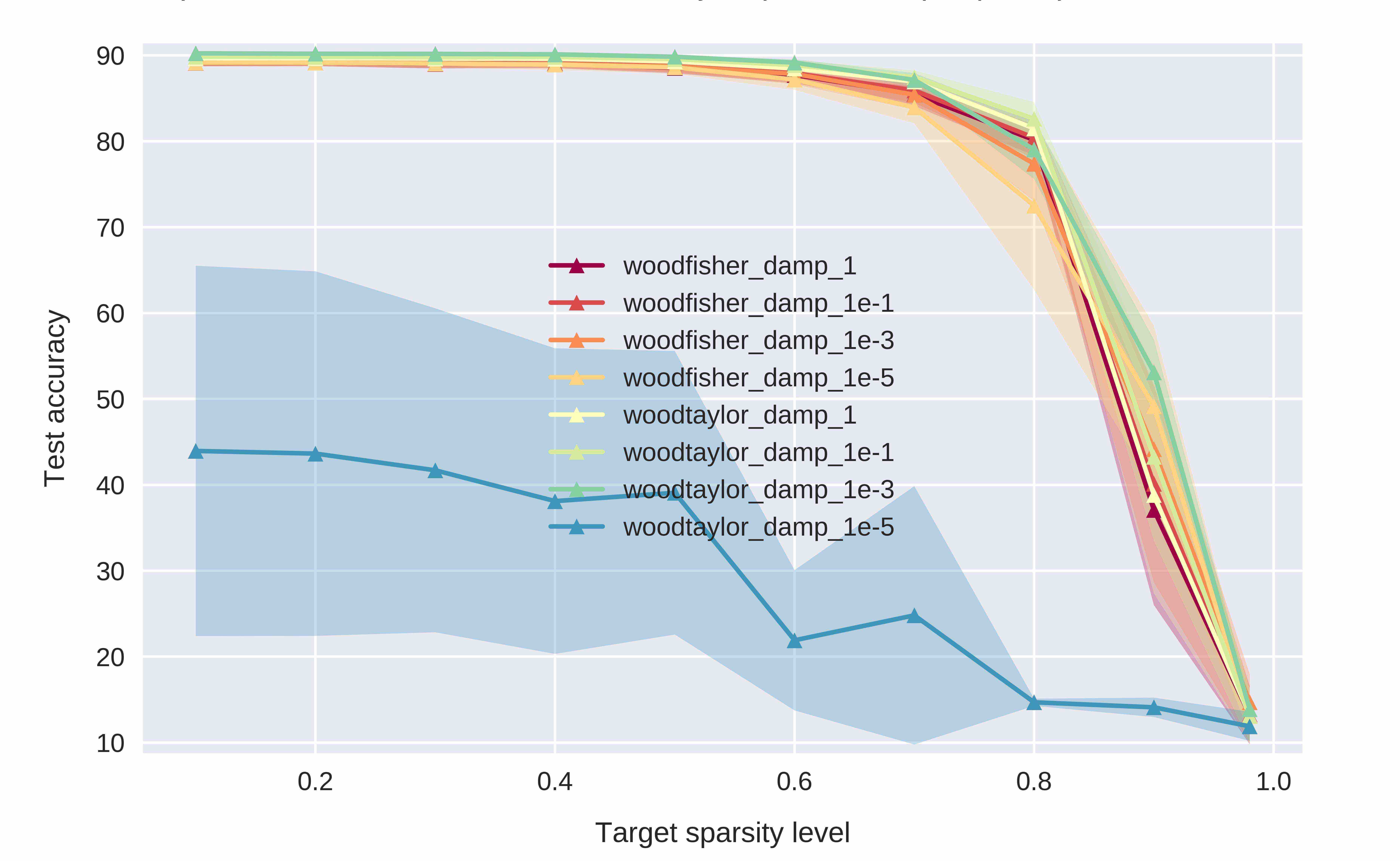}
	\caption{Comparing one-shot sparsity results for WoodTaylor and WoodFisher on CIFAR-10 for ResNet-20.}
	\label{fig:one-shot-woodtaylor}
\end{figure}

\begin{figure}[h]
	\centering
	\includegraphics[width=0.8\linewidth]{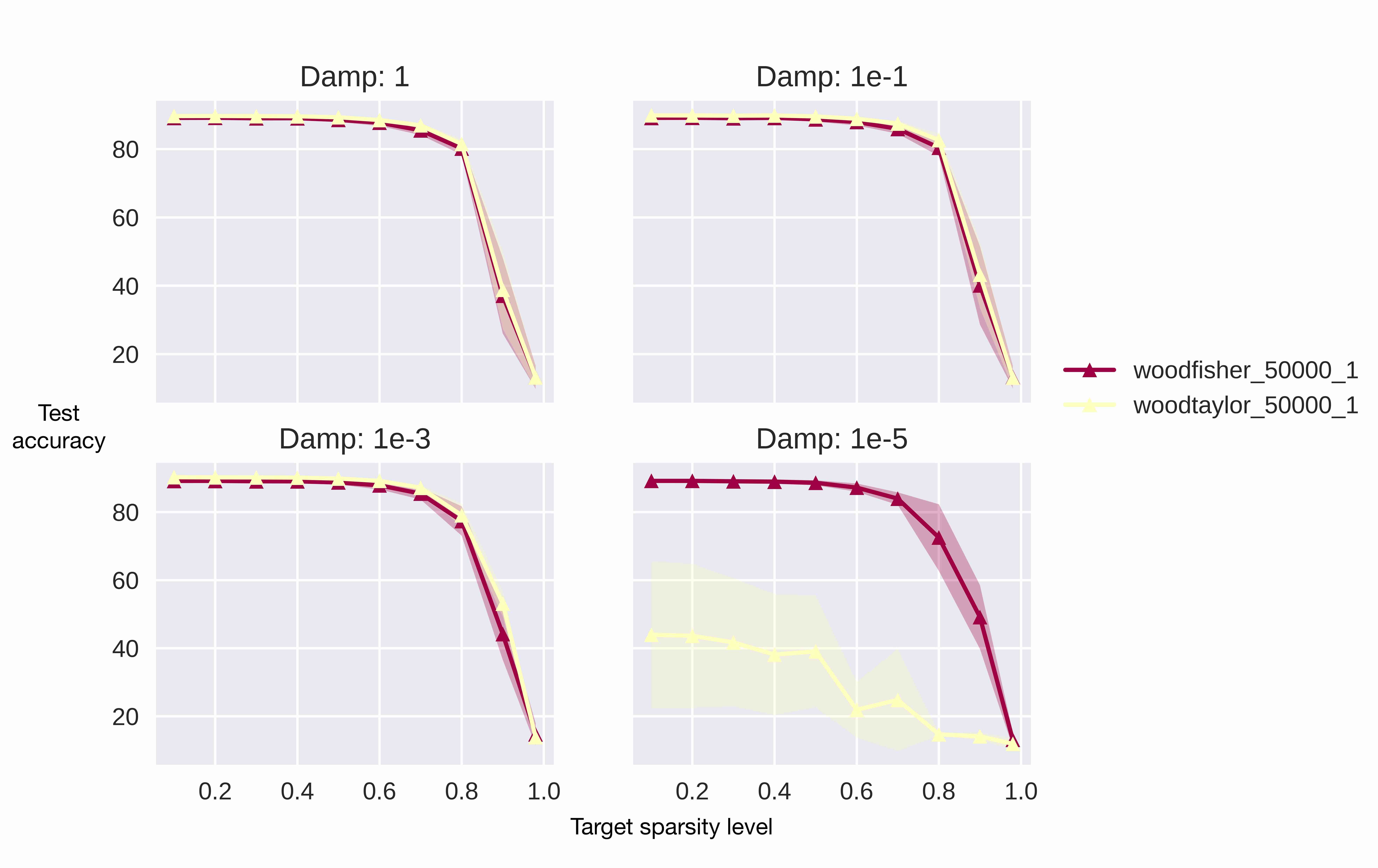}
	\caption{A simplified comparison of one-shot sparsity results for WoodTaylor and WoodFisher on CIFAR-10 for ResNet-20.}
	\label{fig:ablation-woodtaylor-simplified}
\end{figure}

Next, we focus on the comparison between WoodFisher and WoodTaylor for the setting of ResNet-20 pre-trained on CIFAR10, where both the methods are used in their `full-matrix' mode. In other words, no block-wise assumption is made, and we consider pruning only the `layer1.0.conv1', `layer1.0.conv2' and `layer2.0.conv1'. In Figures~\ref{fig:one-shot-woodtaylor}, ~\ref{fig:ablation-woodtaylor-simplified}, we present the results of one-shot experiments in this setting. 
We observe that WoodTaylor (with damp=$1e-3$) outperforms WoodFisher (across various dampening values) for almost all levels of target sparsity. This confirms our hypothesis of factoring in the gradient term, which even in this case where the model has relatively high accuracy, can lead to a gain in performance. 
However, it is important to that in comparison to WoodFisher, WoodTaylor is more sensitive to the choice of hyper-parameters like the dampening value, as reflected in the Figure~\ref{fig:one-shot-woodtaylor}. This arises because now in the weight update, Eqn.~\eqref{eq:grad_opt_delw}, there are interactions between the Hessian inverse and gradient terms, due to which the scaling of the inverse Hessian governed by this dampening becomes more important. To give an example, in the case where damp=$1e-5$, the resulting weight update has about $10\times$ bigger norm than that of the original weight. 

This can be easily adjusted via the dampening, but unlike WoodFisher, it is not hyper-parameter free. Also, for these experiments, the number of samples used was 50,000, which is higher in comparison to our previously used values.  

\subsection{Ablation Study}
\begin{figure}[h]
	\centering
	\includegraphics[width=\linewidth]{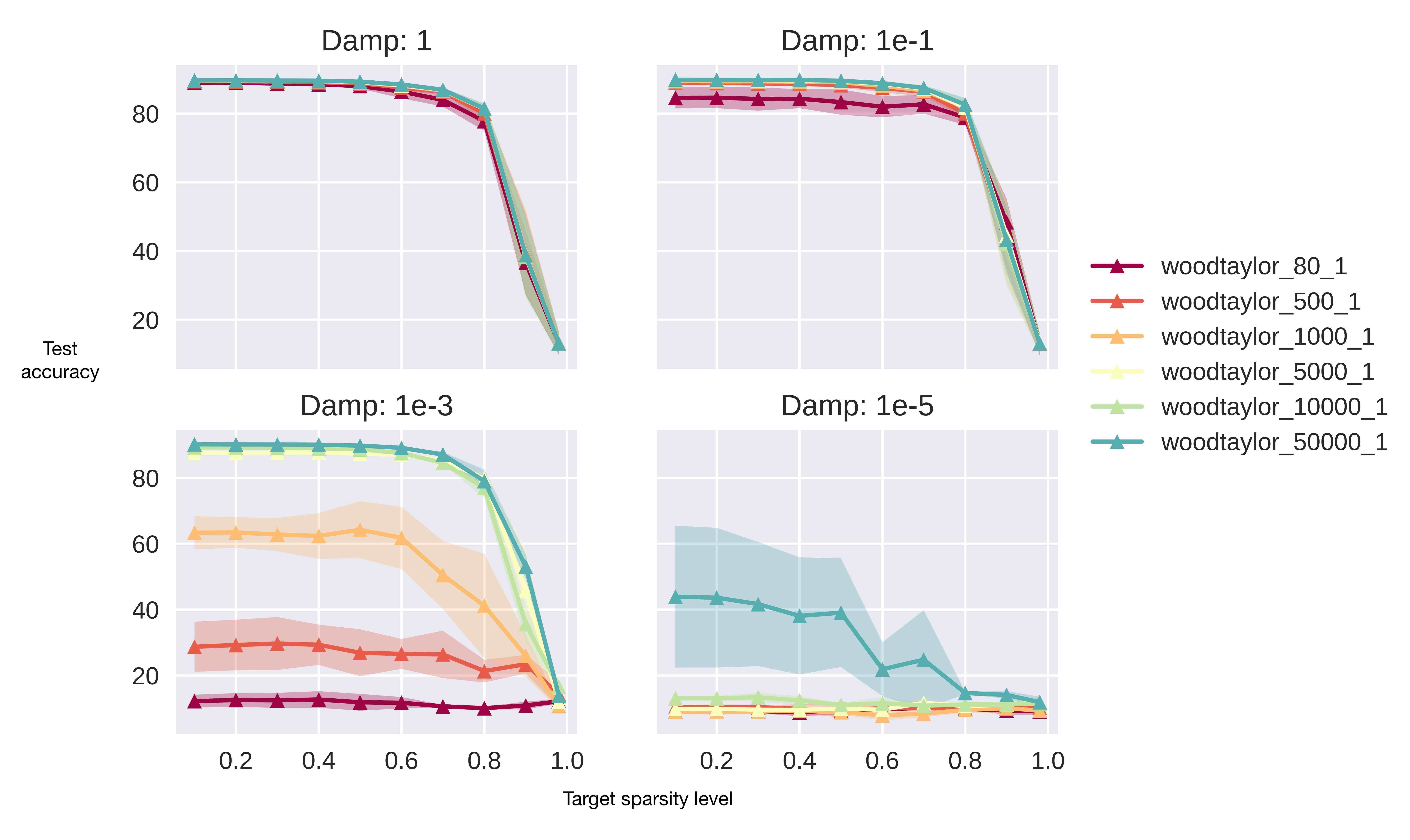}
	\caption{Ablation study for WoodTaylor that shows the effect of dampening and the number of samples used on the performance.}
	\label{fig:ablation-woodtaylor}
\end{figure}

In order to better understand the sensitivity of WoodTaylor with respect to these hyper-parameter choices, we present an ablation study in Figure~\ref{fig:ablation-woodtaylor} that measures their effect on WoodTaylor's performance.

In the end, we conclude that incorporating the first-order term helps WoodTaylor to gain in performance over WoodFisher, however, some hyper-parameter tuning for the dampening constant might be required. Future work would aim to apply WoodTaylor in the setting of gradual pruning discussed in Section~\ref{sec:app_gradual_detail}.

\end{document}